\documentclass[preprint,12pt,authoryear]{elsarticle}
\usepackage[T1]{fontenc}

\usepackage{siunitx}
\usepackage{multirow}
\usepackage{booktabs}
\usepackage{graphicx}
\usepackage[export]{adjustbox}
\usepackage{subcaption} 
\usepackage{pdflscape}

\usepackage[english]{babel}

\usepackage{amsmath}
\usepackage{amssymb}
\usepackage{algorithm}
\usepackage{algpseudocode}
\usepackage[table]{xcolor}
\DeclareMathOperator{\I}{I} 

\DeclareMathOperator*{\argmax}{argmax}
\newcommand{\x}{{\bf x}}

\newcommand{\T}{\mathcal{T}}
\newcommand{\M}{\mathcal{M}}

\newcommand{\R}{\mathbb{R}}

\usepackage[letterpaper,top=2cm,bottom=2cm,left=3cm,right=3cm,marginparwidth=1.75cm]{geometry}

\usepackage[colorlinks=true, allcolors=blue]{hyperref}
\usepackage{cleveref}
\crefname{figure}{Figure}{Figures} 
\Crefname{figure}{Figure}{Figures} 

\journal{Journal of Visual Communication and Image Representation}

\begin{document}

\begin{frontmatter}



\title{Realistic Continual Learning Approach \\ using Pre-trained Models}
\author[uah,ua]{Nadia Nasri}
\affiliation[uah]{organization={University of Alcalá, Department of Signal Theory and Communications, GRAM Research group},
            city={Alcalá de Henares},
            postcode={28805}, 
            country={Spain}}
            
\affiliation[ua]{organization={University of Alicante, Department of Computer Science and Artificial Intelligence},
            city={San Vicente del Raspeig},
            postcode={03690}, 
            country={Spain}}
\author[uah]{Carlos Gutiérrez-Álvarez}
\author[uah]{Sergio Lafuente-Arroyo}
\author[uah]{Saturnino Maldonado-Bascón}
\author[uah]{Roberto J. López-Sastre}

\begin{abstract}
Continual learning (CL) evaluates adaptability in learning solutions to retain knowledge. Our research addresses the challenge of catastrophic forgetting, where models lose proficiency in previously learned tasks as they acquire new ones. While numerous solutions have been proposed, existing experimental setups often rely on idealized class-incremental learning scenarios. We introduce Realistic Continual Learning (RealCL), a novel CL paradigm where class distributions across tasks are random. We also present CLARE (Continual Learning Approach with pRE-trained models for RealCL scenarios), a pre-trained model-based solution designed to integrate new knowledge while preserving past learning. Our contributions include pioneering RealCL as a generalization of traditional CL setups, proposing CLARE as an adaptable approach for RealCL tasks, and conducting extensive experiments demonstrating its effectiveness across various RealCL scenarios. Notably, CLARE outperforms existing models on RealCL benchmarks, highlighting its versatility in unpredictable learning environments.
Code to reproduce all our experiments can be found at \url{https://github.com/gramuah/clare}.
\end{abstract}


\end{frontmatter}

\section{Introduction}
Continual learning (CL), also referred to as lifelong or incremental learning, is a vibrant sub-field of machine learning focused on creating algorithms that can learn from sequential input data streams in an incremental manner. Its major purpose is to mimic human beings’ ability to continuously gather and refine new information over time, allowing models to adapt to new information without sacrificing the previously learned one. The implication of CL is built on the notion that artificial intelligence (AI) should become more robust and adaptable in an ever-changing world.
In other words, defining an experimental setup for continual learning requires structuring learning into tasks and evaluating how the accuracy of proposed solutions evolves as tasks progress. 
Is the model able to maintain good recognition accuracy in the last task it was evaluated on? 
How has the degradation been in previous tasks when this new task was learned? 
The main factor driving research in CL is what is known as catastrophic forgetting. 
It is well known that in a continual learning training scheme based on fine-tuning for deep learning models, for example, when we adjust the model's weights to learn the last task, there is no guarantee that the model will retain the knowledge it has previously learned \citep{French1993CatastrophicII,McCloskey1989CatastrophicII}. 
What we observe is called catastrophic forgetting, as the performance in tasks where the model was previously an expert gradually decreases.

\begin{figure}[t]
\centering
\includegraphics[width=0.80\linewidth]{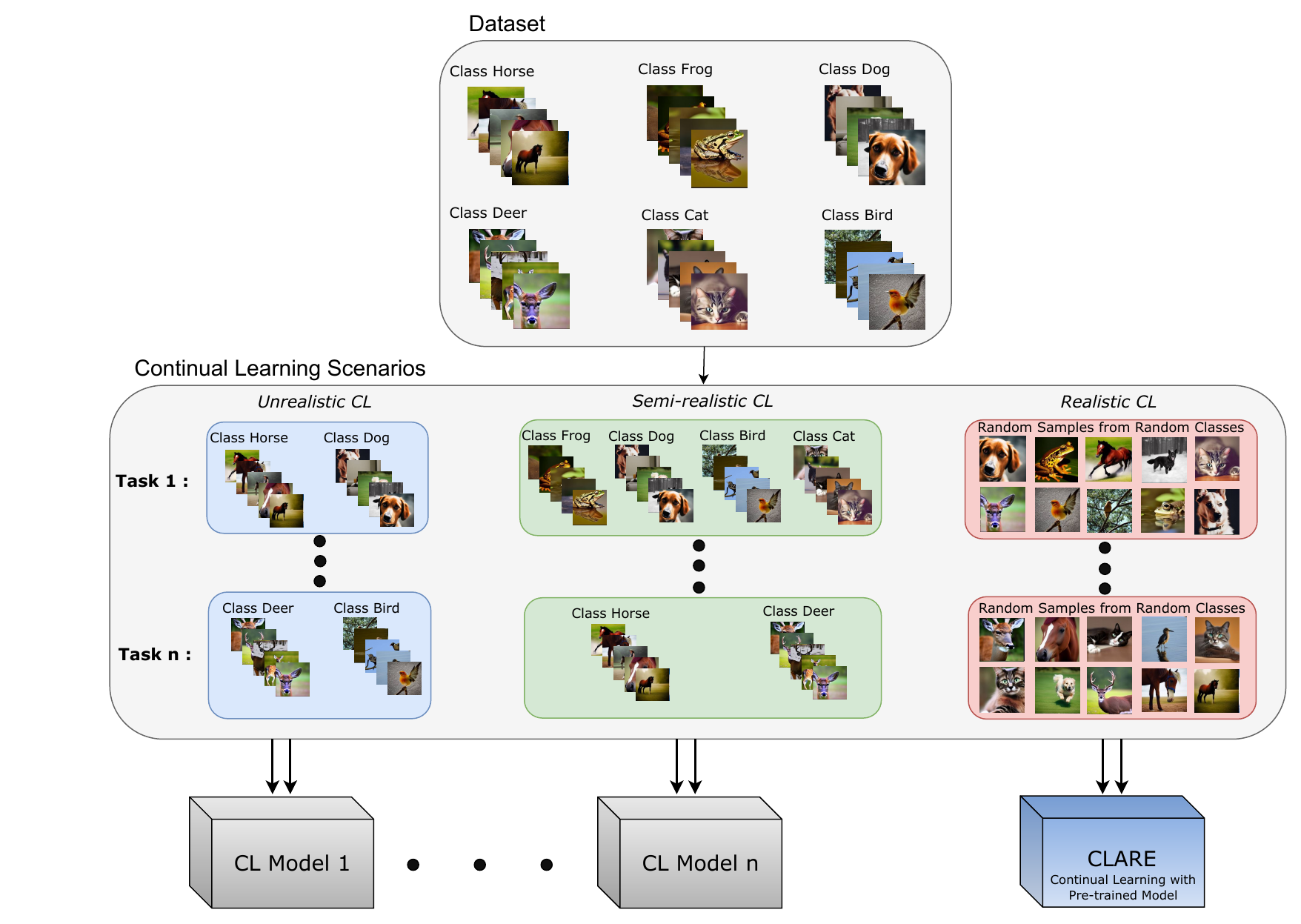}
\caption{\label{fig:abstractfig} 
In this figure, we illustrate the possible experimental evaluation scenarios that can be used to validate continual learning solutions. The scenario referred to as Unrealistic CL is the most commonly used in the literature, also known as the class-incremental setup. In this work, we propose to explore the RealCL scenario as a generalization of all continual learning setups. In RealCL, the categories to be recognized by CL models are not distributed in a structured manner across tasks, but rather randomly. Halfway between Unrealistic CL and RealCL, we have the semi-realistic scenario, where it is tolerated for classes to be organized across tasks, but the number classes assigned to each task is not controlled. In this work, we provide an experimental evaluation in the new RealCL scenario of both our novel CLARE proposal and the state-of-the-art CL models.
}
\end{figure}

Research in CL aims to mitigate the impact of this catastrophic forgetting. 
Numerous solutions have been proposed based on different principles (e.g. \citep{Shi2023,Aljundi2019GradientBS, Buzzega2020RethinkingER, Shim2020OnlineCC, Zhao2020MemoryEfficientCL, Castro2018EndtoEndIL,Wang2023BEEFBC, Douillard2021DyToxTF, Zhou2022AMO,Zhao2019MaintainingDA, Li2016LearningWF, Zhou2019M2KDMA, Douillard2020PODNetPO}), but they all share an aspect that in our opinion has gone unnoticed: the idealized or unrealistic experimental setup. 
To compare different continual learning models, researchers have defined different experimental setups in which we simulate a continuous learning process by controlling the distribution of data into different tasks. 
We take a specific dataset, such as CIFAR-100, and organize the different object categories into tasks, so we ask our continual learning models to gradually learn to recognize, for example, between the categories dog and cat first. Then in a second task between the categories car, motorcycle, dog and cat. And so on. This is what is called a class-incremental learning scenario.
The models proposed in the literature present ad hoc solutions for this experimental situation. In other words, the models have been specifically designed to optimize their learning when categories grow in a structured and incremental manner. 
The question we ask from this work is: what happens when this assumption is not met? 
Imagine a realistic continual learning environment where these idealizations do not exist. The data that the continual learning model has to face is also structured into tasks, but the distribution of categories is not controlled or structured but randomly formed. Categories like dog and cat may appear in task 1, but they may also reappear later in task $n$. In this work, we propose to advance in the problem of continual learning by focusing on how state-of-the-art models have benefited from this idealization of the experimental setup. Therefore, we propose a generalization of the problem that we have called Realistic CL (RealCL), in which CL models will have to deal with task streams where the class distribution is not structured (see Figure \ref{fig:abstractfig}).

To explore this new RealCL scenario, we also introduce a novel solution for continual learning that exploits the benefits of using what is known as pre-trained models (e.g. \citep{Radford2021LearningTV,Dosovitskiy2020AnII}).
Our model is named CLARE, for a Continual Learning Approach with pRE-trained models for RealCL scenarios.
Distinguished by its simplicity and pre-trained-model-agnostic nature, CLARE leverages pre-trained models to seamlessly integrate new knowledge while preserving past learning.
This breakthrough eliminates the need for extensive retraining learning process.
In summary, our main contributions are three-fold:
\begin{itemize}
    \item We pioneer the challenging task of RealCL, a new continual learning setup where the class distribution is not structured across tasks. We proof that RealCL is the generalization of all the standard CL scenarios typically employed in all previous scientific works. In our study we explore a spectrum of CL scenarios, encompassing highly structured to entirely unstructured. Our objective is to reveal whether the state-of-the-art CL models have taken advantage of the artificial and forced structuring that characterizes traditional experimental evaluation scenarios, evaluating  them in the novel RealCL scenario.    
    \item We propose the pre-trained model based continual learning strategy CLARE, which is an innovative and adaptable approach tailored for RealCL tasks. CLARE is based on the ability that pre-trained models seem to exhibit to capture and retain knowledge when combined with a specific dynamic module on which learning is performed.    
    \item Extensive experiments are conducted to verify the effectiveness of our CLARE model in all the continual learning scenarios that can be derived from RealCL. Impressively, our method achieves a strong performance even improving some state-of-the-art models when tested on the RealCL benchmarks. Through meticulous comparisons, we assess CLARE's performance against alternative approaches, demonstrating its versatility and robustness in handling the unpredictability imposed by the RealCL setup, which simulates what a realistic CL scenario would be like, where we have no control over the tasks that the models are being confronted with.
\end{itemize}


\section{Related Work} \label{sec:related_work}
\subsection{Continual learning and its idealized evaluation scenarios} 
Continual learning is a sub-field of machine learning concerned with how models can be trained on a stream of data arriving sequentially without performance decay or forgetting. This requires solving the stability-plasticity trade-off, which is a core problem in CL \citep{Mermillod2013TheSD, Lee2022BalancingTS, Araujo2022EntropybasedSF}. This trade-off refers to how a model should balance retaining old knowledge (stability) and adding new knowledge (plasticity).
Therefore, it is a problem of catastrophic forgetting that stems from the necessity to combine two fairly outstanding strengths – stability and plasticity \citep{French1993CatastrophicII,McCloskey1989CatastrophicII}. Where stability makes sure a model holds on to the knowledge it previously gained and plasticity allows it to integrate new data. However, quantifying the importance of these two characteristics and finding an algorithm that accounts for both puts the upfront rule in this task. Usually, the most recent data takes the most weight, eroding previous insights, which is a phenomenon of catastrophic forgetting. It is most fatal in tasks that need to adjust to ever-changing datasets and new contexts. Given this obstacle, research in the area of CL has shown certain ways that are trying to reduce catastrophic forgetting, so as not to lose valuable knowledge and transfer it between different tasks. 

Three main techniques are employed in the realm of CL to address the issue of catastrophic forgetting. 
The first group relies on replaying old samples or exemplars from previous classes, and it is commonly known as rehearsal \citep{Aljundi2019GradientBS, Buzzega2020RethinkingER, Shim2020OnlineCC, Zhao2020MemoryEfficientCL, Castro2018EndtoEndIL}. In this method, data from previous tasks is sampled based on some strategy and fed to the model, which is then again integrated with new data during the training process. Rehearsal was initially presented as a prevalent and most effective approach in CL literature due to its potential to retain diversity and representation of data across tasks and avoid the forgetting of previous knowledge \citep{Lomonaco2020CVPR2C}.

The second group addresses the challenge of limited representation capacity by dynamically expanding the network architecture upon the introduction of new tasks \citep{Wang2023BEEFBC, Douillard2021DyToxTF, Zhou2022AMO}. In this case, it is ensured by network self-growing in a dynamic way upon each arriving task. Techniques involving the addition of new neurons, layers, or modules enable this smooth integration of new information without disturbing previously acquired knowledge. The improvement of the feature space expressiveness and its diversity makes these kinds of expansions reduce the model parameter interference and saturation. Some well-known examples include progressive neural networks, dynamic expandable networks, and lifelong learning via network growth domains.

The final category of CL strategies is based on knowledge distillation, where the objective is to facilitate knowledge extraction from old models and transfer it into new ones \citep{Zhao2019MaintainingDA, Li2016LearningWF, Zhou2019M2KDMA, Douillard2020PODNetPO}. This technique ensures that the transfer of knowledge from a larger model to a smaller model does not compromise its validity. In terms of CL, knowledge distillation makes sure that outputs of older models are aligned with those of newer ones such that output probabilities in the new model match those in the old model when subjected to the same input. This synchronization guarantees the retention of knowledge pertaining to earlier concepts, thereby preventing forgetting during the acquisition of new tasks. It is important to acknowledge that some of the studies do not fit to one type of CL methods since they leverage different techniques in combination for getting better outcomes.

Most of the above-mentioned works were conducted within the context of well-defined experimental scenarios, where the classes were also balanced across the tasks. Although such experiments are indispensable for studying the catastrophic forgetting phenomenon, they often fail because real-world data streams are dynamic and non-stationary. Considering the probability of some classes reappearing, it implies that class distributions and their frequencies in these data streams can dynamically alter from task to task in ways that could seem unexpected for the model.

In other studies, there have been several other initiatives with differing settings of continual learning but no less important; which strongly reflect the value of proper realistic evaluation conditions \citep{Chen2019,vandeVen2019ThreeSF, dalange2021, vandeVen2022ThreeTO}. Yet, it is worth noting that the data distribution in these scenarios often remains fixed or controlled, stipulated with class frequencies for the task. The diversities among these scenarios are essential because they reveal different perspectives on how continual learning can be done in real-world situations. 

In our work, we present a new Realistic Continual Learning (RealCL) scenario, carefully designed to learn from a non-stationary series of tasks. This scenario is unique because it does not control the distribution of classes across tasks or the recurrence of a class. Our approach is intentionally crafted to reflect the natural distribution of the data stream, capturing the diversity and complexity found in real-world situations.

\subsection{Pre-trained Models for CL}

Recently, high-performance and flexible pre-trained models (PTMs) have been widely adopted as backbones for continual learning. In particular, the Vision Transformer (ViT) \citep{Dosovitskiy2020AnII} has become one of the most popular architectures due to its strong representation capabilities. The original ViT architecture processes an image as a sequence of embedded image patches using a standard transformer encoder and contains only the parameters associated with the patch embedding, positional embeddings, transformer blocks, and classification token.

Building upon this architecture, several prompt-based continual learning methods introduce a small set of additional learnable prompt parameters that are optimized while keeping the pre-trained ViT backbone frozen or largely unchanged. 
These prompts act as task-specific conditioning tokens, enabling the model to retrieve task-relevant knowledge while mitigating catastrophic forgetting and promoting knowledge transfer across tasks. 
Representative methods include L2P \citep{Wang2021LearningTP}, DualPrompt \citep{Wang2022DualPromptCP}, S-Prompt \citep{Wang2022SPromptsLW}, DAP \citep{Jung2023GeneratingIP}, CODA-Prompt \citep{Smith2022CODAPromptCD}, and other related approaches \citep{Wang2022IsolationAI}. 
In some cases, prompt learning has also been extended to multimodal continual learning scenarios by jointly learning image and text prompts to better exploit complementary visual and semantic information \citep{Wang2022SPromptsLW, Zhou2021LearningTP}.

Recent advances have taken advantage of the generalizable embeddings of PTMs to address the CL challenges and propose an innovative framework, AdaPt and mERge (APER) \citep{aper}, that significantly outperforms traditional methods without requiring downstream task training. It does this by combining the adaptivity of model updating with the generalizability of knowledge transfer.

Different methodologies, in the alternative direction, tend to have very low parameter costs while they also allow effective adaptation of PTMs to downstream tasks through focusing the fine-tuning process on a few parameters. In particular, approaches like Visual Prompt Tuning (VPT) \citep{Jia2022VisualPT} add prefix tokens that are tunable, and Low-Rank Adaptation (LoRA) \citep{Hu2021LoRALA}, which employs low-rank matrices in updating the parameters, are methods that can achieve close to the fine-tuned model accuracy at negligible computational costs.

These methods, however, have their limitations despite the progress made. In most cases, prompts have few static dimensions and a simplistic way of selecting them. On the other hand, the performance of a tuning method, which is also parameter-efficient, has to balance with the number of parameters it can tune. But still, this integration process did turn out to be very well-working because it enables CL to work with PTMs and they both provide this potential for knowledge preservation; as such models outperform scratch-trained models and open new vistas for research in this area \citep{Cao2023RetentiveOF}.

Building on the progress of PTMs, in this paper we introduce CLARE, a novel model that is able to exploit the powerful representation of a frozen PTM, combined with a Dynamic Neural Adaptation Network (Dyn-NAN) module. We do not follow any prompting-based pipeline, but propose with CLARE a simple and effective architecture that exhibits a remarkable performance in the realistic CL setting proposed in this work, as well as in the traditional CL scenarios.
Unlike standard replay-based methods, CLARE introduces this novel combination of a frozen pre-trained model with a Dyn-NAN.
This architecture enables efficient task adaptation without retraining the pre-trained model, focusing instead on dynamically adjusting the Dyn-NAN module to integrate new knowledge while minimizing interference with previously learned tasks.
By leveraging the strong representations of pre-trained models, CLARE not only mitigates forgetting but also achieves robustness in unstructured and realistic continual learning scenarios, distinguishing it from traditional replay approaches.

\section{From Unrealistic to Realistic Continual Learning}\label{sec:realistic}
In this section we detail the novel realistic continual learning setting we aim to explore in this work.
We demonstrate that the standard scenarios typically employed in all previous scientific works function as mere specializations of our realistic scenario, given its broader scope, which encompasses them all.

\subsection{Realistic Continual Learning}
Realistic continual learning (RealCL) aims to learn from an evolving task stream where no control over the distribution of new classes in the new tasks is applied.
The objective of a RealCL approach is to learn a model for a large number of tasks sequentially, without forgetting knowledge obtained from the preceding tasks, where the data in the old tasks are not available anymore during training new ones.
We propose a training scheme that does not impose \emph{any} control over how the data stream is distributed across different tasks, as it gradually arrives.
Neither the classes nor the distribution of samples arriving with each task are controlled. 
Note that this is different from a naive fine-tuning based training process in CL, where the new tasks provide access to new categories, typically balancing the number of features and the class distributions.
In our RealCL, all the task definition process is random.
We assume, therefore, that the data flow randomly into our system, and once the data has been used for learning, our model cannot access it as new tasks arrive. 

We now formalize the mathematical problem for RealCL, which will allow us, in Section \ref{sec:unreal_scenarios}, to proof how traditional scenarios used in CL are particularizations of our proposal.
In a RealCL setting we assume that there is a sequence of $K$ training tasks 
$\left\{\mathcal{T}^{1}, \mathcal{T}^{2},\cdots, \right.$
$\left.\mathcal{T}^{K}\right\}$, where $\mathcal{T}^{k}=\left\{\left(\mathbf{x}_{i}^{k}, y_{i}^{k}\right)\right\}_{i=1}^{n_k}$ is the $k$-th incremental step with $n_k$ instances.
Here, the training instance $\x_i^k \in \R^D$, where $D$ is the dimensionality of the feature space, belongs to class $y_i \in Y_k$, where $Y_k$ is the label space of task $k$.
Note that this label space is random, so there is no control over the labels that should appear in each task.
Furthermore, another property of the proposed RealCL paradigm is that the original training data from previous tasks are no longer accessible once a task has been completed. In other words, during incremental step $k$, the learner cannot revisit the \emph{complete} datasets associated with previous tasks. This assumption reflects realistic deployment scenarios in which historical data cannot be permanently stored or reloaded.
Notice that this restriction does not preclude continual learning algorithms from maintaining a limited auxiliary memory (e.g., a replay buffer or a compact set of exemplars), provided that the complete historical datasets remain unavailable.

The aim of RealCL is to incrementally construct a unified model encompassing all observed classes, i.~e. assimilating information from new classes while simultaneously retaining knowledge from previous ones.
During training, the learner sequentially updates the model using only the current training task $\mathcal{T}^{k}$, without accessing previous training data. The learning objective at each incremental step is therefore defined exclusively over the available training set. Overall, the goal is to learn a model $f(\x): \mathbb{R}^D \rightarrow \mathcal{Y}_k$ using only the training data available at each incremental step.

After completing task $k$, the model is evaluated on the union of the \emph{testing} sets corresponding to all observed tasks in order to assess both knowledge acquisition and forgetting, i.e. $\mathcal{T}_{\text{test}}^1\cup\cdots\mathcal{T}_{\text{test}}^k$.

For the evaluation metric we use the empirical risk,
\begin{equation} \label{eq:realCLmodel} 
\sum_{(\mathbf{x}_j, y_j) \in \mathcal{T}_{\text{test}}^1\cup\cdots\mathcal{T}_{\text{test}}^k} \ell \left(f \left(\x_{j} \right), {y}_{j}\right) \,,
\end{equation}
where $\ell(\cdot,\cdot)$ measures the discrepancy between prediction and ground-truth labels, and $\mathcal{T}_{\text{test}}^k$ encodes the \emph{testing} samples for task $k$. 
A well-performing RealCL model that meets the requirements of Eq.~\ref{eq:realCLmodel} exhibits discriminability across all classes, achieving a harmonious equilibrium between learning novel classes and retaining memory of existing ones.

\subsection{The unrealistic scenarios}\label{sec:unreal_scenarios}

Traditionally, in all scientific publications, the problem of continual learning has been addressed through experimental evaluations that are particularizations of the RealCL model.
These particularizations arise from imposing some restrictions on the more general RealCL model in different aspects, which we detail as follows.

Most continual learning benchmarks proposed in the literature rely on what is commonly referred to as the non-overlapping tasks paradigm. In these settings, the label space is explicitly partitioned into disjoint subsets, each assigned to a different task. Consequently, once a class has appeared in one task, it never reappears in subsequent tasks.
Unlike these formulations, RealCL imposes no constraints on the assignment of classes to tasks. Since tasks are generated directly from the incoming data stream, previously observed classes may naturally reappear in later tasks, while entirely new classes may also emerge. Therefore, the non-overlapping tasks paradigm can be regarded as a \emph{particular restriction} of the more general RealCL formulation.

Therefore, in the non-overlapping setup, the label space is partitioned into distinct, non-overlapping subsets, each corresponding to a task, and the abrupt shifts between these subsets are denoted as task boundaries.
While in the RealCL scenario there is no control over the balancing of categories or their distribution in each task, in the traditional non-overlapping tasks setting there is.
Typically, the task boundaries delineate data splits that are balanced, ensuring an approximately equal distribution of categories. 
This formulation substantially simplifies the overarching continual learning problem, constraining the previously unknown expansion of the label space and rendering it comprehensible.
Technically, the traditional non-overlapping tasks scenario can be viewed as a constrained version of RealCL, obtained by enforcing pairwise-disjoint label spaces $Y_k$ and controlled task composition.

We can further restrict the non-overlapping tasks scenario by assuming that the categories distributed among the different tasks do not overlap and gradually increase, being balanced across tasks.
This is traditionally known in the continual learning literature as the class-incremental scenario \citep{Rebuffi2017,Shmelkov2017}.
The class-incremental scenario is therefore a particular case of the RealCL setting, so that the label spaces of each task $\T^{k}$ are distributed as follows: $Y_k  \cap Y_{k^\prime} = \varnothing$ for $k\neq k^\prime$. 

Another unrealistic but popular choice for the continual learning formulation is the task-incremental setting (e.g.~\citep{lopez2017}).
In this context, in addition to the non-overlapping and class-incremental task premises, a task descriptor, or task ID, is also conveyed by an oracle both during training and testing phases.
In particular, the continual learning model observes for every task $\T^{k}=\left\{\left(\x_{i}^{k}, k, y_{i}^{k}\right)\right\}_{i=1}^{n_k}$, where $k$ is the task descriptor.
Being aware of this subset of labels a priori significantly reduces the label space during both training and inference, but it is relatively impractical to possess such knowledge in real-world scenarios.

Finally, the scenario known as domain-incremental can be defined (e.g. \citep{Goodfellow2014}).
In this setting, the label space is shared across tasks, that is: $Y_k  = Y_{k^\prime}$ for $k\neq k^\prime$.
However, data $\x_i^k$ may originate from changing input domains across tasks (e.g., varying imaging conditions, different viewpoints, permutations, etc.).

All previous formulations provided in this section are particularizations of the more general formulation provided for RealCL, an aspect that confirms that all previous CL experimental settings are simplifications of the realistic scenario proposed.

\subsubsection{A Semi-realistic scenario}

As a middle ground between the unrealistic scenarios and RealCL, we propose a semi-realistic scenario (SemiRealCL) where we enforce RealCL to be of class-incremental nature.

In the SemiRealCL setting, we ensure that once a class is introduced in a task, it does not reappear in subsequent tasks. However, the number of classes per task is not predefined, it is random, leading to an uneven distribution.
See Figure \ref{fig:abstractfig} for a better understanding of the proposed setting.
In contrast, the RealCL setting introduces no constraints on class distribution, allowing for random recurrence of previously seen classes in future tasks.
Therefore, since SemiRealCL must be a class-incremental process, once a set of classes is randomly assigned to a task, those categories cannot be used again in future tasks.
The main difference from the standard class-incremental setting is that the categories are not evenly distributed across tasks.
For example, in a SemiRealCL setting where we define 5 tasks for a dataset with 10 classes, the distribution of classes across tasks is randomized rather than being grouped in pairs. For instance, task 1 might include classes $\{0, 2, 6\}$, task 2 could have $\{5, 8\}$, task 3 might be assigned $\{1\}$, and so on.

\section{CLARE: Continual Learning Approach with pre-trained Models}\label{sec:model_clare}

To explore the new RealCL paradigm, we propose a novel solution based on pre-trained models.
We present \textbf{CLARE}, our \textbf{C}ontinual \textbf{L}earning \textbf{A}pproach with p\textbf{RE}-trained models for RealCL scenarios.
Pre-trained models inherently possess a degree of generalizability, which can be leveraged and transferred to downstream tasks. 
In this context, we introduce CLARE, a framework designed to adapt pre-trained models' recognition capabilities for the RealCL setup.

\begin{figure}[t]
\centering
\includegraphics[width=0.8\textwidth]{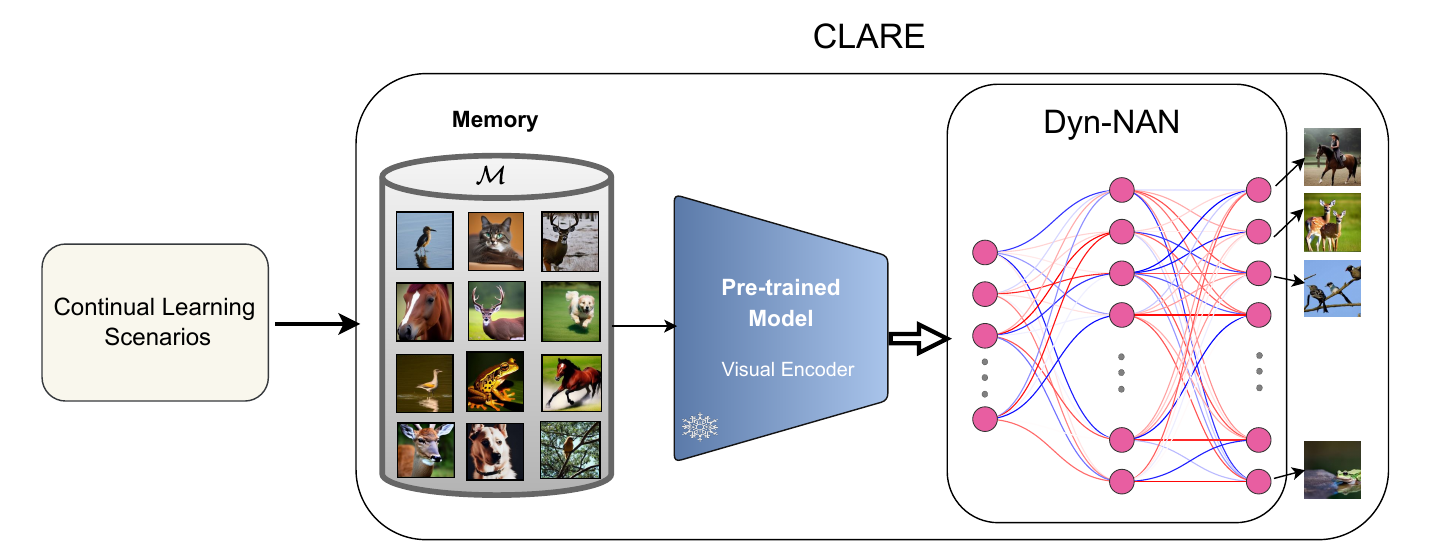}
\caption{We show the main components of CLARE: the memory buffer $\M$, the frozen pre-trained model that works as the visual encoder, and the Dyn-NAN module.}\label{fig:model_clare}
\end{figure}

We show CLARE's main components in Figure \ref{fig:model_clare}.
As can be observed, CLARE is essentially a continual learning model based on a replay buffer, where we combine:  a) a memory module $\M$ with its corresponding sample selection mechanism;
b) a pre-trained model used as an information extractor; and  c) a learning process aimed at minimizing catastrophic forgetting, which integrates a Dynamic Neural Adaptation Network (Dyn-NAN) block into the optimization process.

Note that CLARE is a model designed to operate under the proposed RealCL protocol. Although RealCL assumes that \emph{complete} historical datasets are unavailable after each task, CLARE follows the standard rehearsal-based continual learning strategy of maintaining a small bounded replay buffer $\M$ containing representative samples from previous tasks. 

The use of this bounded replay buffer should not be interpreted as contradicting the motivation behind RealCL. The objective of the proposed RealCL scenario is to remove the unrealistic assumption that complete historical datasets remain permanently available. Maintaining a small exemplar memory is a common compromise adopted in rehearsal-based continual learning, balancing performance with practical constraints on storage and computation.

Therefore, the memory module in CLARE is responsible for retaining a subset of training samples from both current and previously encountered tasks.
How does it work?
We follow a simple and efficient strategy: we randomly sample instances from the classes of the current task $\T^{k}$ and store them in the memory module $\M$, together with instances that belong to the classes considered in previous tasks $\{ \T^{k-1}, \T^{k-2}, \ldots\}$.
So, for each task $\T^{k}$, we randomly sample instances from the current task's data and add them to the memory buffer $\M$.
If the buffer has available space, the new samples are appended directly.
Otherwise, a replacement strategy is applied, where a portion of the existing samples is randomly replaced by new ones.
This process does not enforce explicit class balancing, and the resulting memory content may reflect the inherent class imbalance of the task stream.
This design choice aligns with the RealCL paradigm, where class distributions are unstructured and potentially overlapping across tasks.
Therefore, our memory module has only one parameter: the size of samples it can store.

It is noteworthy that this parameter can be dynamically changed, allowing CLARE to adapt to the difficulty of the problem dynamically as well.
In the experimental evaluation, we have demonstrated the impact that this parameter has on the system's performance.
So, the memory represents a subset of the whole data distribution, which is used by the pre-trained model encoder and the Dyn-NAN module to solve the current task $\T^{k}$ minimizing the impact over previous tasks.

The second component of CLARE is a pre-trained model used as visual encoder.
Note that CLARE is designed to be model agnostic, ensuring seamless integration with any pre-trained model, which enhances its capability to recognize and process a diverse range of classes it has been exposed to.
For our implementation and the experimental evaluation, we have opted to integrate in CLARE the Contrastive Language-Image Pre-training (CLIP) \citep{Radford2021LearningTV} pre-trained model.
CLIP learns a common embedding space for natural language and visual concepts.
CLIP employs a dual-encoder architecture that comprises two separate encoders: a Transformer model \citep{Vaswani2017AttentionIA} for text and a CNN or a ViT \citep{Dosovitskiy2020AnII} for images.
The encoders are trained on a large-scale dataset of 400 million image-caption pairs, collected from various internet domains, to maximize the similarity of text and image pairs that are semantically related and minimize the similarity of those that are not.
This enables CLIP to acquire transferable representations for both text and images, which can adapt to new data and be applied to downstream tasks such as image classification, captioning, retrieval, and generation.
For CLARE we only employ the visual encoder in CLIP, whose output is connected to the learnable Dyn-NAN block.

Finally, CLARE includes the Dyn-NAN module, which consists of fully connected layers set up with full parameterization, integrated in the optimization process in a dynamic way.
Technically, the Dyn-NAN is implemented as a dynamic fully-connected neural network.
Its design is specifically tailored to address the challenges posed by the RealCL scenario, where the number of classes is unknown and evolves unpredictably across tasks. The key feature of Dyn-NAN is its dynamic adaptability: for each new task $\mathcal{T}^k$, the output layer is dynamically reconfigured to include neurons corresponding to the newly encountered classes, while preserving the weights associated with previously learned ones.
This incremental expansion mechanism allows Dyn-NAN to maintain a unified classification head that grows with the label space, without requiring retraining of the frozen pre-trained encoder.
For the experiemntal evaluation, we employed an specific architecture that consists of three fully-connected layers with ReLU activations, and the final layer is dynamically resized to match the total number of classes observed up to task $k$. This design ensures that CLARE can flexibly accommodate new categories while minimizing interference with previously acquired knowledge.
Despite its simplicity, Dyn-NAN plays a pivotal role in enabling CLARE to perform effectively under the RealCL setting, where task boundaries and class distributions are unstructured and potentially overlapping. Its dynamic nature allows the model to adapt to evolving classification demands while preserving stability in the learned representations.
In the following section, we detail how the Dyn-NAN module is integrated into the training procedure to ensure that new categories are smoothly incorporated into the model, while minimizing the forgetting of previously acquired knowledge.

\subsection{Training procedure for CLARE}

Following the RealCL setup, for each task $\T^{k}$, CLARE first updates the memory module $\M$.
This memory buffer contains a selection of the current-task $\T^{k}$ samples together with a subset of representative samples previously stored.
For the RealCL image classification problem we are dealing with, these data come in the form of annotated image sets.
Then, the images in the memory module $\M$, denoted by $I_i^k$, are processed by the frozen visual encoder provided by the pre-trained model, resulting in a mapping from the input images to vectors $x_i^k$ by the encoder $E$, as stated by the operation $x_i^k \gets E(I_i^k)$.

In other words, to use the pre-trained model involves a mapping from the input images to vectors $\x_i^k$.
This way, for task $\T^{k}$, module Dyn-NAN receives the set $\M=\left\{\left(\x_{i}^{k}, y_{i}^{k}\right)\right\}_{i=1}^{n_k}$.
Dyn-NAN module dynamically adapts the architecture for the possible new number of classes.
Dyn-NAN works as a mapping from embeddings $\x_{i}^{k}$ to scalar scores for each category, i.e. $\{g_y: \mathbb{R}^d \to \mathbb{R}\}_{y \in \mathcal{Y}_k}$.
The output of Dyn-NAN module is defined as the concatenation of the output heads for all the classes seen so far, i.e. $\mathcal{Y}_k$, as $f_k(\x): X\rightarrow\mathcal{Y}_k$, where:
\begin{equation} \label{eq:f_k}
f_k(\x) = [g_y(\x)]_{y \in \mathcal{Y}_k} \, .
\end{equation}

CLARE model final prediction is obtained as the argmax over the output layer of Dyn-NAN module as follows:
\begin{equation} \label{eq:argmax}
\hat{y}_k(\x) = \argmax_{y \in \mathcal{Y}_k} f_k(\x)[y]\; .
\end{equation}

During training, CLARE uses the cross-entropy loss function.
The entire training process of CLARE is depicted in Algorithm \ref{algo:clare}.

\begin{algorithm}[t]
\caption{CLARE Training under the RealCL Setting}\label{algo:clare}
\begin{algorithmic}[1]
\Require Memory buffer $\mathcal{M}$ (capacity $B$), frozen encoder $E$, task sequence $\{\mathcal{T}^{1}, \dots, \mathcal{T}^{K}\}$, Dyn-NAN module $DN$
\Ensure Updated $DN$ over $\cup_{k=1}^{K}\mathcal{Y}^{k}$

\For{$k = 1$ to $K$}
    
    \State Sample $\tilde{\mathcal{T}}^{k} \subset \mathcal{T}^{k}$ and update $\mathcal{M} \leftarrow \mathcal{M} \cup \tilde{\mathcal{T}}^{k}$ \Comment{\textbf{Memory update}}
    \If{$|\mathcal{M}| > B$}
        \State Apply random replacement
    \EndIf

    \State Freeze $E$ weights.
    \For{$\I_i^k$ in $\mathcal{M}$} \Comment{\textbf{Feature extraction}}
        \State  $x_i^k \gets E(\I_i^k)$.
    \EndFor   
    
    \State Expand $DN$ output layer to $\mathcal{Y}_{k}$ \Comment{\textbf{Dynamically adapt Dyn-NAN}}     

    \For{mini-batches $\mathcal{B} \subset \mathcal{M}$}
        \State $s_i^k \gets DN(x_i^k)$. \Comment{Apply Eq. \ref{eq:f_k}}
        \State Minimize cross-entropy loss $\mathcal{L}(s_i^k, y_i^k)$. 
        \State Update $DN$ by gradient descent
    \EndFor

\EndFor
\end{algorithmic}
\end{algorithm}

Overall, in our training process, we can follow with CLARE two learning strategies.
The first one simply involves training all the model's weights from scratch in every task iteration, as the memory module always contains samples from all the classes observes so far.
Pre-trained models result effective for such training strategy.
The second possibility is to follow a fine-tuning strategy, starting from the weights that the model has from the learning done until the previous task  $\T^{k-1}$.
Due to the generalization capability of pre-trained models, it is interesting to evaluate the impact of following one strategy or the other. 
This aspect is shown in the experimental evaluation.

\section{Experiments}
\label{sec:experiments}

In this section, we present the experimental evaluation designed for the RealCL problem.
We first detail the description of the experimental methodology, highlighting the key factors that influenced our experimental design and execution. 
Our main goals are to answer the following research questions that enable us to thoroughly evaluate and comprehend the implications of the novel RealCL paradigm:
\begin{itemize}
\item[-] How do pre-trained models enhance the adaptability and performance of the system in RealCL scenarios?
\item[-] What is the efficiency of the novel CLARE approach in the RealCL setting and in its simplifications? How do the number of tasks and the memory budget affect the trade-off between effectiveness and efficiency for CLARE?
\item[-] What are the advantages of CLARE over state-of-the-art methods in terms of performance in continual learning scenarios? What are the underlying mechanisms that enable these advantages?
\end{itemize}

\subsection{Experimental setup}

\textbf{Datasets:} We have conducted experiments using four diverse datasets that pose distinct challenges in continual learning.
The first dataset, CIFAR-10 \citep{Krizhevsky2009LearningML}, consists of 60,000 color images of size  $32\time32$ pixels, distributed across 10 classes.
Each class contains 6,000 images, and the dataset is partitioned into 50,000 training images and 10,000 test images. 
The second dataset, CIFAR-100 \citep{Krizhevsky2009LearningML}, shares similarities with CIFAR-10 but encompasses 100 classes, each containing 600 images. Similar to CIFAR-10, the dataset is divided into 50,000 training images and 10,000 test images.
We also use TinyImagenet dataset \citep{Chrabaszcz2017ADV}, a subset of the extensive ImageNet \citep{Krizhevsky2012ImageNetCW} collection. It features 200 classes with 500 images each, sized at $64\time64$ pixels. The dataset is further categorized into 100,000 training images, 10,000 validation images, and 10,000 test images.
Finally, we also employ the CUB200 dataset \citep{cub200}. It consists of 11,788 images of 200 bird species. Most of the categories contain 60 images. For this dataset, we follow the same experimental setup as in previous CL work \citep{aper}.

These four are standard datasets used in the literature about continual learning.
This allows us to compare our method with state-of-the-art models and demonstrate its robustness and accuracy.
We here propose to use these datasets for the first time in the novel RealCL setting.
For all the datasets, we use the training sets for learning the models, and report the performance using the test sets.
By using four datasets with different numbers of classes, we can evaluate the performance of CLARE under various classification scenarios.

\textbf{Evaluation Metrics:} To evaluate the effectiveness of our approach, we have used various metrics that capture different relevant aspects of continual learning. 
Let $\mathcal{D} = \{\mathcal{T}^{1}, \mathcal{T}^{2}, \ldots, \mathcal{T}^{K}\}$ be the set of all tasks.
We first employ the last task accuracy $A_k$, as the average accuracy of the model on the test set of the classes that it has seen until task $\T^{k}$.
As a global measure of the performance of the models, we report also the average accuracy $A_{\text{Avg}}$ over all tasks.


We quantify the method's ability to retain knowledge from previous tasks during the process of learning new tasks. This is achieved by calculating the difference in accuracy between the previous task and the current task, a measure we refer to as Global Forgetting: $F_{\text{G}}=A_{k-1} - A_k$.
Then, the Average Global Forgetting is also reported, $F_{\text{AvgG}} = \frac{1}{K}\sum_{k=1}^{K} (A_{k-1} - A_k)$.

To measure the degree of forgetting in a continual learning model, we examine the reduction in accuracy on prior tasks after training on new tasks. The Task Forgetting metric is defined as,$F_T = A_k - A'_k$, where $A_k$ is the average accuracy on the test set of classes seen up to task $\T^{k}$ before learning a new task, and $A'_k$ is the average accuracy on the \emph{same} test set after learning the new task $\T^{k+1}$. A positive value of $F_T$ indicates forgetting, whereas a zero or negative value suggests retention or improved generalization post-learning the new task.
The Average Task Forgetting is then measured by: $F_{\text{AvgT}} = \frac{1}{K}\sum_{k=1}^{K} (A_k - A'_k)$.
For the first task $\T^{1}$, we assume the absence of measurable forgetting.

\textbf{Implementation details:} 
For our proposed method CLARE, we have integrated the CLIP pre-trained model \citep{Radford2021LearningTV}.
In the experiments, different memory sizes have been utilized for CLARE's memory module, including 2K, 4K, and 8K samples.

To populate the memory module, we follow a task-aware random sampling strategy.
For each task $\T^{k}$, a subset of samples is randomly selected from the current task's data.
These samples are added to the memory buffer, which also retains samples from previously seen tasks. When the buffer reaches its capacity, a replacement policy is applied that randomly discards existing samples to accommodate new ones. Importantly, this process does not enforce explicit class balancing, which aligns with the unstructured nature of the RealCL scenario. As a result, the memory buffer may contain an imbalanced representation of classes, depending on the random composition of each task.

The architecture of the Dyn-NAN module is initialized with three fully-connected layers with dimensions of \(512 \times 1024\), \(1024 \times 1024\), and \(1024 \times \text{number of classes}\).
Two ReLU layers are positioned subsequent to the first and second fully-connected layers, serving to augment the efficacy of network operations.

To evaluate the performance of CLARE, we report the average for the evaluation metric used on the test set over five different runs for each dataset, along with the corresponding standard deviation. 

We conduct all our experiments on a single machine with an Intel i7 8700 CPU, 64 GB RAM, and a GeForce RTX 2080 Ti GPU. 
For the optimization, we use stochastic gradient descent (SGD) with a batch size of 64 and a weight decay of 1e-4. 
We also employ stochastic gradient descent with restarts (SGDR) \citep{Loshchilov2016SGDRSG} as a learning rate scheduler, with the following parameters: the initial cycle length ~$T_0 = 1$, the cycle length multiplier $T_{\text{mult}} = 2$, a warm start of one epoch, a maximum learning rate of 0.005, and a minimum learning rate of 5e-5. 
Moreover, we apply cutmix \citep{Yun2019CutMixRS}, a data augmentation technique that randomly mixes two images and their labels, with a probability of 0.5 and a regularization strength of 1.0. 
We apply cutmix to all datasets and RealCL scenarios, regardless of the number of classes or the size of the images.

\subsection{Results}

\subsubsection{CLARE's Performance in all RealCL scenarios}

This section starts reporting the results obtained by applying CLARE to three different continual learning scenarios proposed in Section \ref{sec:realistic}: RealCL, SemiRealCL and unrealistic. 
We then conduct a detailed analysis and comparison of the accuracy and forgetting rates for each scenario, providing a thorough evaluation of the performance of CLARE across three of the proposed datasets with varying class numbers and memory sizes.

\begin{figure}[h!]
  \centering
  \begin{subfigure}{0.33\textwidth}
    \centering
    \includegraphics[width=\textwidth]{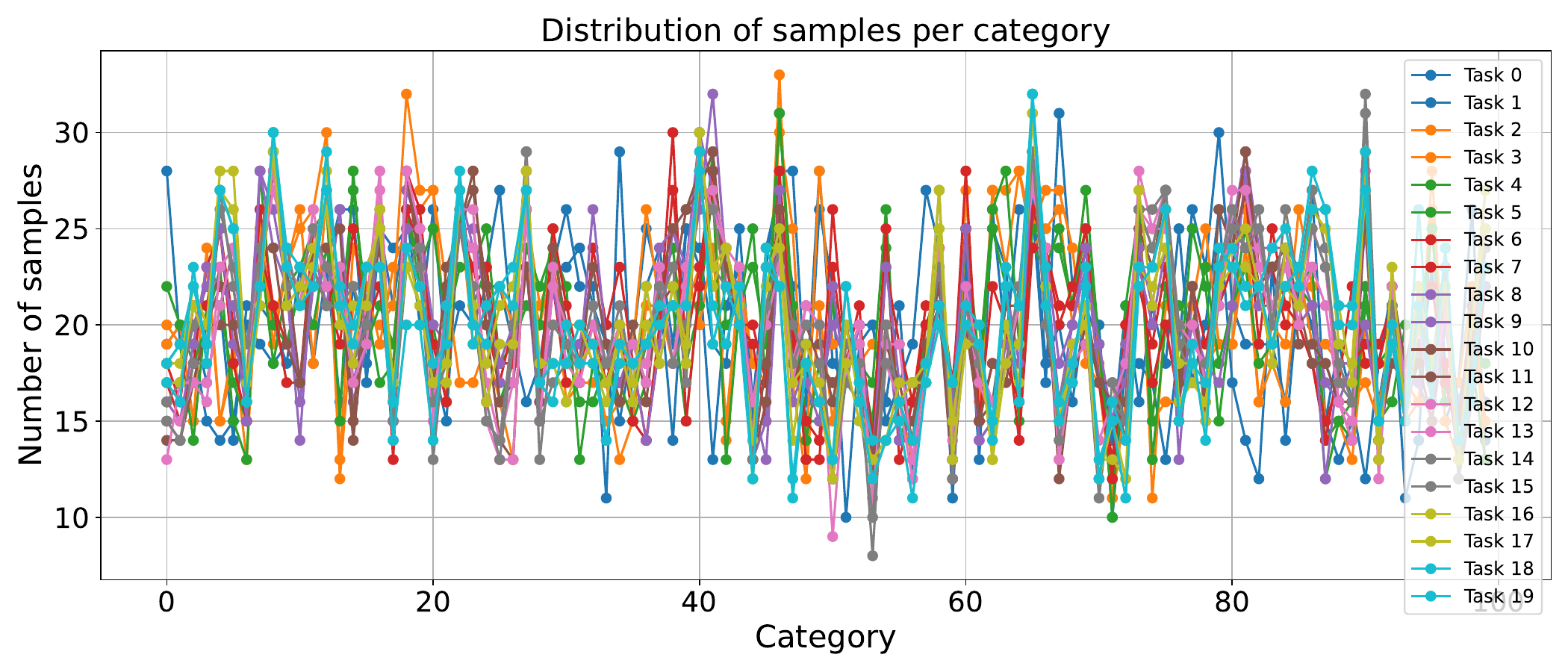}
    \caption{RealCL.}
  \end{subfigure}%
  \begin{subfigure}{0.33\textwidth}
    \centering
    \includegraphics[width=\textwidth]{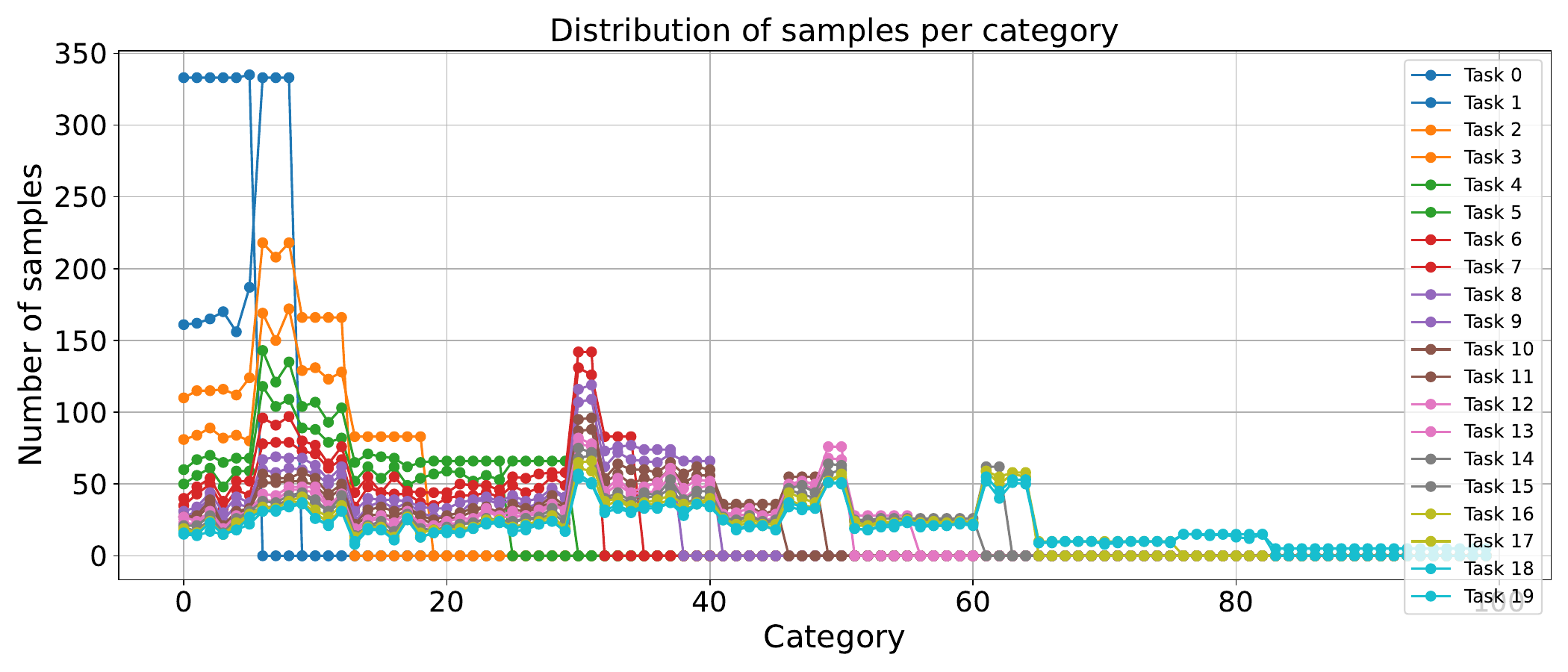}
    \caption{SemiRealCL.}
  \end{subfigure}%
  \begin{subfigure}{0.33\textwidth}
    \centering
    \includegraphics[width=\textwidth]{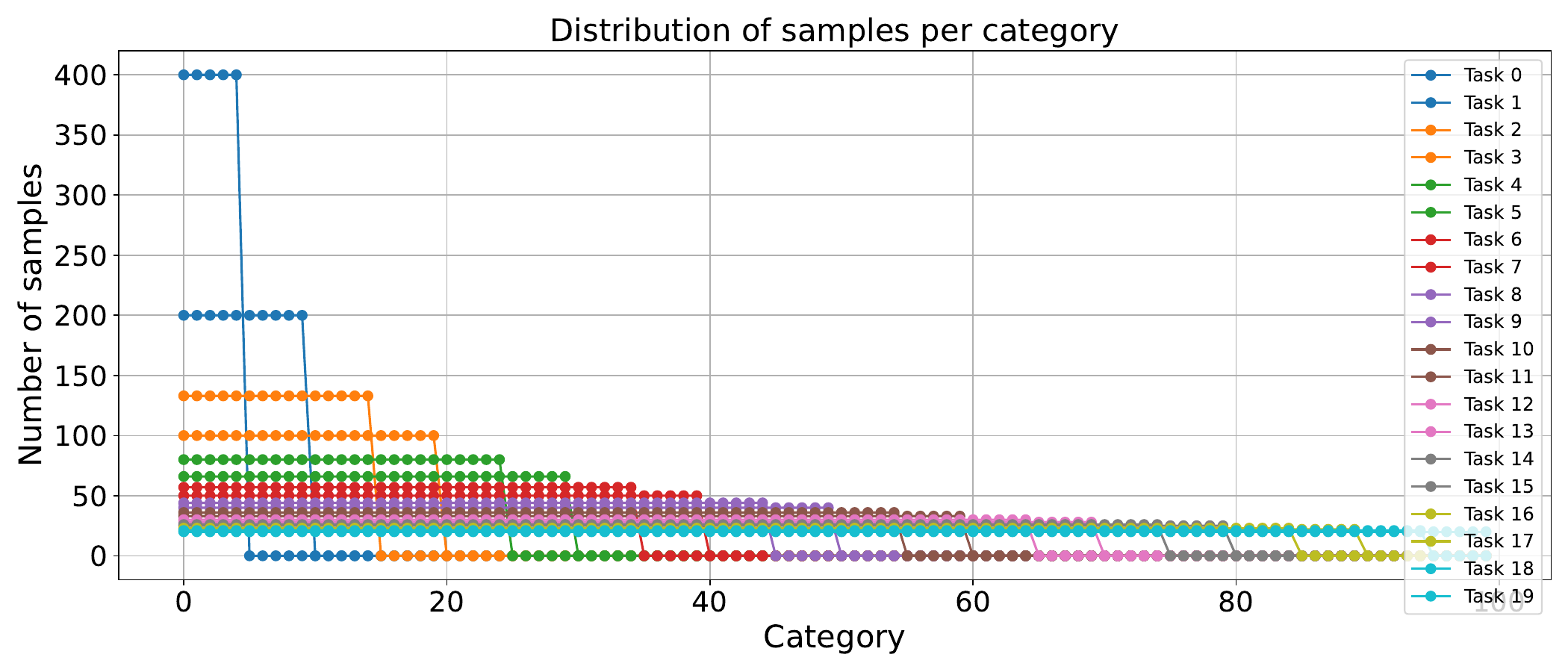}
    \caption{Unrealistic.}
  \end{subfigure}
  \caption{Class distributions for all the RealCL scenarios when CLARE model is used with the CIFAR100 dataset for 20 tasks. Note that the classes have been arranged in ascending order to facilitate the visualization of the figure.}
   \label{fig:class_distributions}
\end{figure}

To better understand the implications of the proposed scenarios and the results reported in each of them, we analyze the class distributions across tasks in each setup. In the unrealistic scenario, classes are evenly distributed and introduced incrementally, resulting in balanced and non-overlapping tasks. SemiRealCL relaxes this constraint by allowing a variable number of classes per task, though without recurrence. RealCL, in contrast, introduces tasks with randomly sampled classes, allowing both imbalance and recurrence, which better reflects real-world data streams.
Figure~\ref{fig:class_distributions} illustrates these differences by showing the number of samples per class across 20 tasks when we use CLARE with the CIFAR100 dataset, for example.
This visualization highlights the structural variability and unpredictability of RealCL, which poses a greater challenge for continual learning models. The lack of control over class composition and recurrence in RealCL demands adaptive mechanisms capable of handling evolving and unstructured label spaces—precisely the motivation behind our proposal.

Tables \ref{tablecifar10_results}, \ref{tablecifar100_results}, and \ref{tabletiny_results} report the results of CLARE for datasets CIFAR-10, CIFAR-100 and TinyImagenet, respectively.
By analyzing all their results, we first observe that CLARE achieves the best results under the Unrealistic setup, but only in experiments conducted on the CIFAR-100 and TinyImageNet datasets. When considering the CIFAR-10 dataset, the performance between the Unrealistic and the rest of setups is nearly equivalent.
In the Unrealistic scenario, all classes are equally distributed across tasks and the memory buffer has a balanced representation of each class.
Ensuring that the model has no class imbalance prevents the problem of poor performance of the model on minority classes. Interestingly, as the memory capacity increases, the model benefits from observing more samples from each class while training, which can result in higher accuracy. 
This proves that a greater number of examples available for generalization contribute to a more profound understanding of data by the model.
However, it’s important to recognize that expanding memory capacity comes with a rise in computational demands. This necessitates a considered compromise to balance the expanded memory with the overall efficiency of the model.
We argue that this unrealistic scenario does not represent the best environment to validate continual learning models. The simplifications it assumes are artificial and benefit models that can leverage the potential of having well-distributed and balanced classes in memory, or that have been specifically designed for the situation where classes appear incrementally and in a controlled number.
Therefore, the RealCL and SemiRealCL scenarios emerge as better candidates for evaluating the continual learning capabilities that the proposed models actually possess for this purpose.

Second, we find that in the RealCL scenario, for all datasets, CLARE reports the lowest global and task forgetting.
Remember that in the RealCL scenario one does not control the sample distribution across tasks.
CLARE employs a random selection process for samples stored in memory, which may result in an imbalance in the memory, characterized by varying sample counts for each class.
Despite all this, CLARE is capable of offering the lowest forgetting rates, which reinforces its suitability for these more challenging scenarios.
This demonstrates the effectiveness and robustness of CLARE.
Furthermore, we observe that accuracy and memory capacity are positively correlated, indicating that the model advantages from a larger memory module. This shows how memory size affects CLARE performance and how using additional memory resources may lead to higher accuracy.

In the SemiRealCL setting, where the classes are randomly assigned to different tasks and have unequal distributions within each task, the memory buffer exhibits this imbalance in the way it preserves the representations of the classes. 
This leads to a problem where the model experiences the highest degree of forgetting, both in terms of task forgetting and global forgetting. 
Consequently, the last task accuracy, which reflects the model’s performance after learning all the tasks, is the lowest in this scenario compared to the others, because the model is learning with a severely imbalanced memory buffer. 
Class imbalance in the tasks is another element that makes a difference to the average accuracy of the model because some tasks may have more classes to learn while others are simpler. However, from the tables \ref{tablecifar100_results} and \ref{tabletiny_results}, it can be seen that if the model has a larger memory budget, then its accuracy will also improve as it can hold more representations of each class and learn better.

Overall, to quantify the impact of memory size, we have conducted an in-depth analysis across all continual learning scenarios. 
As expected for rehearsal-based continual learning methods, increasing the replay memory consistently improves performance by preserving a more representative subset of previously observed samples. Nevertheless, the objective of these experiments is not to demonstrate that larger memories alone improve continual learning, but rather to analyze the robustness of CLARE under different memory budgets. 
The observed improvements therefore reflect the complementary effect of the replay mechanism together with the proposed representation learning strategy.
While it is evident that larger memory sizes improve accuracy due to enhanced class representation, our results demonstrate that they also reduce both task and global forgetting rates. This relationship is particularly crucial in realistic scenarios, where the lack of structured task organization challenges the model’s ability to retain and integrate knowledge effectively. These findings underscore the importance of optimizing memory usage to balance performance and computational efficiency.

\begin{table}[h!]
\centering
\sisetup{table-align-text-post=false}
\begin{tabular}{l|c|c|c}
\toprule
\multicolumn{1}{c|}{CIFAR - 10} & \multicolumn{1}{c|}{Unrealistic} & \multicolumn{1}{c|}{SemiRealCL} & \multicolumn{1}{c}{RealCL} \\ \midrule
\multicolumn{1}{c|}{Memory Size} & M=1K & M=1K & M=1K \\ \midrule
Last Task Accuracy & 89.39 {\small $\pm0.05$}  & 89.38 {\small $\pm0.24$} & \hspace{1mm}89.90 {\small $\pm0.18$}  \\
Average Accuracy & 94.45 {\small $\pm0.02$}  & 94.05 {\small $\pm0.64$}  & 89.86 {\small $\pm0.06$} \\
Average Global Forgetting & 2.04 {\small $\pm0.01$} & 2.00 {\small $\pm0.06$} & 0.01 {\small $\pm0.06$} \\
Average Task Forgetting & 1.49 {\small $\pm0.02$} & 1.83 {\small $\pm0.51$} & 0.01 {\small $\pm0.06$} \\
\bottomrule
\end{tabular}
\caption{CLARE's performance on CIFAR-10 Dataset with Memory Module size of 1K  (Values in Percentage).}
\label{tablecifar10_results}
\end{table}

\begin{table}[h!]
\centering
\scalebox{0.55}{ 
\begin{tabular}{l|l|l|l}
\toprule
\multicolumn{1}{c|}{CIFAR - 100} & \multicolumn{1}{c|}{Unrealistic} & \multicolumn{1}{c|}{SemiRealCL} & \multicolumn{1}{c}{RealCL} \\ \midrule
\multicolumn{1}{c|}{Memory Size} &  M=2K  \hspace{12mm}  M=4K  \hspace{12mm} M=8K &  M=2K  \hspace{12mm} M=4K \hspace{12mm} M=8K&  M=2K \hspace{12mm} M=4K \hspace{12mm} M=8K\\ \midrule

Last Task Accuracy & 63.95 {\small $\pm$0.07}  \hspace{2.1mm} 69.22 {\small $\pm$0.06} \hspace{1.5mm} 72.98 {\small $\pm$0.15}  & 56.42 {\small $\pm$3.24}  \hspace{2.7mm} 63.11 {\small $\pm$2.44} \hspace{2.7mm} 68.76 {\small $\pm$1.80} & 63.11 {\small $\pm$0.50}  \hspace{2mm} 68.76 {\small $\pm$0.11} \hspace{2mm} \hspace{1mm} 72.26 {\small $\pm$0.27} \\

Average Accuracy & 77.81 {\small $\pm$0.01} \hspace{2.1mm} 81.07 {\small $\pm$0.03} \hspace{1.5mm}   83.42{\small $\pm$0.01}&  79.61 {\small $\pm$1.68}\hspace{3.5mm} 82.35 {\small $\pm$1.52} \hspace{3mm} 84.77 {\small $\pm$1.41} &  63.19 {\small $\pm$0.27}   \hspace{1.5mm} 68.53 {\small $\pm$0.10}  \hspace{4.5mm} 71.88 {\small $\pm$0.15}   \\

Average Global Forgetting & 1.79 {\small $\pm$0.00} \hspace{4mm} 1.53 {\small $\pm$0.00} \hspace{4mm} 1.34 {\small $\pm$0.01} &  2.04 {\small $\pm$0.17}  \hspace{4mm} 1.71 {\small $\pm$0.12} \hspace{5mm} 1.43 {\small $\pm$0.10}  & 0.00 {\small $\pm$0.03}       \hspace{3mm}  -0.17 {\small $\pm$0.01} \hspace{5.5mm} -0.34 {\small $\pm$0.01}\\

Average Task Forgetting & 1.70 {\small $\pm$0.04}   \hspace{4mm} 1.42 {\small $\pm$0.03}  \hspace{4mm} 1.24 {\small $\pm$0.02} & 2.28 {\small $\pm$0.31}\hspace{5mm} 1.99 {\small $\pm$0.23} \hspace{5mm} 1.74 {\small $\pm$0.23} & 0.00 {\small $\pm$0.03} \hspace{3mm}  -0.17 {\small $\pm$0.02}\hspace{6.5mm} -0.34 {\small $\pm$0.01}\\
\bottomrule

\end{tabular}
}
\caption{CLARE's performance on CIFAR-100 Dataset with different Memory Module sizes (2K, 4K and 8K).}
\label{tablecifar100_results}
\end{table}


\begin{table}[h!]
\centering
\scalebox{0.55}{ 
\begin{tabular}{l|l|l|l}
\toprule
\multicolumn{1}{c|}{TinyImageNet} & \multicolumn{1}{c|}{Unrealistic} & \multicolumn{1}{c|}{SemiRealCL} & \multicolumn{1}{c}{RealCL} \\ \midrule
\multicolumn{1}{c|}{Memory Size} &  M=2K  \hspace{12mm}  M=4K  \hspace{12mm} M=8K &  M=2K  \hspace{12mm} M=4K \hspace{12mm} M=8K&  M=2K \hspace{12mm} M=4K \hspace{12mm} M=8K\\ \midrule

Last Task Accuracy & 56.40 {\small $\pm$0.13}  \hspace{2.1mm} 62.55 {\small $\pm$0.08} \hspace{1.5mm} 66.86 {\small $\pm$0.07}  & 47.71 {\small $\pm$3.40}  \hspace{2.7mm} 56.26 {\small $\pm$3.20} \hspace{2.7mm} 61.94 {\small $\pm$2.73} & 54.95 {\small $\pm$0.24}  \hspace{2mm} 61.61 {\small $\pm$0.09} \hspace{2mm} \hspace{1mm} 66.19 {\small $\pm$0.29} \\

Average Accuracy & 71.94 {\small $\pm$0.02} \hspace{2.1mm} 75.72 {\small $\pm$0.02} \hspace{1.5mm}   78.39 {\small $\pm$0.03}&  74.21 {\small $\pm$2.19}\hspace{3.5mm} 77.78 {\small $\pm$1.70} \hspace{3mm} 80.58 {\small $\pm$1.61} &  54.84 {\small $\pm$0.17}   \hspace{2mm} 61.68 {\small $\pm$0.23}  \hspace{4.5mm} 66.13 {\small $\pm$0.15}   \\

Average Global Forgetting & 1.93 {\small $\pm$0.01} \hspace{4mm} 1.69 {\small $\pm$0.01} \hspace{4mm} 1.47 {\small $\pm$0.01} &  2.49 {\small $\pm$0.24}  \hspace{4mm} 2.07 {\small $\pm$0.23} \hspace{5mm} 1.78 {\small $\pm$0.20}  & 0.00 {\small $\pm$0.02} \hspace{4mm}  0.02 {\small $\pm$0.01} \hspace{6mm}  -0.11 {\small $\pm$0.03}\\

Average Task Forgetting & 1.97 {\small $\pm$0.02}   \hspace{4mm} 1.67 {\small $\pm$0.04}  \hspace{4mm} 1.45 {\small $\pm$0.02} & 2.64 {\small $\pm$0.41}\hspace{5mm} 2.24 {\small $\pm$0.33} \hspace{5mm} 1.98 {\small $\pm$0.31} & 0.00 {\small $\pm$0.02} \hspace{4mm}  0.02 {\small $\pm$0.01}\hspace{7mm} -0.11 {\small $\pm$0.03}\\
\bottomrule

\end{tabular}
}
\caption{CLARE's performance on TinyImageNet Dataset with different Memory Module sizes (2K, 4K and 8K).}
\label{tabletiny_results}
\end{table}

\begin{figure} [h!]
  \centering
   \begin{subfigure}{0.33\textwidth}
    \centering
    \includegraphics[width=\textwidth]{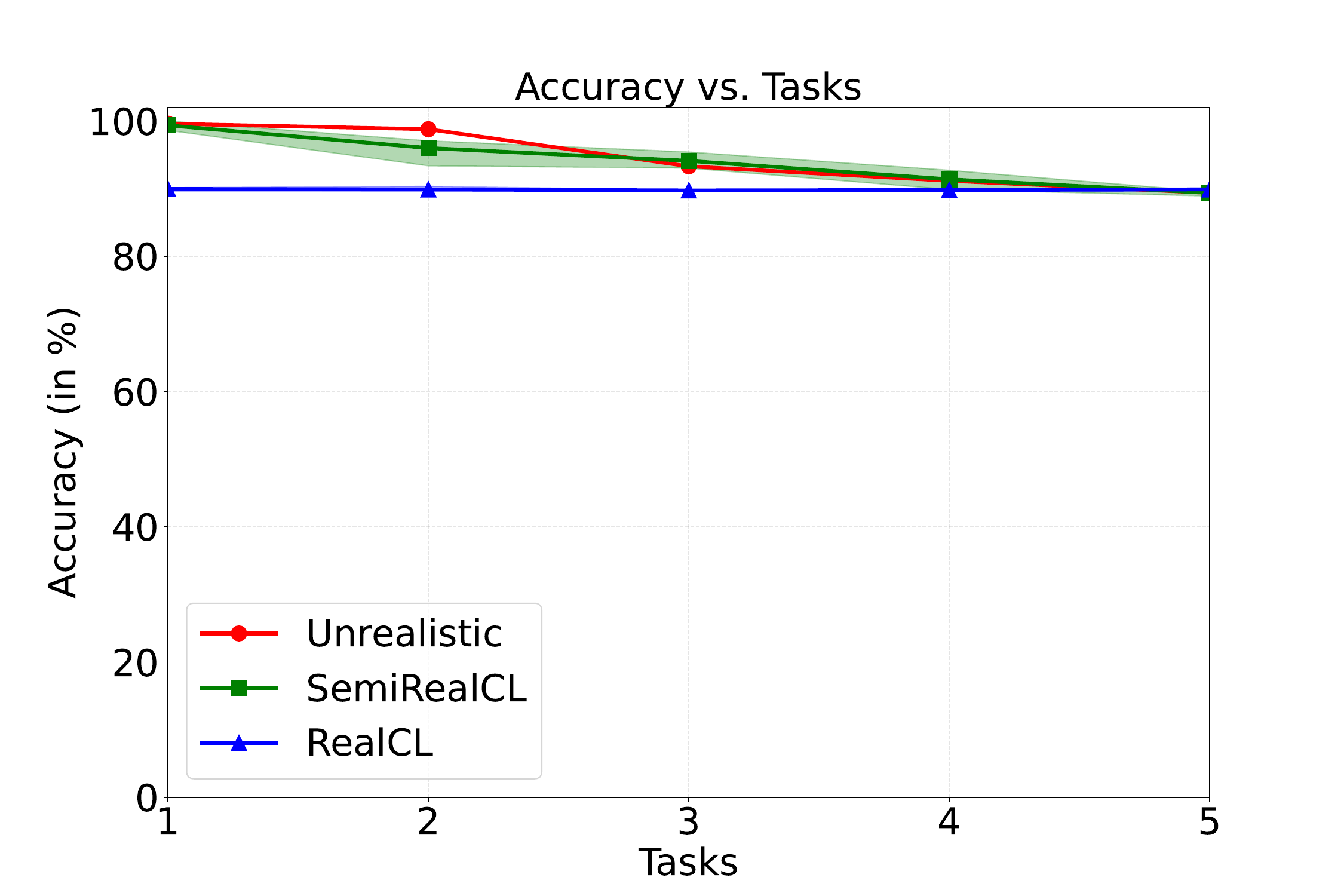}
    \caption{\small Task accuracy.}
  \end{subfigure}%
  \hfill
  \begin{subfigure}{0.33\textwidth}
    \centering
    \includegraphics[width=\textwidth]{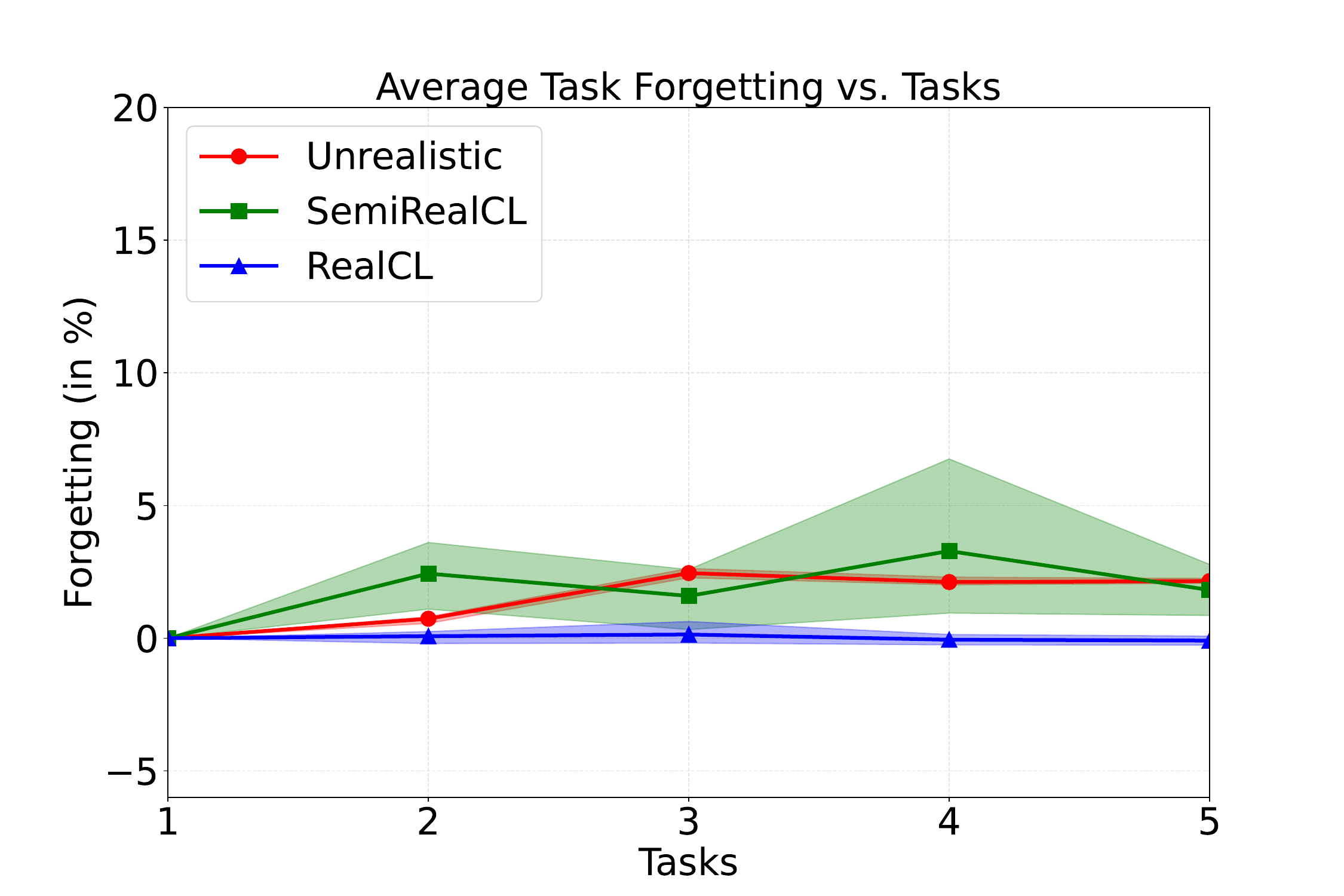}
    \caption{ \small Average task forgetting.}
  \end{subfigure}%
  \hfill
  \begin{subfigure}{0.33\textwidth}
    \centering
    \includegraphics[width=\textwidth]{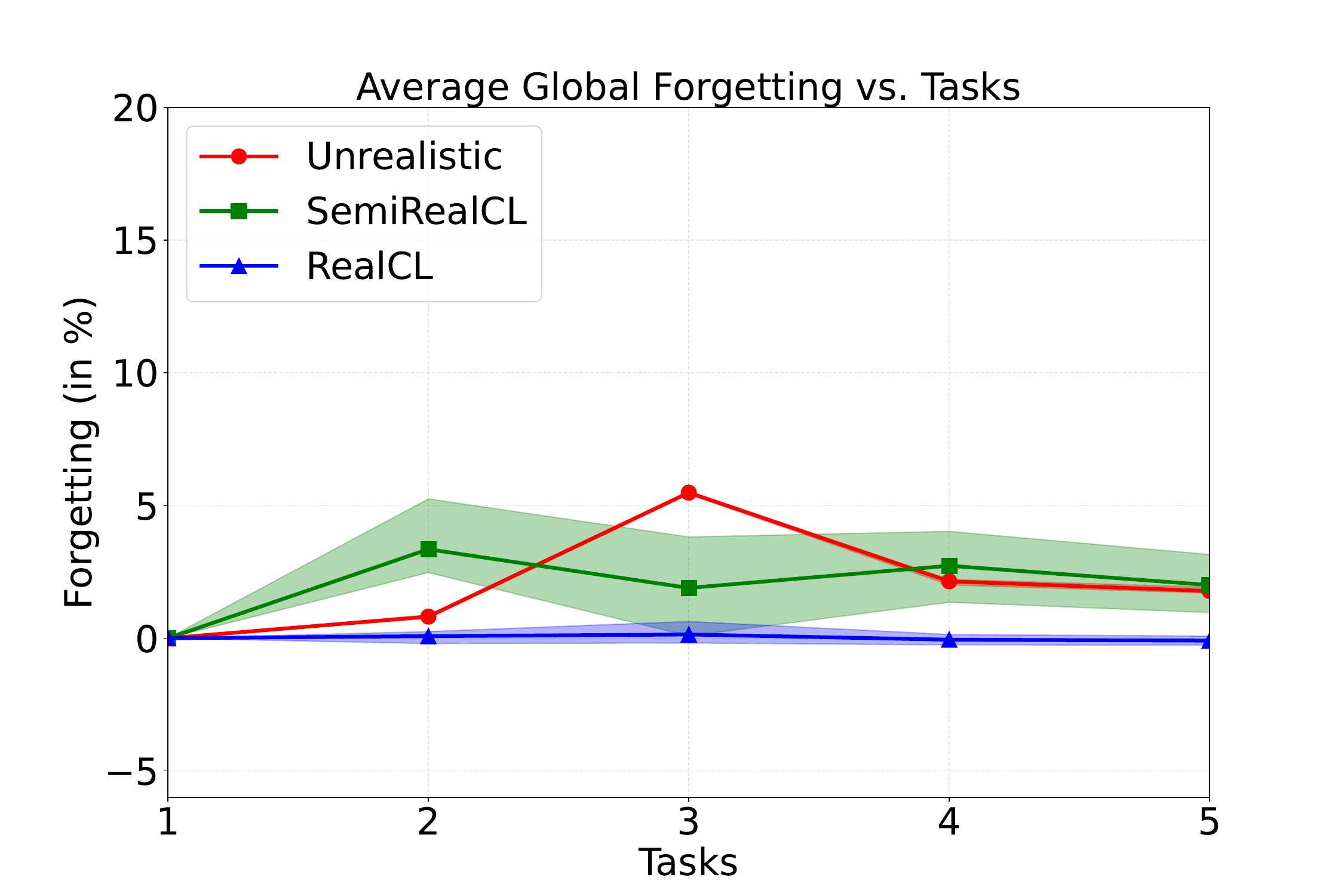}
    \caption{\small Average global forgetting.}
  \end{subfigure}%
  \caption{Results of CLARE for CIFAR-10 in the three scenarios with memory module size of 1K.}
   \label{cifar_results_1000}
\end{figure}

\begin{figure}[h!]
  \centering
  \begin{subfigure}{0.33\textwidth}
    \centering
    \includegraphics[width=\textwidth]{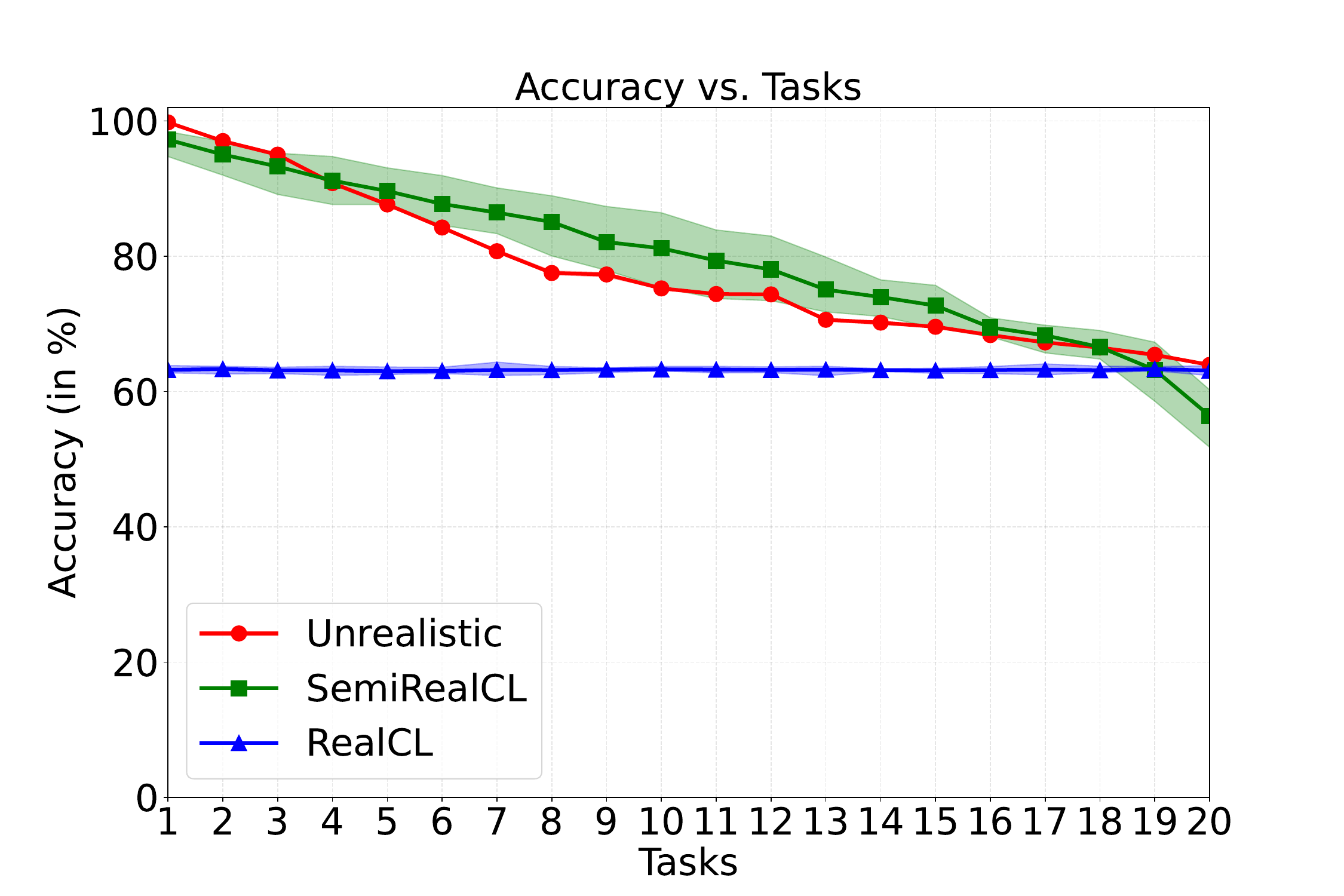}
    \caption{\small Task accuracy.}
  \end{subfigure}%
  \begin{subfigure}{0.33\textwidth}
    \centering
    \includegraphics[width=\textwidth]{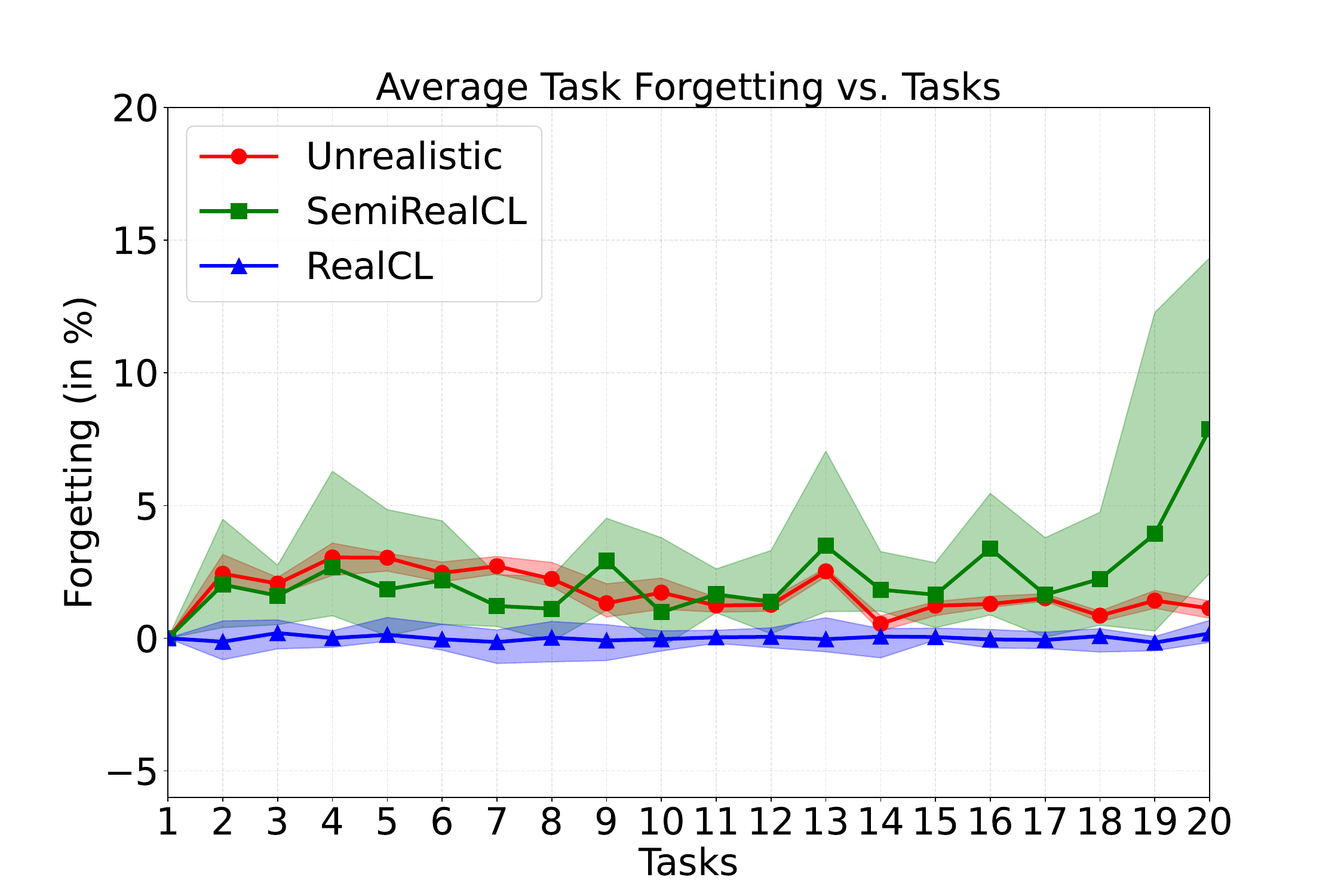}
    \caption{\small Average task forgetting.}
  \end{subfigure}%
  \begin{subfigure}{0.33\textwidth}
    \centering
    \includegraphics[width=\textwidth]{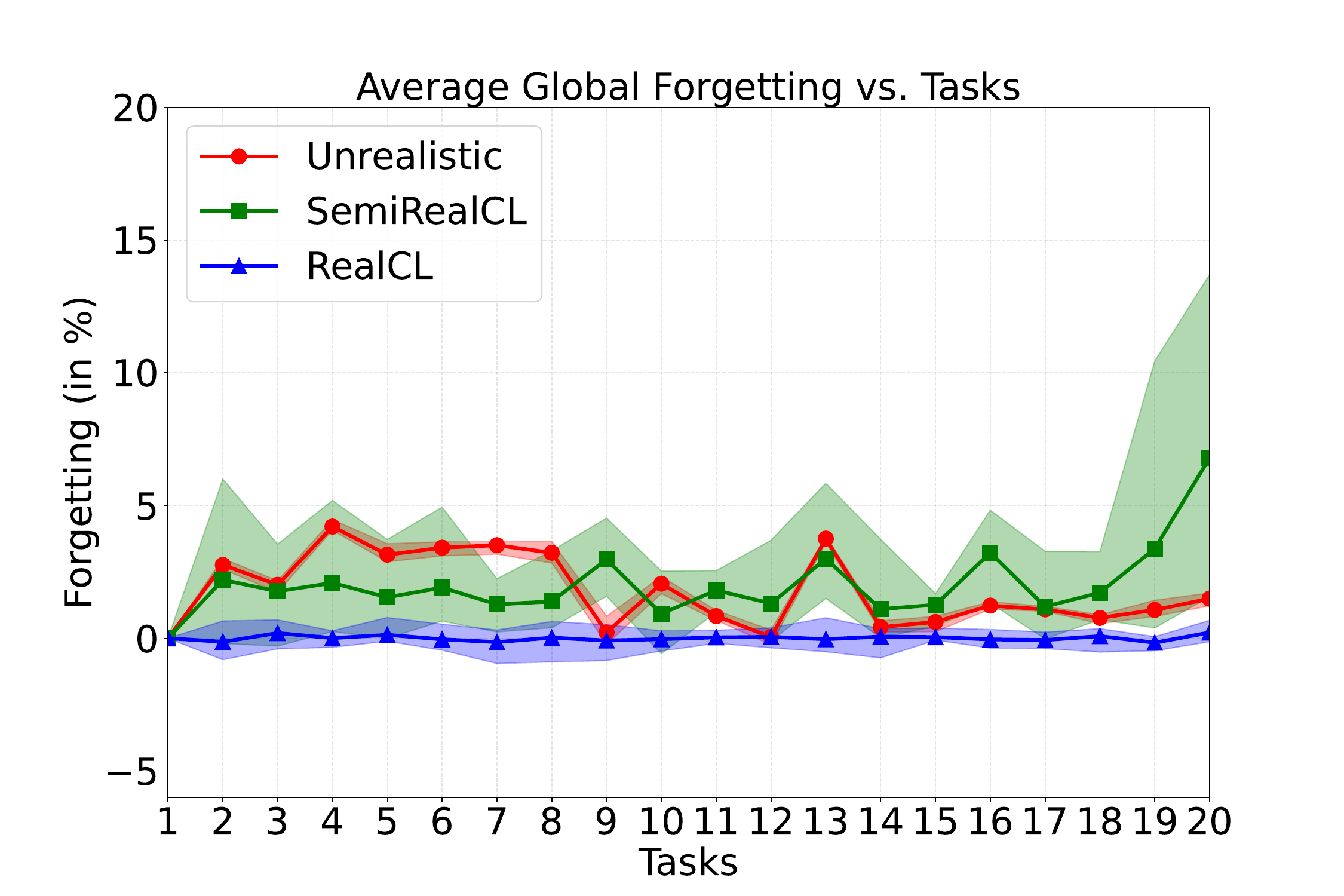}
    \caption{\small Average global forgetting.}
  \end{subfigure}
  \caption{Results of CLARE for CIFAR-100 in the three scenarios with memory module size of 2K.}
  \label{fig:cifar100_results_2000}
\end{figure}

\begin{figure} [h!]
  \centering
  \begin{subfigure}{0.33\textwidth}
    \centering
    \includegraphics[width=\textwidth]{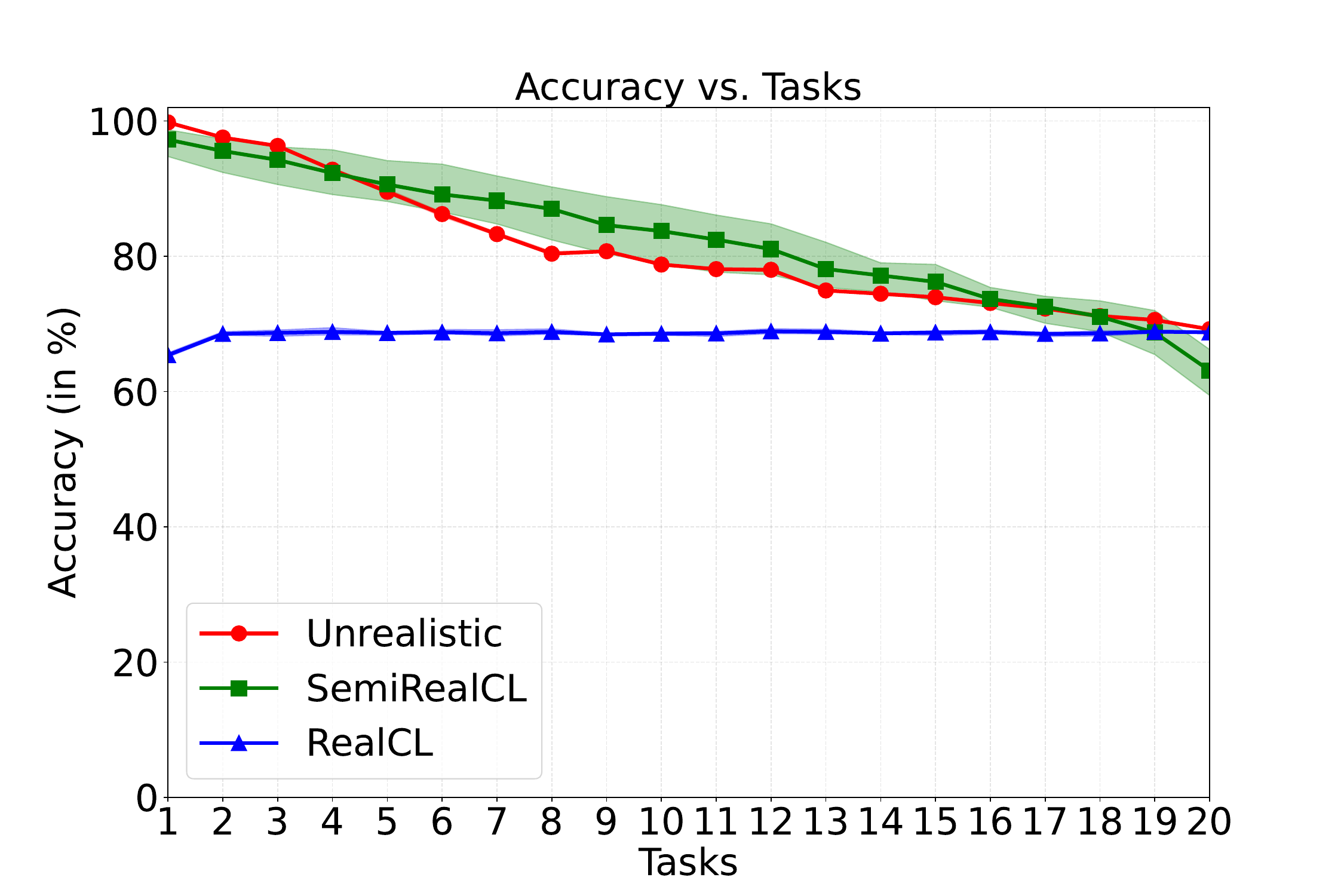}
    \caption{\small Task accuracy.}
  \end{subfigure}%
  \hfill
    \begin{subfigure}{0.33\textwidth}
    \centering
    \includegraphics[width=\textwidth]{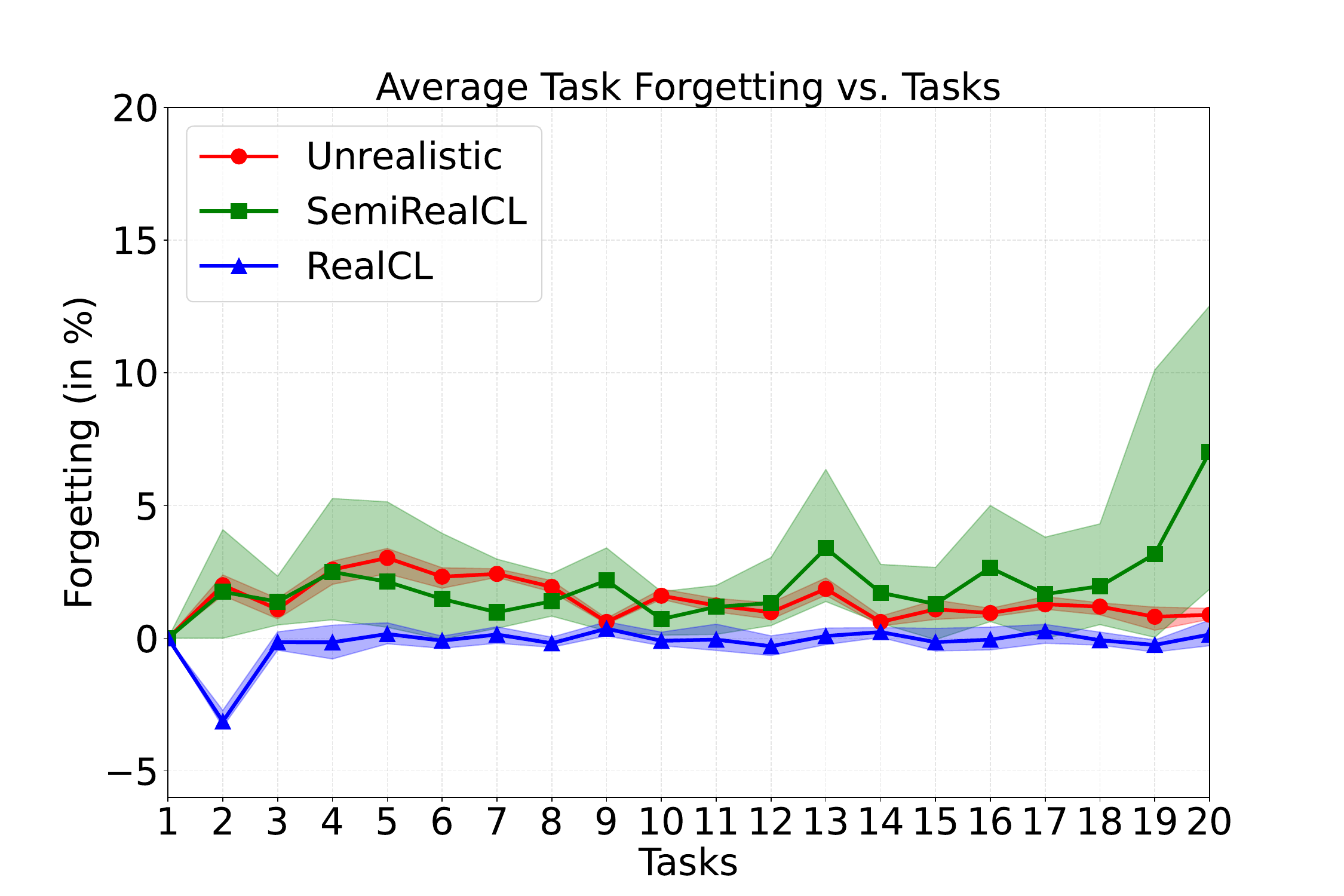}
    \caption{ \small Average task forgetting.}
  \end{subfigure}%
  \hfill
  \begin{subfigure}{0.33\textwidth}
    \centering
    \includegraphics[width=\textwidth]{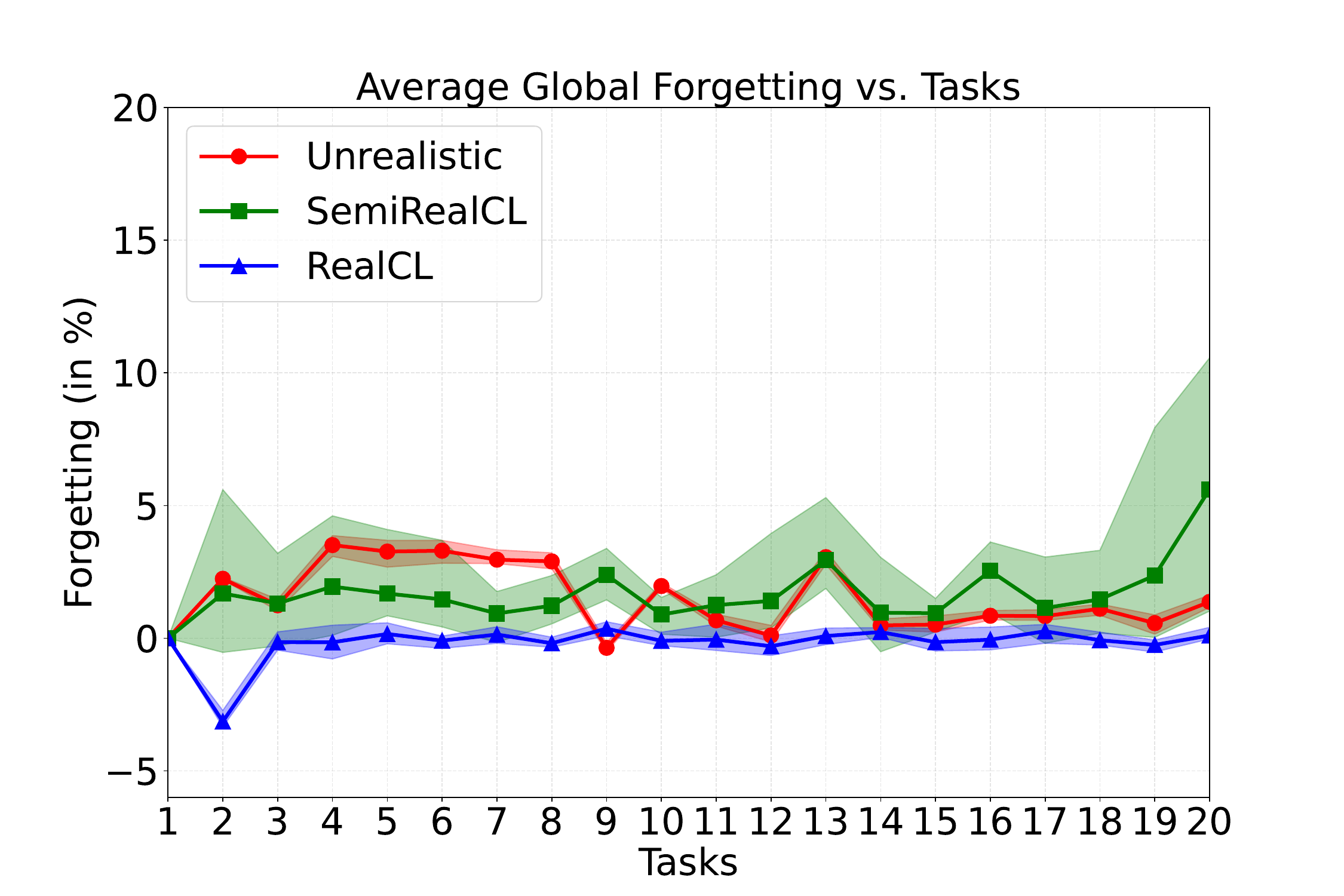}
    \caption{\small Average global forgetting.}
  \end{subfigure}%
  
  \caption{Results of CLARE for CIFAR-100 in the three scenarios with memory module size of 4K.}
  \label{fig:cifar100_results_4000}
\end{figure}

\begin{figure} [h!]
  \centering
   \begin{subfigure}{0.33\textwidth}
    \centering
    \includegraphics[width=\textwidth]{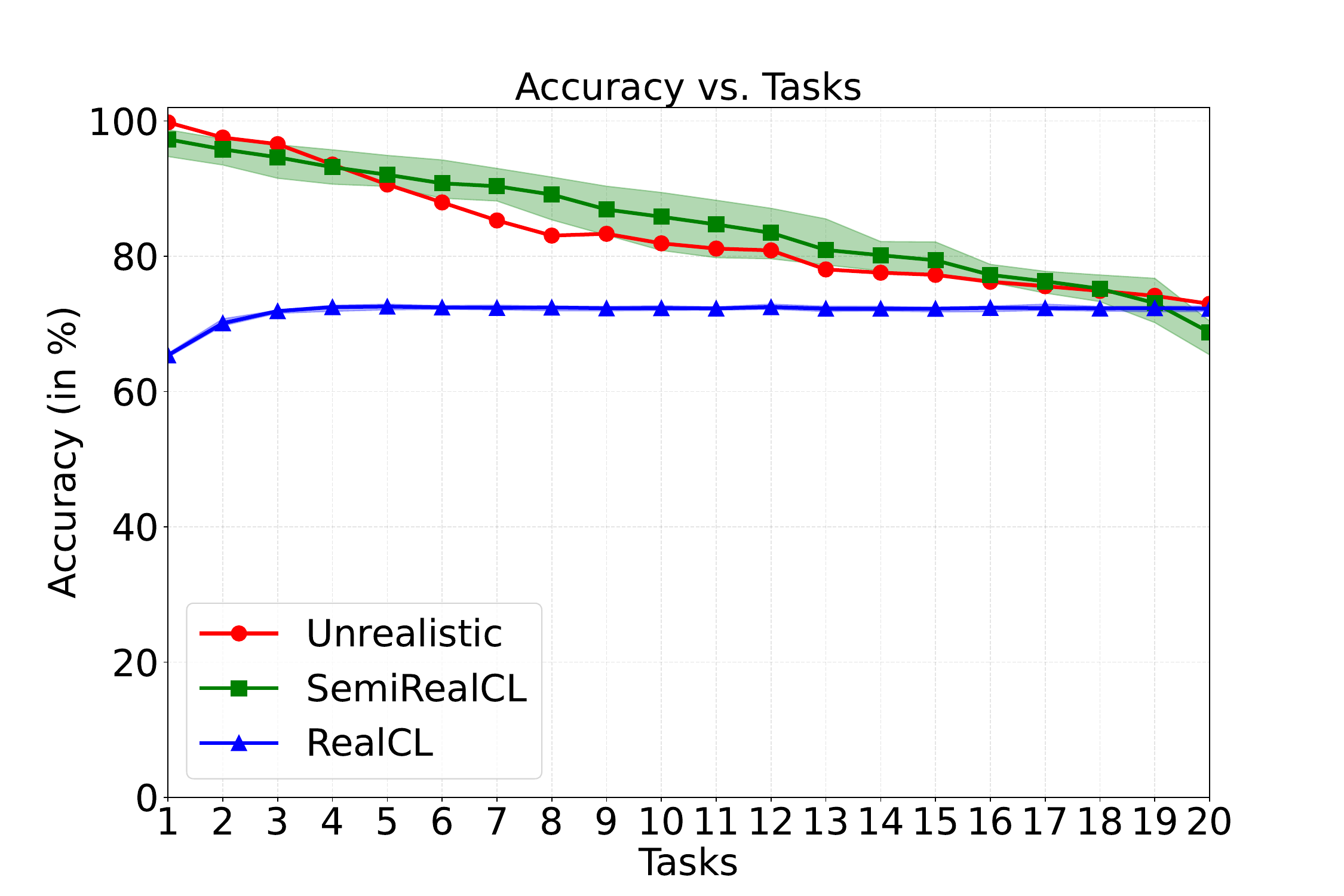}
    \caption{\small Task accuracy.}
  \end{subfigure}%
  \hfill
  \begin{subfigure}{0.33\textwidth}
    \centering
    \includegraphics[width=\textwidth]{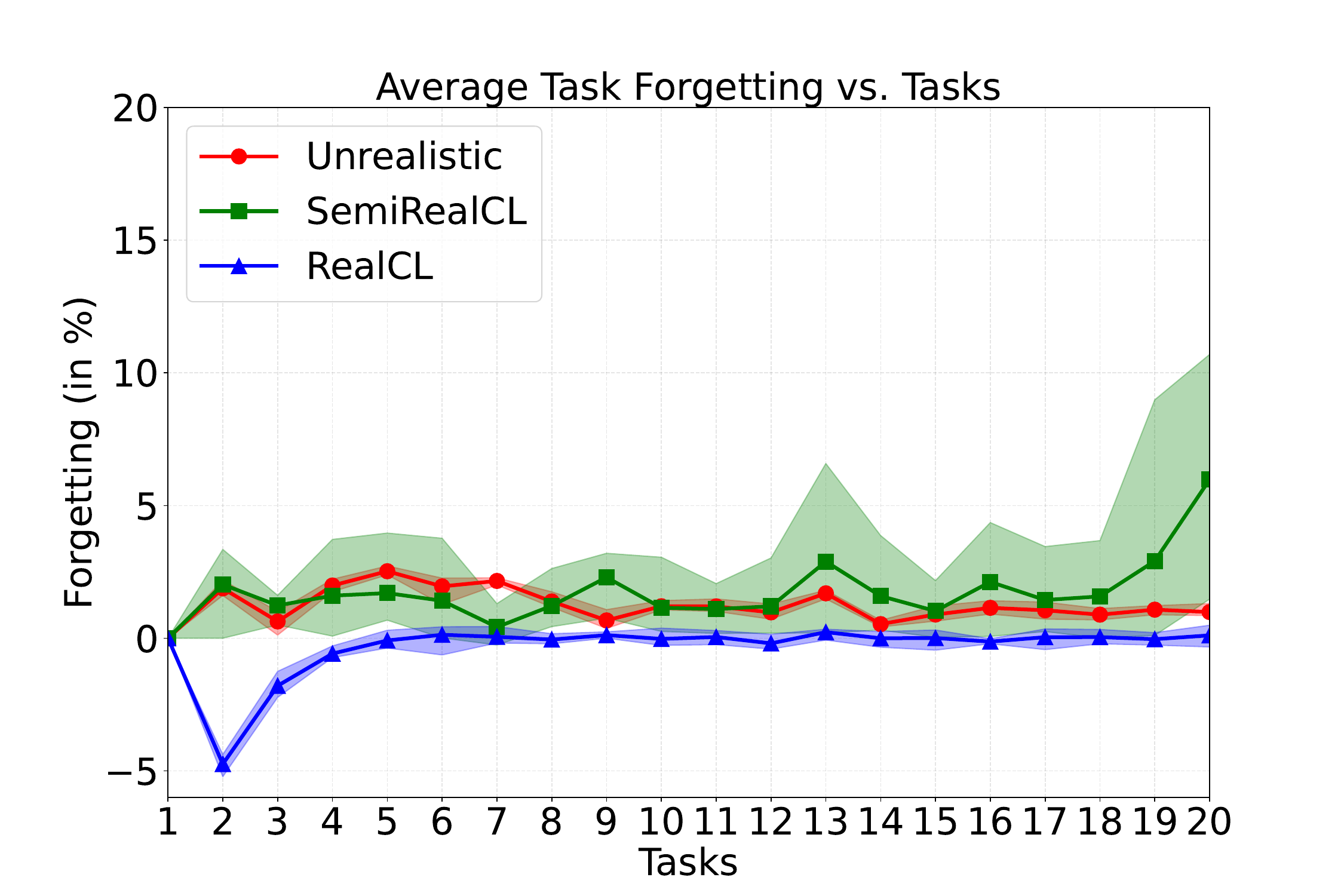}
    \caption{ \small Average task forgetting.}
  \end{subfigure}%
  \hfill
  \begin{subfigure}{0.33\textwidth}
    \centering
    \includegraphics[width=\textwidth]{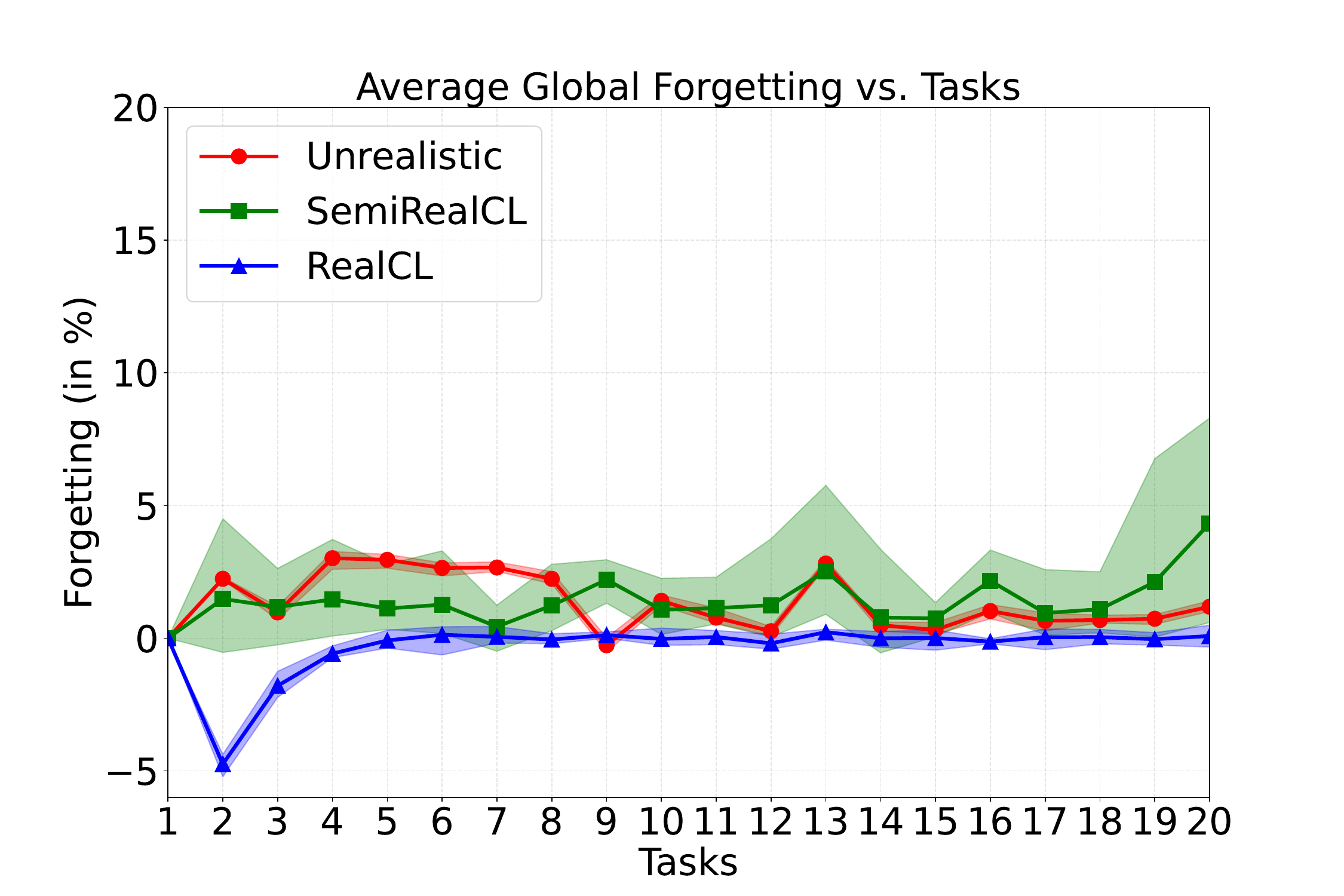}
    \caption{\small Average global forgetting.}
  \end{subfigure}
  \caption{Results of CLARE for CIFAR-100 in the three scenarios with memory module size of 8K.}
    \label{fig:cifar100_results_8000}
\end{figure}

\begin{figure}[h!]
  \centering
  \begin{subfigure}{0.33\textwidth}
    \centering
    \includegraphics[width=\textwidth]{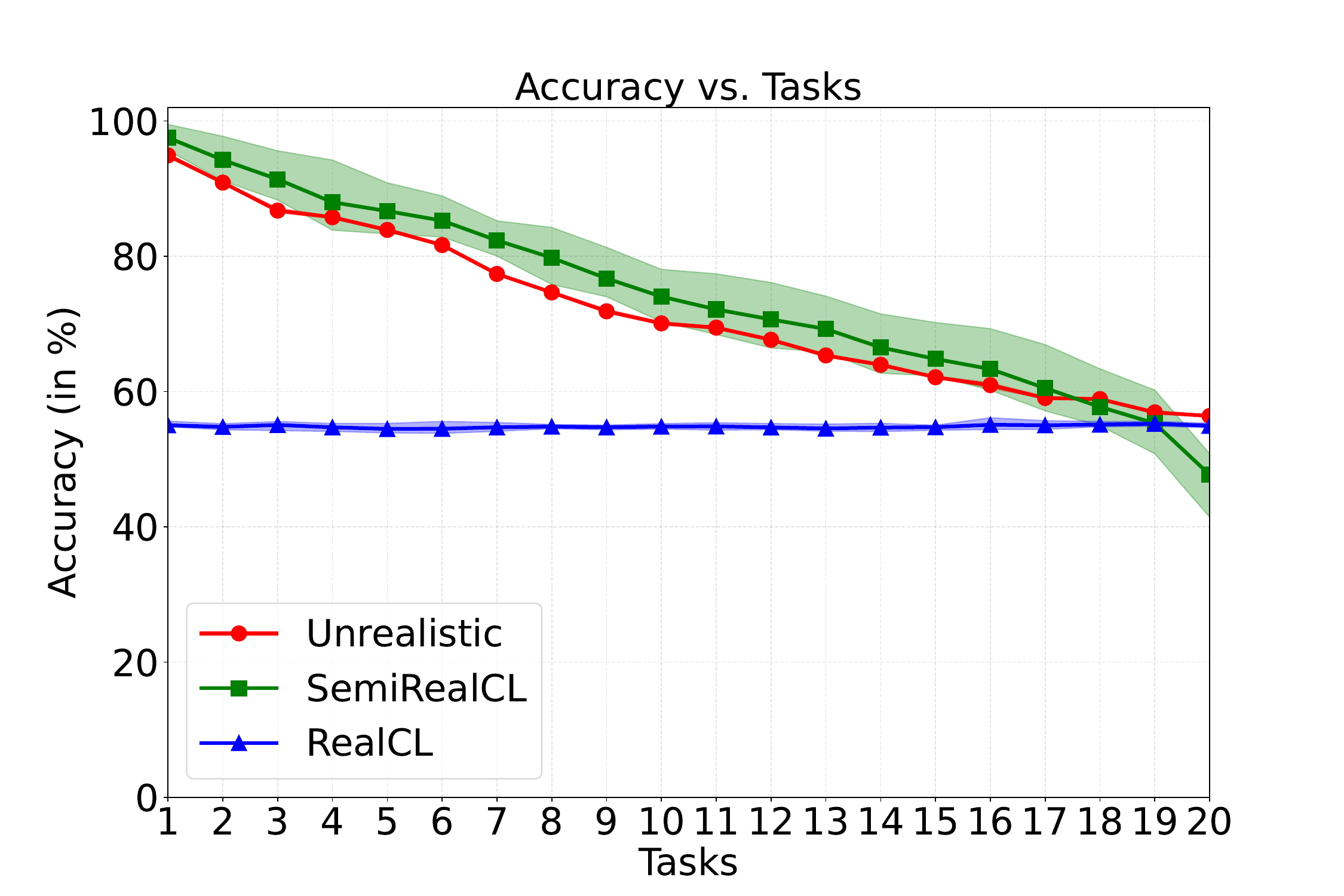}
    \caption{\small Task accuracy.}
  \end{subfigure}%
  \begin{subfigure}{0.33\textwidth}
    \centering
    \includegraphics[width=\textwidth]{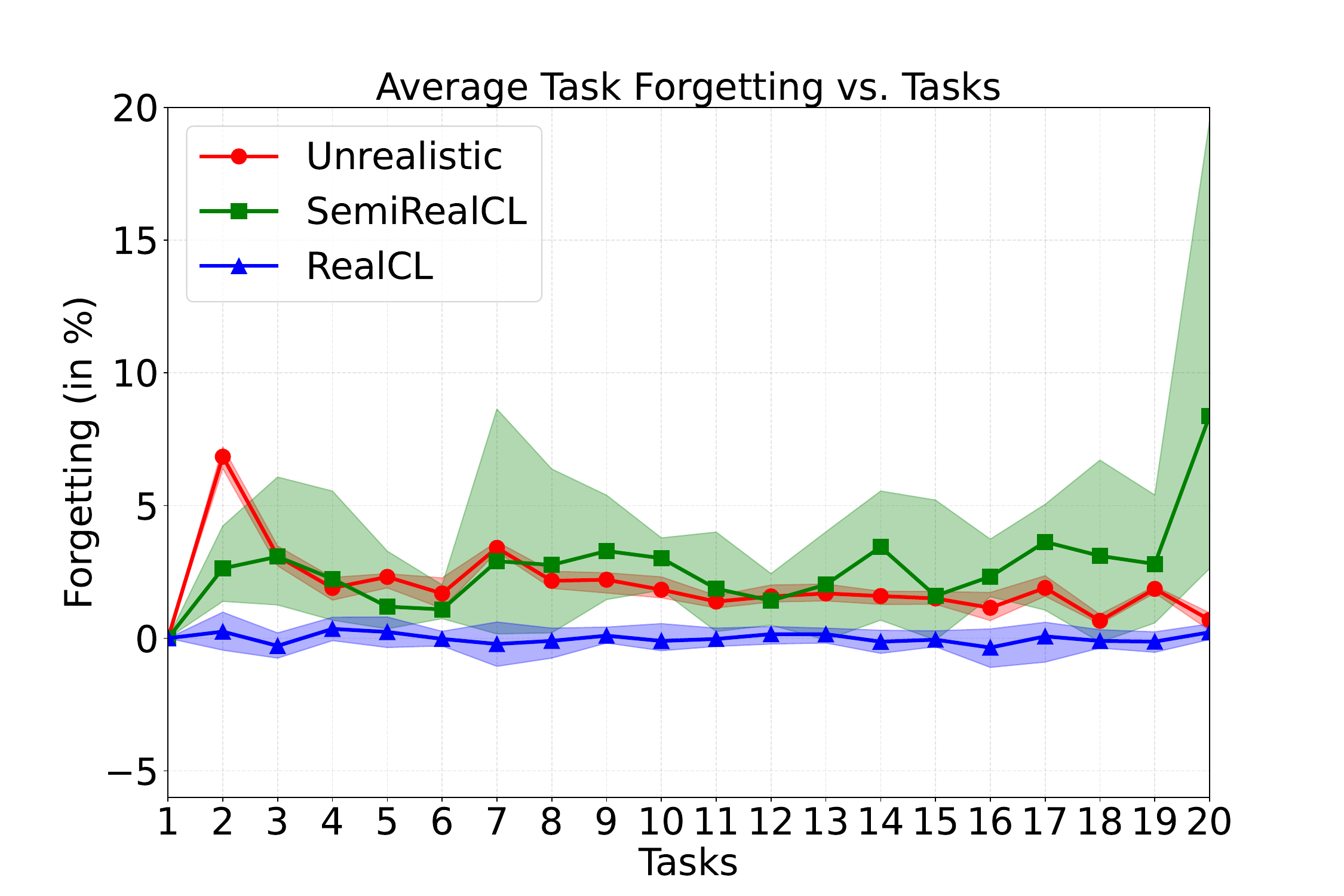}
    \caption{\small Average task forgetting.}
  \end{subfigure}%
  \begin{subfigure}{0.33\textwidth}
    \centering
    \includegraphics[width=\textwidth]{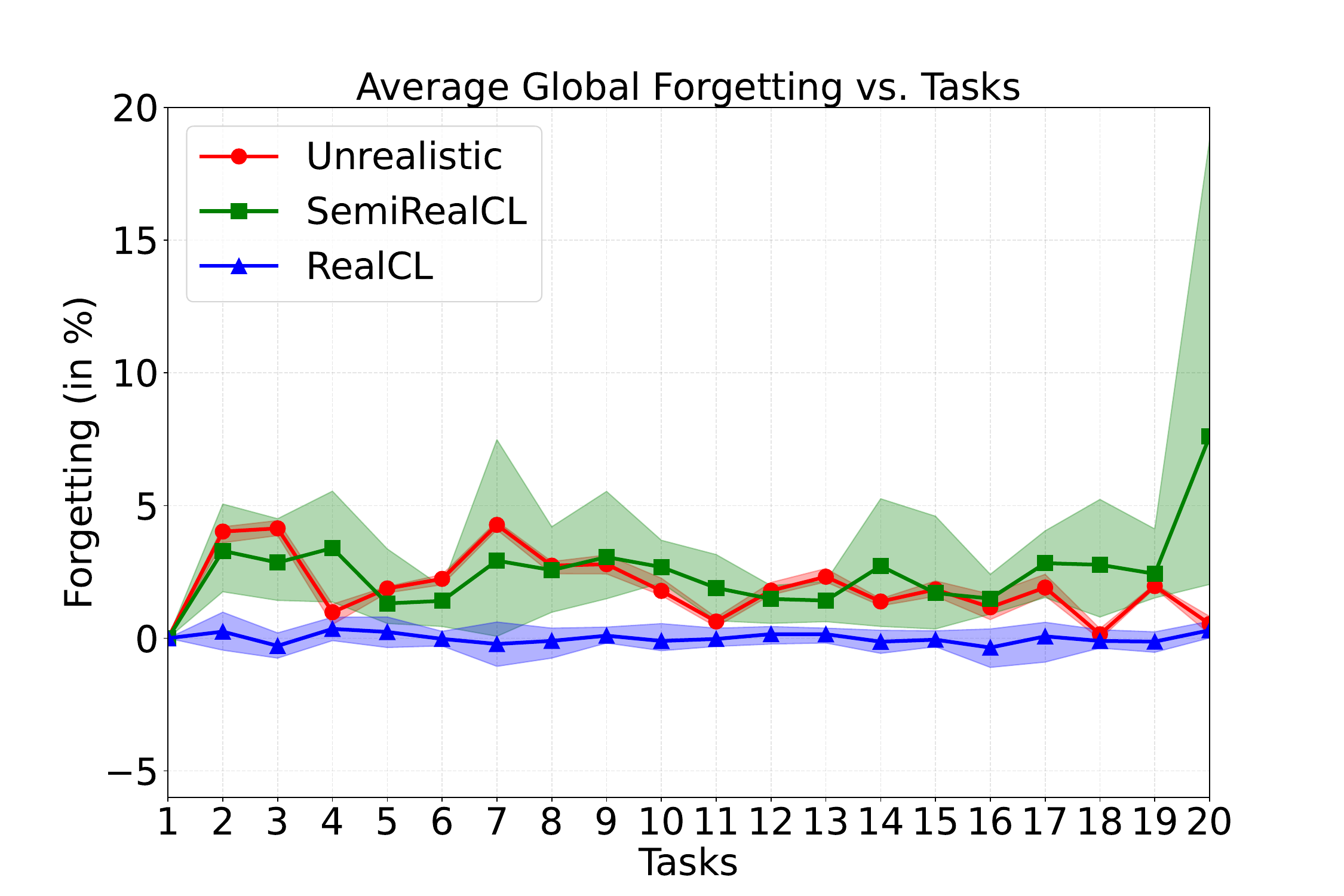}
    \caption{\small Average global forgetting.}
  \end{subfigure}
  \caption{Results of CLARE for TinyImagenet in the three scenarios with memory module size of 2K.}
   \label{fig:tiny_results_2000}
\end{figure}

\begin{figure}[h!]
  \centering
  \begin{subfigure}{0.33\textwidth}
    \centering
    \includegraphics[width=\textwidth]{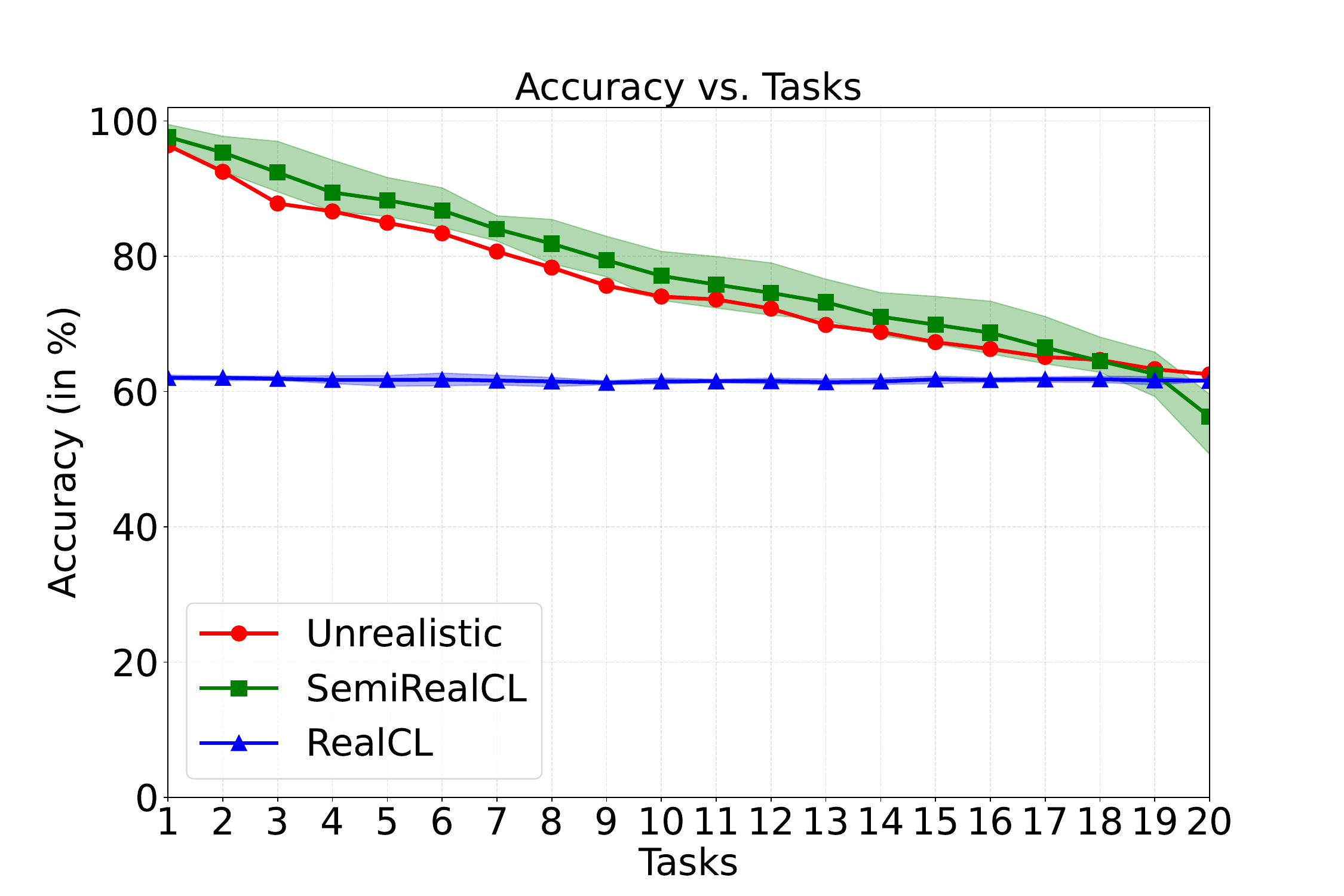}
    \caption{\small Task accuracy.}
  \end{subfigure}%
  \begin{subfigure}{0.33\textwidth}
    \centering
    \includegraphics[width=\textwidth]{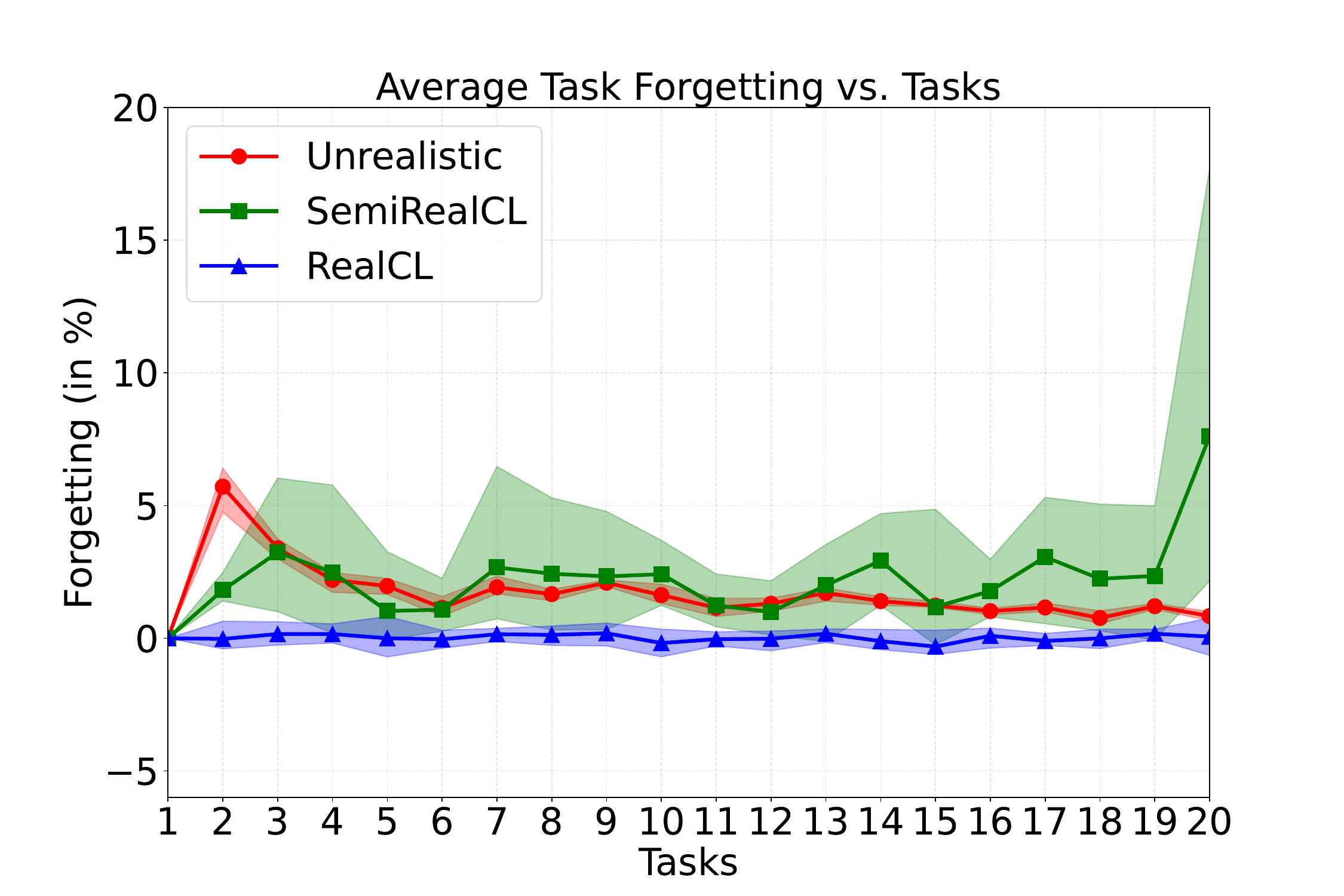}
    \caption{\small Average task forgetting.}
  \end{subfigure}%
  \begin{subfigure}{0.33\textwidth}
    \centering

    \includegraphics[width=\textwidth]{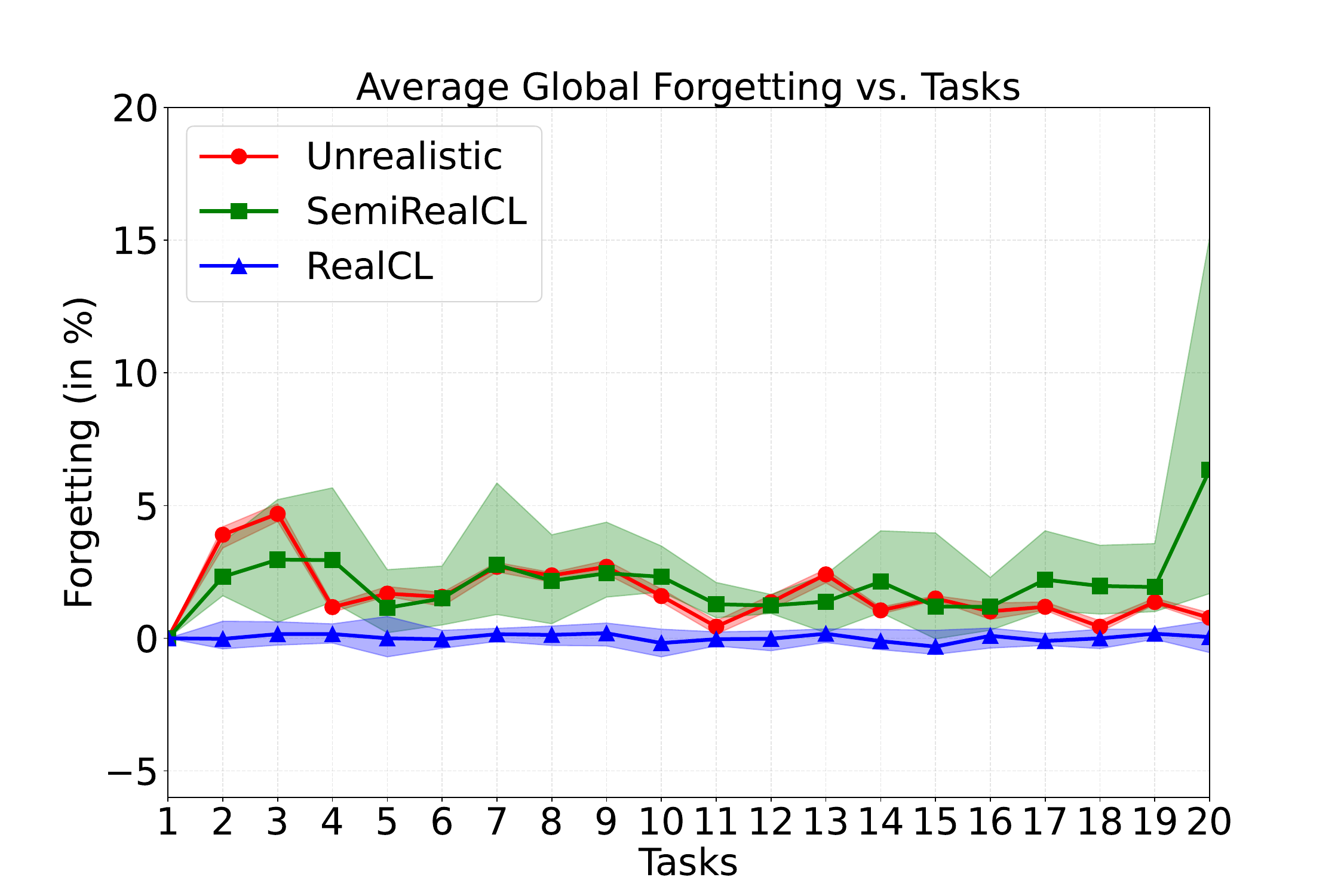}
    \caption{\small Average global forgetting.}
  \end{subfigure}
  \caption{Results of CLARE for TinyImagenet in the three scenarios with memory module size of 4K.}
   \label{fig:tiny_results_4000}
\end{figure}

\begin{figure}[h!]
  \centering
  \begin{subfigure}{0.33\textwidth}
    \centering
    \includegraphics[width=\textwidth]{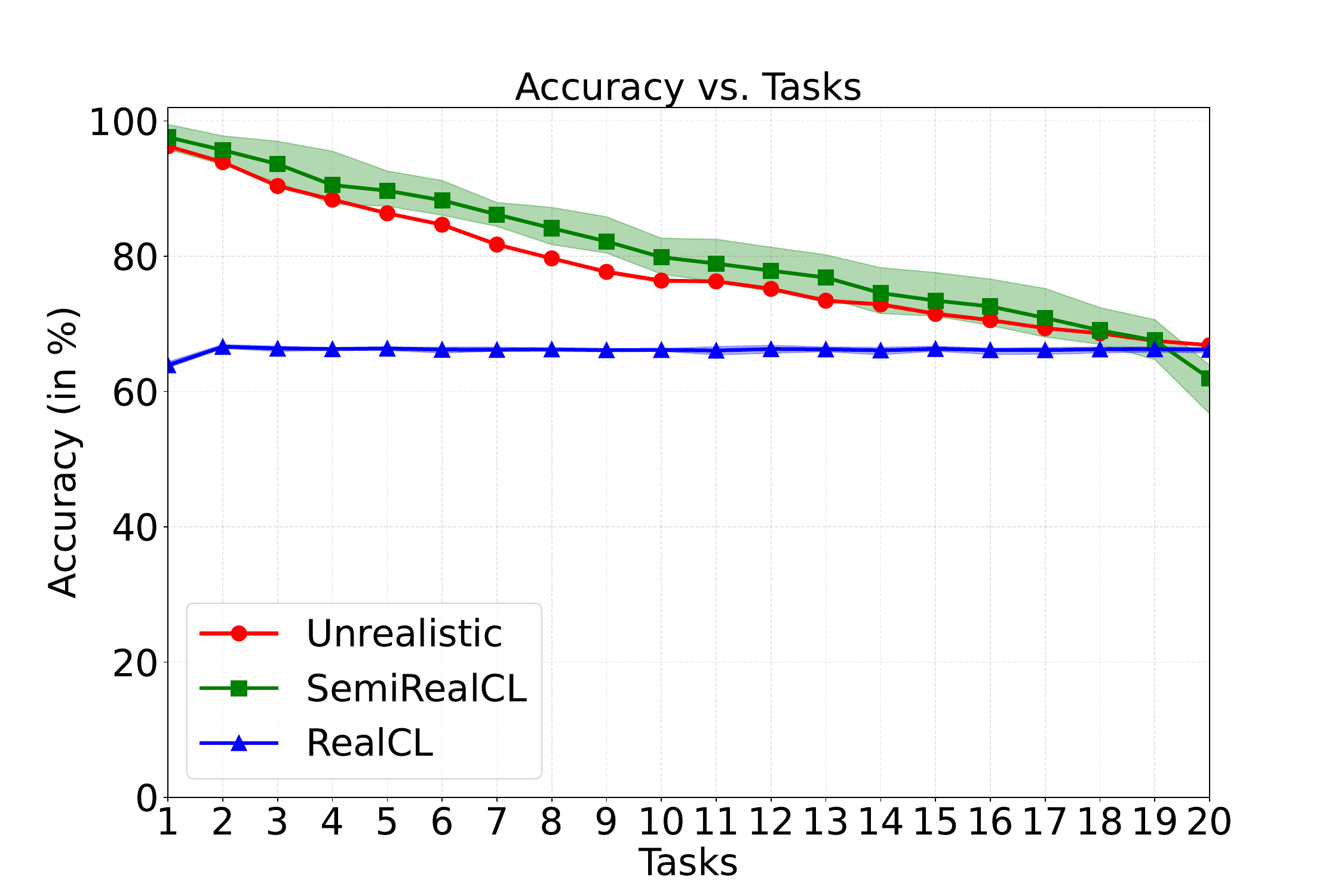}
    \caption{\small Task accuracy.}
  \end{subfigure}%
  \begin{subfigure}{0.33\textwidth}
    \centering
    \includegraphics[width=\textwidth]{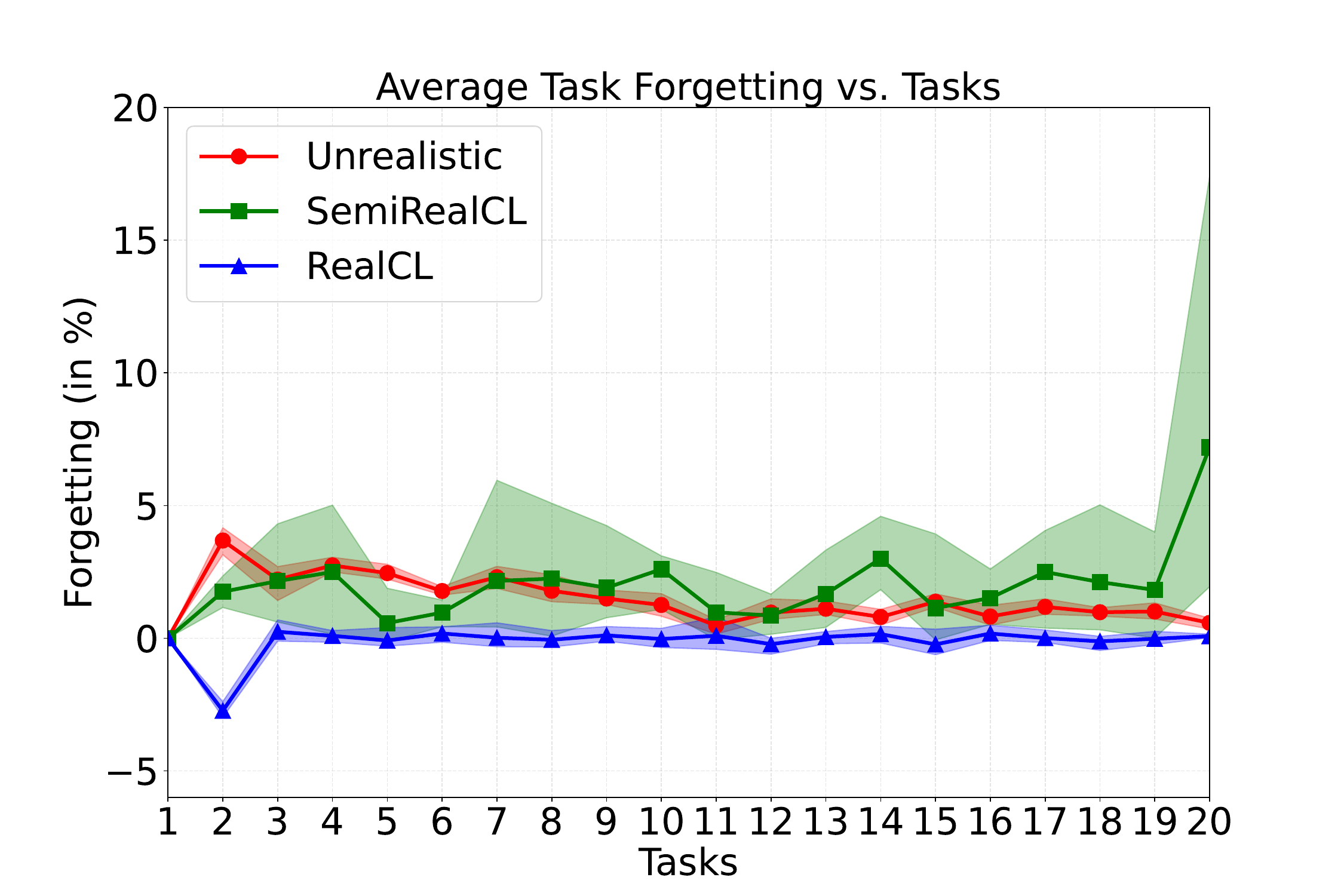}
    \caption{\small Average task forgetting.}
  \end{subfigure}%
  \begin{subfigure}{0.33\textwidth}
    \centering

    \includegraphics[width=\textwidth]{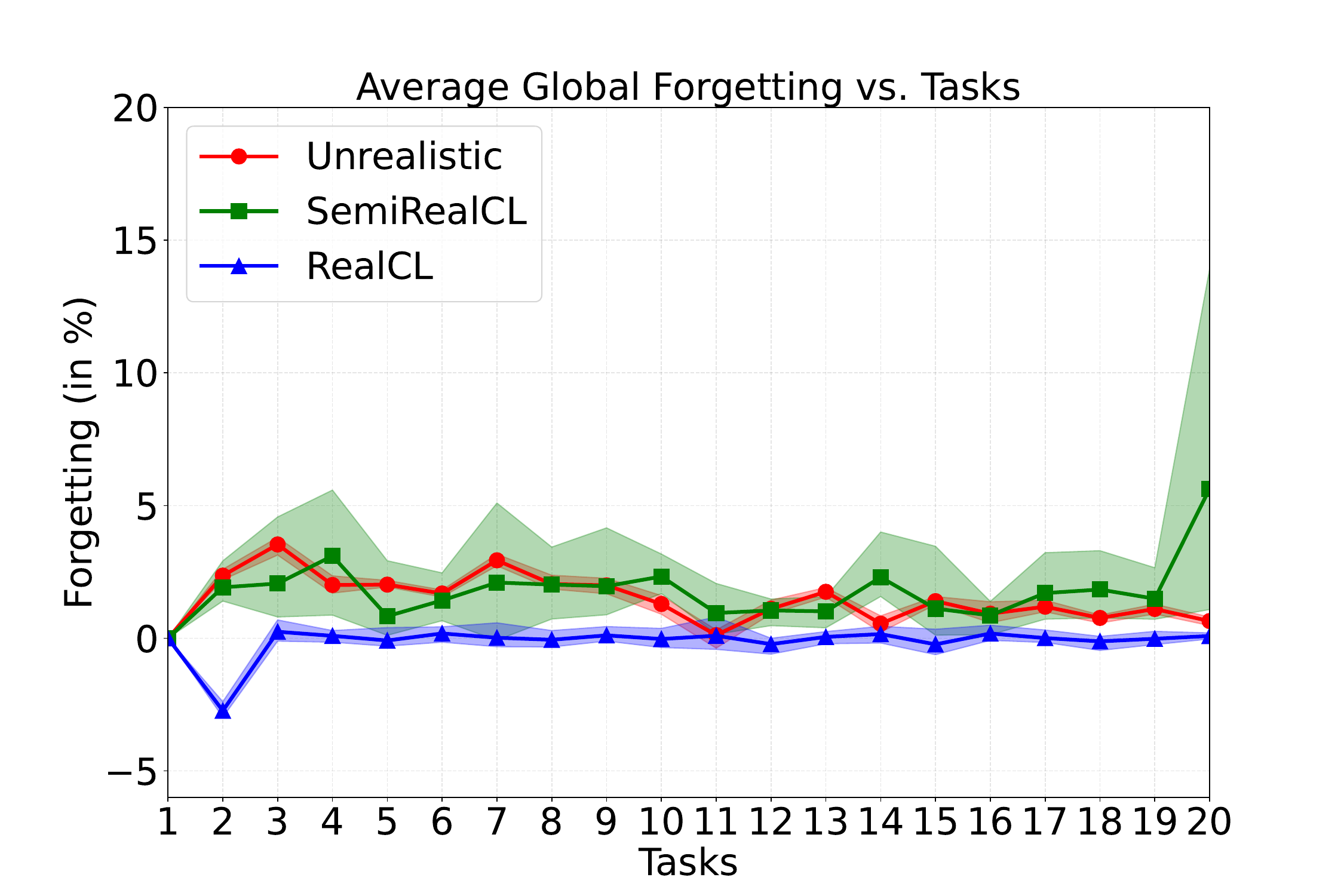}
    \caption{\small Average global forgetting.}
  \end{subfigure}
  \caption{Results of CLARE for TinyImagenet in the three scenarios with memory module size of 8K.}
  \label{fig:tiny_results_8000}
\end{figure}

\cref{cifar_results_1000,fig:cifar100_results_2000,fig:cifar100_results_4000,fig:cifar100_results_8000,fig:tiny_results_2000,fig:tiny_results_4000,fig:tiny_results_8000} show the detailed results for different scenarios and datasets. 
CLARE model reports the highest accuracy on the CIFAR10 dataset, which has only 10 classes, and can classify and learn them in all scenarios. 
However, CIFAR100 and TinyImagenet datasets have more classes (100 and 200, respectively), which makes the classification task more challenging. 
The average accuracies of CLARE model on these datasets are lower, but still comparable to other state-of-the-art methods. Moreover, it is noteworthy that CLARE model achieves this accuracy by only training on the existing samples in the memory buffer.
Interestingly, these results indicate that forgetting is indeed mitigated in the realistic setting.
A detailed analysis reveals the following aspects and causes.
First, we believe that the main reason is due to the artificiality of the scenario: it is not a purely realistic scenario, but rather a simulation of a real situation, with a closed data set, such as CIFAR-100 or TinyImageNet.
This makes the difficulty of the scenario less than in a real situation, where not handling a closed set of classes would present a greater variability over the training time of the models. Nevertheless, our RealCL scenario is much more realistic than previous proposals handled in the literature.
Second, forgetting has been mitigated, as our experiments reveal, by our own CLARE model.
The minimal performance degradation observed in the realistic setting can be attributed to CLARE’s architecture, which leverages the fixed representations of a robust pre-trained model alongside the dynamic adjustment capabilities of Dyn-NAN. This design effectively mitigates forgetting by isolating task-specific adaptations within the Dyn-NAN module while preserving shared knowledge in the pre-trained encoder. While forgetting is not entirely absent, the reduced degradation highlights CLARE’s capacity to handle the unstructured nature of the realistic incremental setup, as demonstrated by its low global and task forgetting rates. These results emphasize the model’s ability to balance stability and plasticity effectively.

\begin{figure}[h!]
\centering
\includegraphics[width=0.50\linewidth]{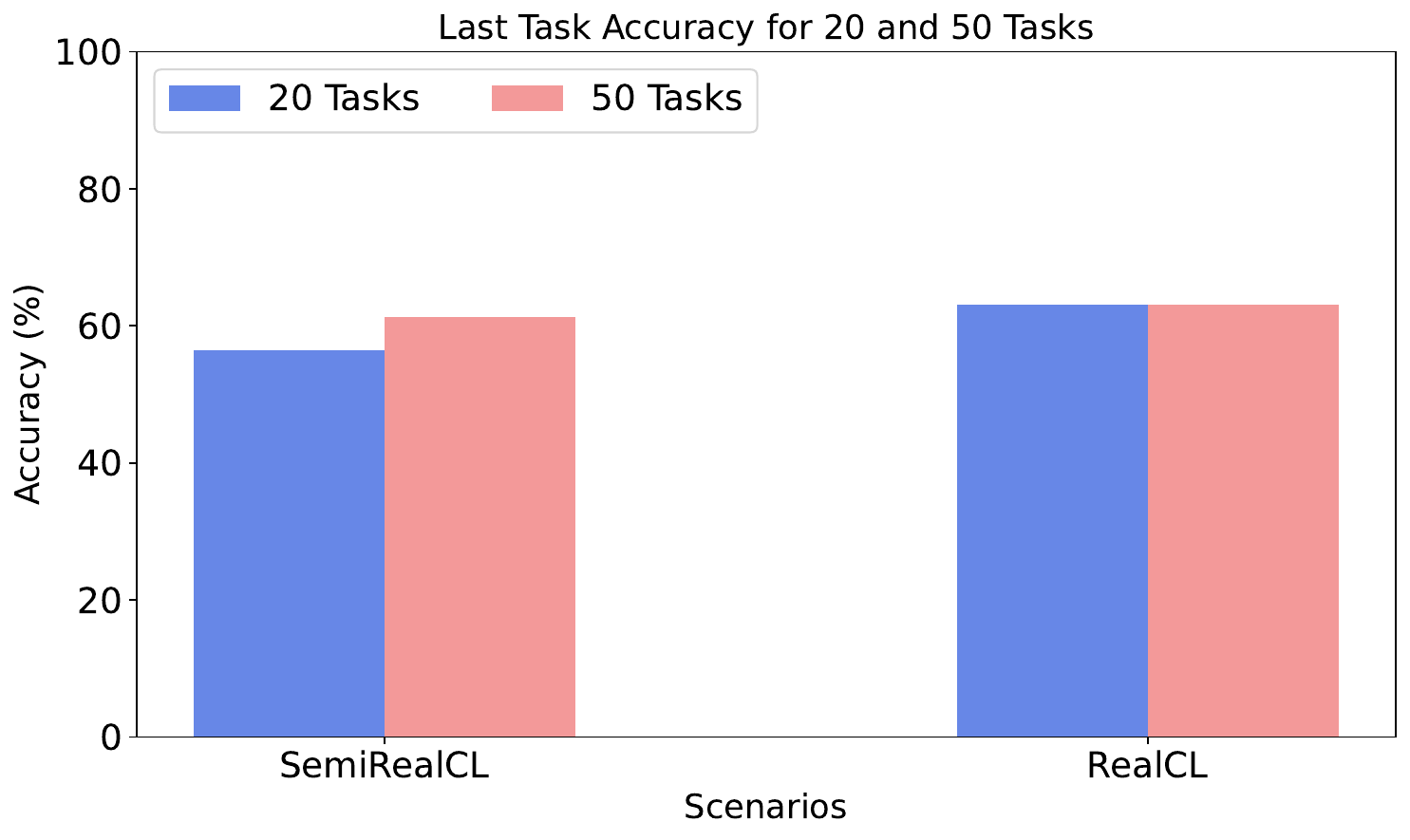}
\caption{\label{fig:20_50}Comparative analysis of the last task accuracy of CLARE for 20 and 50 tasks in both SemiRealCL and RealCL scenarios.}
\end{figure}

As the size of the memory module increases, CLARE demonstrates improved performance metrics.
The question that we want to address now is: How does CLARE perform when the number of tasks increases?
We have evaluated CLARE, which uses a powerful pre-trained model, on the RealCL and SemiRealCL settings.
These scenarios pose more challenges as the number of tasks increases. 
Classic continual learning models are prone to catastrophic forgetting, which means losing old knowledge when learning new tasks. 
This aspect can be relevant specially when the number of tasks grows. 
We have performed the evaluation using the CIFAR-100 dataset with 2K memory size.
Table \ref{cifar_50tasks} shows that CLARE can keep its performance on all tasks, even in the RealCL setup. 
Additionally, in the SemiRealCL scenario CLARE outperforms the results obtained for 20 tasks in terms of Last Task Accuracy (see Figure \ref{fig:20_50}).

These results reinforce the observation that, although increasing the replay memory generally leads to better performance, the effectiveness of CLARE cannot be explained solely by the replay buffer.
Notably, when the number of incremental tasks increases from 20 to 50 in the CIFAR-100 benchmark, CLARE maintains consistently strong performance using the same 2K memory budget despite the substantially higher task fragmentation. This suggests that the proposed learning strategy effectively complements the replay mechanism, enabling the model to preserve knowledge under increasingly challenging continual learning conditions.
Therefore, the replay buffer should be regarded as an enabling component of CLARE rather than as the sole explanation for its performance.


\begin{table}[h!]
\centering
\sisetup{table-align-text-post=false}
\begin{tabular}{l|c|c}
\toprule
\multicolumn{1}{c|}{CIFAR - 100} & \multicolumn{1}{c|}{SemiRealCL} & \multicolumn{1}{c}{RealCL} \\ \midrule
\multicolumn{1}{c|}{Number of Tasks} & 50 Tasks & 50 Tasks\\ \midrule
Last Task Accuracy & 61.23 {\small $\pm$0.71} & 63.08 {\small $\pm$0.06} \\
Average Accuracy & 77.72 {\small $\pm$2.40} & 63.01 {\small $\pm$0.52} \\
Average Global Forgetting & 0.74 {\small $\pm$0.00} & -0.01 {\small $\pm$0.00}  \\
Average Task Forgetting & 0.72 {\small $\pm$0.03} & 0.32 {\small $\pm$0.15} \\
\bottomrule
\end{tabular}
\caption{Performance metrics for 50 Tasks on the CIFAR-100 dataset with a 2K memory module in SemiRealCL and RealCL scenarios (Values in Percentage).}
\label{cifar_50tasks}
\end{table}

\subsubsection{Comparison with state-of-the-art models}

We report a comparison of CLARE with two recent and representative continual learning methods: MEMO~\citep{Zhou2022AMO} and APER~\citep{aper}.
They are strong state-of-the-art baselines that exemplify two relevant paradigms, memory-based rehearsal and pre-trained model adaptation, respectively.
At the time of submisision, both MEMO and APER reported the best results in traditional continual learning settings, i.e. our unrealistic scenario.
Note that the methods to be compared with CLARE do not all employ the same backbone architectures.
Consequently, the reported results should not be interpreted as an isolated comparison of feature extractors, but as a
comparison of performance when we move from traditional unrealistic scenarios to the proposed RealCL one, with state-of-the-art models. For this reason, we compare complete continual learning systems following the experimental protocols adopted in the corresponding original works.

MEMO~\citep{Zhou2022AMO} is a sophisticated method that integrates a memory module to preserve and recall significant instances from prior classes in a continual learning setting.
MEMO's technique creatively introduces extra residual layers instead of reconstructing the full backbone, effectively reducing memory expenses.
MEMO has been specifically developed for class-incremental scenarios.
In fact, it includes a sample selection mechanism for the memory buffer it employs, which requires categories to sequentially arrive.
To tailor MEMO for our new RealCL paradigm, we have had to adapt this selection mechanism.
In RealCL, training samples are randomly allocated across different tasks without predetermined class distributions within each task.
This allows for the possibility of class recurrence across different tasks.
Given this setup, a task may not provide access to all samples from a class, as MEMO needs, thereby limiting the selection of the most representative samples available within that task.
Consequently, there is no predetermined quota for the number of samples per class in the memory buffer.
We initially select the most significant samples from the current task’s classes using MEMO’s sample selection method based on \citep{welling2009}. These chosen samples are then added to the memory buffer through a random selection process that balances between the important samples of previous tasks and the new samples of the current task. To accommodate the new samples, we clear out an equivalent space in the memory buffer, ensuring that the space is equally distributed among the classes presented in each task.
These samples, once chosen, are incorporated into the training data for subsequent tasks.
MEMO model \citep{Zhou2022AMO}, which utilizes a ResNet32-based architecture \citep{He2015DeepRL}, processes the data from the current task alongside the representative samples from previous tasks preserved in the memory buffer.
Consequently, the absence of new classes may not necessitate significant alterations or expansions in the specialized layers, as their primary role is to adapt to the nuances of new class features.

APER~\citep{aper} belongs to the family of pre-trained models for continual learning, like our CLARE approach. 
APER employs a pre-trained model as input for an elaborated adaptation process that integrates some parameter efficient tuning techniques.
As it is reported by \citet{aper}, APER obtains the best results when it is combined with the Adaptformer~\citep{adapter} technique.
Therefore, this is the version of APER that we have decided to integrate in our novel RealCL paradigm.
As for the implementation of APER, we have used the official one released with \citep{sun2023pilot}.

Finally, we also include in this analysis a baseline model, termed naive fine-tuning, that simply involves a learning strategy that consists in updating the model’s parameters with newly acquired data, albeit in the absence of any regularization mechanisms or memory-based strategies. The fine-tuning is conducted using a ResNet-32 architecture \citep{He2015DeepRL}. Given the RealCL scenario, where classes may reappear across various tasks, the model is less likely to experience significant forgetting. It has the potential to continually refine the representations of previous classes as new instances of these classes emerge in subsequent tasks.

\begin{figure}[h!]
    \centering
    \begin{subfigure}[b]{0.45\linewidth}
        \includegraphics[width=\linewidth]{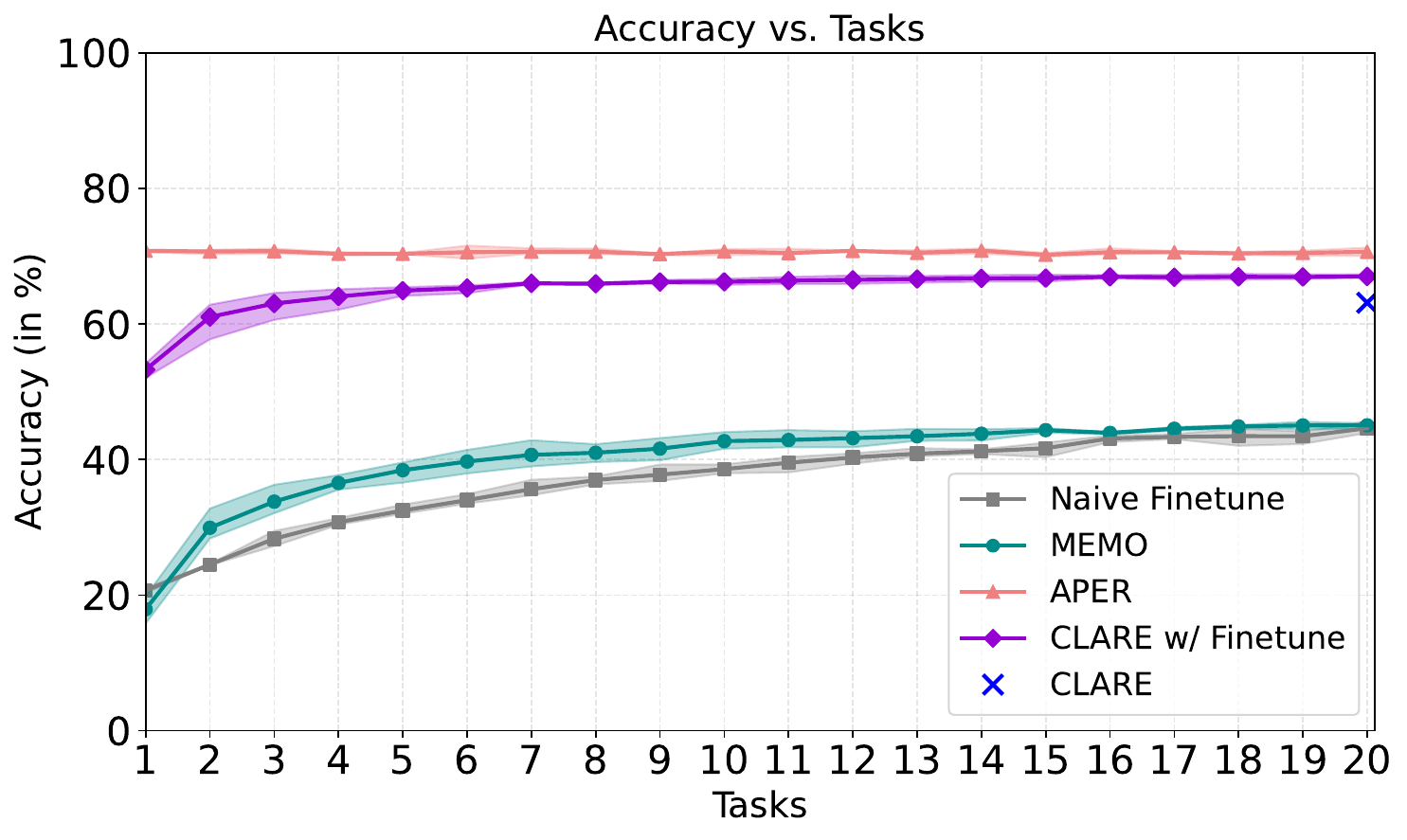}
        \caption{Comparison with state-of-the-art models in the RealCL scenario.}
        \label{fig:memo_fine_real}
    \end{subfigure}
    \hfill 
    \begin{subfigure}[b]{0.491\linewidth}
        \includegraphics[width=\linewidth]{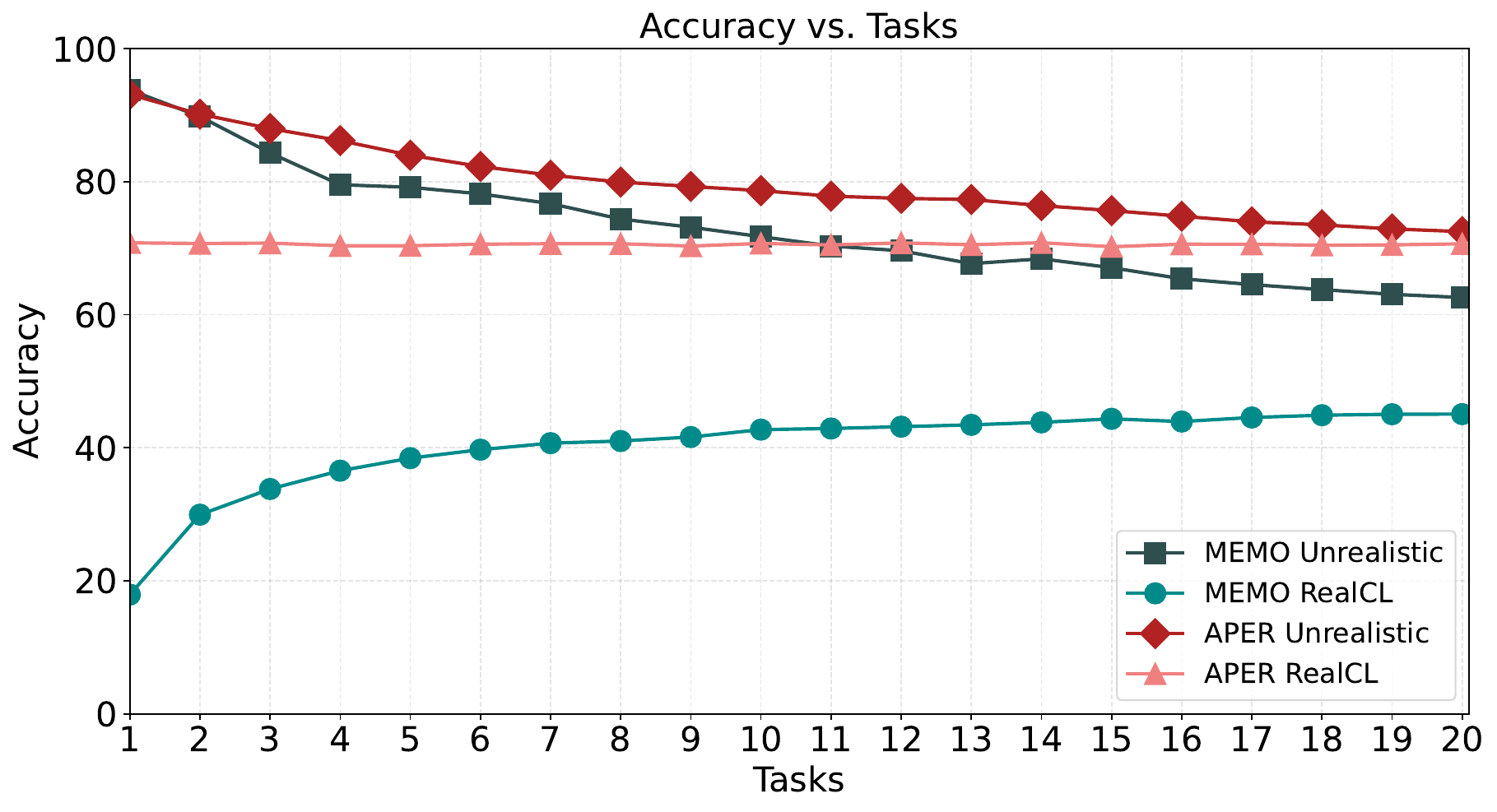}
        \caption{Comparison of MEMO and APER models in the RealCL and the Unrealistic scenarios.}
        \label{fig:memo_un_real}
    \end{subfigure}
    \caption{Results in CIFAR-100 dataset. (a) Performance comparison of CLARE with and without fine-tuning against various state-of-the-art models in RealCL scenario. (b) Comparative analysis of the performance of MEMO and APER in unrealistic scenario versus RealCL outcomes. 
 }
\end{figure}

We start with the comparison with the state of the art in the CIFAR-100 dataset.
Figure \ref{fig:memo_fine_real} and Table \ref{cifar_finetune} show the results obtained in our experiments.
Interestingly, both MEMO and APER experience a reduction in performance when adapted and evaluated in the novel RealCL scenario. As it was reported in \citep{Zhou2022AMO}, in the standard unrealistic scenario, MEMO achieved a Last Task Accuracy of 62.59\% after 20 tasks, whereas in our RealCL implementation, it reports 45.06\%.
For the APER model, the Last Task Accuracy in the unrealistic setting was of 72.5\%, while for our RealCL we report 70.66\%.
We argue that MEMO, APER and their traditional competitors perform well in an ideal CL setting (i.e. our unrealistic scenario) with complete access to class distributions and specific data samples, where the setting follows a pure class-incremental process. However, their effectiveness tend to decrease in the RealCL scenario.
It is relevant to compare, in fact, how these algorithms evolve in both scenarios, the RealCL and the unrealistic or traditional one.
In Figure \ref{fig:memo_un_real} we show this interesting comparison.
Clearly, our realistic scenario pushes these models to their limits, which is reflected in a drop in their accuracy and even in a learning stagnation, as observed in the APER model, whose accuracy remains stuck from the first task.

\begin{table}[h!]
\centering
\sisetup{table-align-text-post=false}
\scalebox{0.60}{ 
\begin{tabular}{l|c|c|c|c|c}
\toprule
\multicolumn{1}{c|}{CIFAR - 100} & \multicolumn{1}{c|}{Naive Fine-tuning} & \multicolumn{1}{c|}{MEMO~\citep{Zhou2022AMO}} & \multicolumn{1}{c|}{APER~\citep{aper}}  & \multicolumn{1}{c|}{CLARE}  & \multicolumn{1}{c}{CLARE with Fine-tuning} \\ \midrule
\multicolumn{1}{c|}{Number of Tasks} & 20 Tasks & 20 Tasks & 20 Tasks & 20 Tasks & 20 Tasks \\ \midrule
Last Task Accuracy & 44.63 {\small$\pm$0.73} & 45.06 {\small $\pm$0.37} & 70.66 {\small $\pm$0.51} & 63.39 {\small $\pm$0.40} & 67.03 {\small $\pm$0.22} \\
Average Accuracy & 37.07 {\small $\pm$6.73} & 40.17 {\small $\pm$6.60} & 70.58 {\small $\pm$0.25} & 63.34 {\small $\pm$0.14} & 65.20 {\small $\pm$3.20}\\
Average Task Forgetting & -1.20 {\small $\pm$1.10} & -1.35 {\small $\pm$2.70} & -0.17 {\small $\pm$0.25} & -0.002 {\small $\pm$0.22} & -0.68 {\small $\pm$1.74}\\
\bottomrule
\end{tabular}
}
\caption{Comparison with state-of-the-art models in the RealCL scenario. Performance Metrics on CIFAR-100 Dataset with a 2K Buffer Size for CLARE. Values in Percentage.}
\label{cifar_finetune}
\end{table}

\begin{table}[h!]
\centering
\sisetup{table-align-text-post=false}
\scalebox{0.60}{ 
\begin{tabular}{l|c|c|c}
\toprule
\multicolumn{1}{c|}{CUB - 200} & \multicolumn{1}{c|}{APER~\citep{aper}} & \multicolumn{1}{c|}{CLARE} & \multicolumn{1}{c}{CLARE with Fine-tuning} \\ \midrule
\multicolumn{1}{c|}{Number of Tasks} & 20 Tasks  &  20 Tasks &  20 Tasks \\ \midrule
Last Task Accuracy & 55.73 {\small $\pm$0.025} & 61.41 {\small $\pm$0.66} & 66.48 {\small $\pm$0.91} \\
Average Accuracy & 55.42 {\small $\pm$1.34} & 59.36 {\small $\pm$0.10} & 62.42 {\small $\pm$0.87} \\
Average Task Forgetting & -0.04 {\small $\pm$1.81} & -0.67 {\small $\pm$0.09} & -0.92 {\small $\pm$0.09} \\
\bottomrule
\end{tabular}
}
\caption{Performance metrics for 20 Tasks on the CUB200 dataset, with a 2K memory module, in a RealCL scenario. We compare APER and CLARE (with and without fine-tunning). Values in Percentage.}
\label{tab:CuB200_20tasks}
\end{table}

The results for the experimental evaluation with the dataset CUB200 are shown in Table \ref{tab:CuB200_20tasks}.
Here, CLARE, with and without fine-tuning, improves the performance of APER according to all metrics.

Our main conclusions of this comparison conducted with state-of-the-art methods are as follows.  
First, methods originally proposed for traditional continual learning benchmarks experience a substantial performance degradation under the novel RealCL setting, suggesting that state-of-the-art performance in conventional unrealistic benchmarks does not necessarily translate to more realistic continual learning scenarios.
With our study we show that it is important to subject CL models to a scenario like RealCL, where possible deficiencies are not masked when class distributions do not appear artificially and incrementally ordered with tasks. 
Interestingly, our observations revealed that the naive fine-tuning model, APER and MEMO demonstrate negative values in task forgetting metrics. This suggests that the models are not merely retaining previously acquired knowledge, but are also contributing to an enhancement in overall accuracy as they are exposed to new data from subsequent tasks. However, this trend appears to plateau once the models have processed a certain number of tasks, indicating a stabilization point beyond which the introduction of additional data does not yield significant improvements in model performance.
This effect is particularly relevant for APER, which does not seem to continue learning after the first task (see Figure \ref{fig:memo_un_real}).  
This aligns with how APER is trained, as detailed in \citep{aper}.  
APER adapts the pre-trained backbone only in the first task and then employs a strategy based on prototypes for every category, along with parameter fine-tuning mechanisms such as Adaptformer~\citep{adapter}.
While this strategy seems effective for an the unrealistic scenario, i.e. the standard class-incremental learning scenario, it is not suitable for our new RealCL setup.

Second, our novel CLARE model presents a compelling and competitive option for realistic scenarios.
In the experiments with CIFAR-100, the APER model reports a last task accuracy slightly higher than CLARE. However, this is not the case with the CUB-200 dataset, where CLARE achieves better results in the realistic scenario.
It is worth mentioning that APER, as detailed by \citet{aper}, is a significantly more complex model than CLARE, integrating multiple parameter-efficient tuning techniques along with the pre-trained model.
These results indicate that the observed improvements cannot be attributed solely to the use of a stronger visual backbone, but also to the proposed learning strategy within the RealCL framework.
CLARE depends on a robust pre-trained model capable of identifying general and adaptable features from data. It is complemented by a learning-based memory module that actively selects and updates a set of samples for each task. This dual strategy enables our method to maintain a balance between preserving existing knowledge and assimilating new information, thereby circumventing the pitfalls of catastrophic forgetting or interference.
As demonstrated in Tables \ref{cifar_finetune} and \ref{tab:CuB200_20tasks}, adopting a fine-tuning approach with CLARE leads to improved accuracy across tasks. It is important to note that since the pre-trained model within CLARE remains static, fine-tuning is applied solely to adjust the learned weights of the Dyn-NAN block. Further observations reveal that CLARE, upon fine-tuning, also displays negative task forgetting, indicating an expansion in the model’s learning capabilities as the number of tasks increases.

\section{Conclusions}   \label{sec:conclusion}
With this paper, we had two main objectives. The first one was to propose to the community the RealCL paradigm for evaluating CL solutions.
The novel RealCL formulation has allowed us to explore up to three distinct continual learning scenarios, ranging from a highly structured scenario with controlled data and class distribution to a completely uncontrolled and realistic setting. We conducted a thorough evaluation of our proposed scenarios employing CLARE, a straightforward and model-agnostic methodology that utilizes pre-trained models to facilitate the learning process. This was the second objective of our work.
We believe that especially in these unstructured scenarios, solutions based on pre-trained models can play a fundamental role.
Our CLARE model enables the seamless integration of new knowledge while preserving previously acquired information, eliminating the need for extensive retraining. Moreover, we implemented the proposed realistic scenario for a set of state-of-the-art models and a baseline, and assessed their performance, comparing it with the outcomes achieved by CLARE. Our findings highlight the versatility and robustness of CLARE, particularly when contending with the unpredictability of real-world data.
The code to reproduce all our results can be found at \url{https://github.com/gramuah/clare}.

\textbf{Limitations and future work:}

While CLARE demonstrates strong performance in realistic continual learning scenarios, certain limitations should be acknowledged. 
First, CLARE relies on a compact replay memory, which may introduce privacy concerns or storage constraints in some applications. Although the memory is intentionally bounded and contains only a small set of representative samples, the method still depends on storing part of the past experience. In practice, this design represents a compromise between purely memory-free approaches, which often suffer from severe forgetting, and unrestricted rehearsal strategies that require access to large historical datasets. CLARE mitigates forgetting effectively by combining a small replay buffer with a frozen pre-trained model, but its performance remains influenced by the memory size and the sample selection strategy, both of which could be further optimized.
Additionally, the use of a frozen pre-trained model provides stability and strong transfer capabilities, but it may limit adaptability in highly specialized or rapidly evolving domains. Investigating adaptive fine-tuning strategies that preserve stability while improving domain adaptation constitutes an important direction for future work. Furthermore, although CLARE has been evaluated on standard continual learning benchmarks, its behavior under stronger domain shifts and highly imbalanced real-world data distributions deserves further investigation.

Regarding our second contribution, the proposed RealCL scenario, it is important to emphasize that RealCL defines a continual learning protocol rather than a specific algorithm. One of the assumptions of RealCL is that the complete historical datasets from previous tasks are no longer available once learning progresses. However, this formulation is compatible with both replay-based and replay-free continual learning methods. Therefore, the use of a bounded replay buffer in CLARE, or in any other replay-based model, should be understood as an algorithmic choice rather than a requirement imposed by RealCL. In privacy-sensitive applications, replay-free approaches may be preferable, and evaluating such methods under the RealCL protocol constitutes an interesting avenue for future work.
Finally, although RealCL is designed to be closer to real-world continual learning conditions, it still assumes that data arrive in discrete tasks rather than as a fully continuous stream. Future research could explore continual learning systems that dynamically integrate incoming data without relying on predefined task boundaries, further narrowing the gap between benchmark protocols and a real-world deployment scenario.

\section*{Acknowledgements}
This research was partially funded by projects: GUIDANCE-4D, with reference CM/DEMG/2024-028, of CAM-UAH;
NAVIGATOR-D, with reference PID2023-148310OB-I00 from the
Ministry of Science and Innovation of Spain; NAVISOCIAL, with reference 2023/00405/001 from the University of Alcal\'a; and MOEVE Chair at the University of Alcal\'a.\\

\bibliographystyle{elsarticle-harv}

\begin{thebibliography}{51}
\expandafter\ifx\csname natexlab\endcsname\relax\def\natexlab#1{#1}\fi
\providecommand{\url}[1]{\texttt{#1}}
\providecommand{\href}[2]{#2}
\providecommand{\path}[1]{#1}
\providecommand{\DOIprefix}{doi:}
\providecommand{\ArXivprefix}{arXiv:}
\providecommand{\URLprefix}{URL: }
\providecommand{\Pubmedprefix}{pmid:}
\providecommand{\doi}[1]{\href{http://dx.doi.org/#1}{\path{#1}}}
\providecommand{\Pubmed}[1]{\href{pmid:#1}{\path{#1}}}
\providecommand{\bibinfo}[2]{#2}
\ifx\xfnm\relax \def\xfnm[#1]{\unskip,\space#1}\fi
\bibitem[{Aljundi et~al.(2019)Aljundi, Lin, Goujaud and
  Bengio}]{Aljundi2019GradientBS}
\bibinfo{author}{Aljundi, R.}, \bibinfo{author}{Lin, M.},
  \bibinfo{author}{Goujaud, B.}, \bibinfo{author}{Bengio, Y.},
  \bibinfo{year}{2019}.
\newblock \bibinfo{title}{Gradient based sample selection for online continual
  learning}, in: \bibinfo{booktitle}{Neural Information Processing Systems}.
\bibitem[{Araujo et~al.(2022)Araujo, Hurtado, Soto and
  Moens}]{Araujo2022EntropybasedSF}
\bibinfo{author}{Araujo, V.}, \bibinfo{author}{Hurtado, J.},
  \bibinfo{author}{Soto, {\'A}.}, \bibinfo{author}{Moens, M.F.},
  \bibinfo{year}{2022}.
\newblock \bibinfo{title}{Entropy-based stability-plasticity for lifelong
  learning}.
\newblock \bibinfo{journal}{2022 IEEE/CVF Conference on Computer Vision and
  Pattern Recognition Workshops (CVPRW)} , \bibinfo{pages}{3720--3727}.
\bibitem[{Buzzega et~al.(2020)Buzzega, Boschini, Porrello and
  Calderara}]{Buzzega2020RethinkingER}
\bibinfo{author}{Buzzega, P.}, \bibinfo{author}{Boschini, M.},
  \bibinfo{author}{Porrello, A.}, \bibinfo{author}{Calderara, S.},
  \bibinfo{year}{2020}.
\newblock \bibinfo{title}{Rethinking experience replay: a bag of tricks for
  continual learning}.
\newblock \bibinfo{journal}{2020 25th International Conference on Pattern
  Recognition (ICPR)} , \bibinfo{pages}{2180--2187}.
\bibitem[{Cao et~al.(2024)Cao, Tang, Lin, Han, Chen, Wang and
  Sun}]{Cao2023RetentiveOF}
\bibinfo{author}{Cao, B.}, \bibinfo{author}{Tang, Q.}, \bibinfo{author}{Lin,
  H.}, \bibinfo{author}{Han, X.}, \bibinfo{author}{Chen, J.},
  \bibinfo{author}{Wang, T.}, \bibinfo{author}{Sun, L.}, \bibinfo{year}{2024}.
\newblock \bibinfo{title}{Retentive or forgetful? diving into the knowledge
  memorizing mechanism of language models}, in:
  \bibinfo{booktitle}{LREC-COLING}.
\bibitem[{Castro et~al.(2018)Castro, Mar{\'i}n-Jim{\'e}nez, Mata, Schmid and
  Karteek}]{Castro2018EndtoEndIL}
\bibinfo{author}{Castro, F.M.}, \bibinfo{author}{Mar{\'i}n-Jim{\'e}nez, M.J.},
  \bibinfo{author}{Mata, N.G.}, \bibinfo{author}{Schmid, C.},
  \bibinfo{author}{Karteek, A.}, \bibinfo{year}{2018}.
\newblock \bibinfo{title}{End-to-end incremental learning}, in:
  \bibinfo{booktitle}{European Conference on Computer Vision}.
\bibitem[{Chen et~al.(2019)Chen, Min, Li and Jiang}]{Chen2019}
\bibinfo{author}{Chen, C.}, \bibinfo{author}{Min, W.}, \bibinfo{author}{Li,
  X.}, \bibinfo{author}{Jiang, S.}, \bibinfo{year}{2019}.
\newblock \bibinfo{title}{Hybrid incremental learning of new data and new
  classes for hand-held object recognition}.
\newblock \bibinfo{journal}{Journal of Visual Communication and Image
  Representation} \bibinfo{volume}{58}, \bibinfo{pages}{138--148}.
\bibitem[{Chen et~al.(2022)Chen, Ge, Tong, Wang, Song, Wang and Luo}]{adapter}
\bibinfo{author}{Chen, S.}, \bibinfo{author}{Ge, C.}, \bibinfo{author}{Tong,
  Z.}, \bibinfo{author}{Wang, J.}, \bibinfo{author}{Song, Y.},
  \bibinfo{author}{Wang, J.}, \bibinfo{author}{Luo, P.}, \bibinfo{year}{2022}.
\newblock \bibinfo{title}{Adaptformer: Adapting vision transformers for
  scalable visual recognition}, in: \bibinfo{booktitle}{NeurIPS}.
\bibitem[{Chrabaszcz et~al.(2017)Chrabaszcz, Loshchilov and
  Hutter}]{Chrabaszcz2017ADV}
\bibinfo{author}{Chrabaszcz, P.}, \bibinfo{author}{Loshchilov, I.},
  \bibinfo{author}{Hutter, F.}, \bibinfo{year}{2017}.
\newblock \bibinfo{title}{A downsampled variant of imagenet as an alternative
  to the cifar datasets}.
\newblock \bibinfo{journal}{ArXiv} \bibinfo{volume}{abs/1707.08819}.
\bibitem[{De~Lange and Tuytelaars(2021)}]{dalange2021}
\bibinfo{author}{De~Lange, M.}, \bibinfo{author}{Tuytelaars, T.},
  \bibinfo{year}{2021}.
\newblock \bibinfo{title}{Continual prototype evolution: Learning online from
  non-stationary data streams}, in: \bibinfo{booktitle}{2021 IEEE/CVF
  International Conference on Computer Vision (ICCV)}, pp.
  \bibinfo{pages}{8230--8239}.
\bibitem[{Dosovitskiy et~al.(2021)Dosovitskiy, Beyer, Kolesnikov, Weissenborn,
  Zhai, Unterthiner, Dehghani, Minderer, Heigold, Gelly, Uszkoreit and
  Houlsby}]{Dosovitskiy2020AnII}
\bibinfo{author}{Dosovitskiy, A.}, \bibinfo{author}{Beyer, L.},
  \bibinfo{author}{Kolesnikov, A.}, \bibinfo{author}{Weissenborn, D.},
  \bibinfo{author}{Zhai, X.}, \bibinfo{author}{Unterthiner, T.},
  \bibinfo{author}{Dehghani, M.}, \bibinfo{author}{Minderer, M.},
  \bibinfo{author}{Heigold, G.}, \bibinfo{author}{Gelly, S.},
  \bibinfo{author}{Uszkoreit, J.}, \bibinfo{author}{Houlsby, N.},
  \bibinfo{year}{2021}.
\newblock \bibinfo{title}{An image is worth 16x16 words: Transformers for image
  recognition at scale}, in: \bibinfo{booktitle}{ICLR}.
\bibitem[{Douillard et~al.(2020)Douillard, Cord, Ollion, Robert and
  Valle}]{Douillard2020PODNetPO}
\bibinfo{author}{Douillard, A.}, \bibinfo{author}{Cord, M.},
  \bibinfo{author}{Ollion, C.}, \bibinfo{author}{Robert, T.},
  \bibinfo{author}{Valle, E.}, \bibinfo{year}{2020}.
\newblock \bibinfo{title}{Podnet: Pooled outputs distillation for small-tasks
  incremental learning}, in: \bibinfo{booktitle}{European Conference on
  Computer Vision}.
\bibitem[{Douillard et~al.(2021)Douillard, Ram'e, Couairon and
  Cord}]{Douillard2021DyToxTF}
\bibinfo{author}{Douillard, A.}, \bibinfo{author}{Ram'e, A.},
  \bibinfo{author}{Couairon, G.}, \bibinfo{author}{Cord, M.},
  \bibinfo{year}{2021}.
\newblock \bibinfo{title}{Dytox: Transformers for continual learning with
  dynamic token expansion}.
\newblock \bibinfo{journal}{2022 IEEE/CVF Conference on Computer Vision and
  Pattern Recognition (CVPR)} , \bibinfo{pages}{9275--9285}.
\bibitem[{French(1993)}]{French1993CatastrophicII}
\bibinfo{author}{French, R.M.}, \bibinfo{year}{1993}.
\newblock \bibinfo{title}{Catastrophic interference in connectionist networks:
  Can it be predicted, can it be prevented?}, in: \bibinfo{booktitle}{Neural
  Information Processing Systems}.
\bibitem[{Goodfellow et~al.(2014)Goodfellow, Mirza, Da, Courville and
  Bengio}]{Goodfellow2014}
\bibinfo{author}{Goodfellow, I.J.}, \bibinfo{author}{Mirza, M.},
  \bibinfo{author}{Da, X.}, \bibinfo{author}{Courville, A.C.},
  \bibinfo{author}{Bengio, Y.}, \bibinfo{year}{2014}.
\newblock \bibinfo{title}{An empirical investigation of catastrophic forgeting
  in gradient-based neural networks}, in: \bibinfo{booktitle}{2nd International
  Conference on Learning Representations, {ICLR} 2014}.
\bibitem[{He et~al.(2015)He, Zhang, Ren and Sun}]{He2015DeepRL}
\bibinfo{author}{He, K.}, \bibinfo{author}{Zhang, X.}, \bibinfo{author}{Ren,
  S.}, \bibinfo{author}{Sun, J.}, \bibinfo{year}{2015}.
\newblock \bibinfo{title}{Deep residual learning for image recognition}.
\newblock \bibinfo{journal}{2016 IEEE Conference on Computer Vision and Pattern
  Recognition (CVPR)} , \bibinfo{pages}{770--778}.
\bibitem[{Hu et~al.(2021)Hu, Shen, Wallis, Allen-Zhu, Li, Wang and
  Chen}]{Hu2021LoRALA}
\bibinfo{author}{Hu, J.E.}, \bibinfo{author}{Shen, Y.},
  \bibinfo{author}{Wallis, P.}, \bibinfo{author}{Allen-Zhu, Z.},
  \bibinfo{author}{Li, Y.}, \bibinfo{author}{Wang, S.}, \bibinfo{author}{Chen,
  W.}, \bibinfo{year}{2021}.
\newblock \bibinfo{title}{Lora: Low-rank adaptation of large language models}.
\newblock \bibinfo{journal}{ArXiv} \bibinfo{volume}{abs/2106.09685}.
\bibitem[{Jia et~al.(2022)Jia, Tang, Chen, Cardie, Belongie, Hariharan and
  Lim}]{Jia2022VisualPT}
\bibinfo{author}{Jia, M.}, \bibinfo{author}{Tang, L.}, \bibinfo{author}{Chen,
  B.C.}, \bibinfo{author}{Cardie, C.}, \bibinfo{author}{Belongie, S.J.},
  \bibinfo{author}{Hariharan, B.}, \bibinfo{author}{Lim, S.N.},
  \bibinfo{year}{2022}.
\newblock \bibinfo{title}{Visual prompt tuning}.
\newblock \bibinfo{journal}{ArXiv} \bibinfo{volume}{abs/2203.12119}.
\bibitem[{Jung et~al.(2023)Jung, Han, Bang and Song}]{Jung2023GeneratingIP}
\bibinfo{author}{Jung, D.}, \bibinfo{author}{Han, D.}, \bibinfo{author}{Bang,
  J.}, \bibinfo{author}{Song, H.}, \bibinfo{year}{2023}.
\newblock \bibinfo{title}{Generating instance-level prompts for rehearsal-free
  continual learning}.
\newblock \bibinfo{journal}{2023 IEEE/CVF International Conference on Computer
  Vision (ICCV)} , \bibinfo{pages}{11813--11823}.
\bibitem[{Krizhevsky(2009)}]{Krizhevsky2009LearningML}
\bibinfo{author}{Krizhevsky, A.}, \bibinfo{year}{2009}.
\newblock \bibinfo{title}{Learning multiple layers of features from tiny
  images}.
\bibitem[{Krizhevsky et~al.(2012)Krizhevsky, Sutskever and
  Hinton}]{Krizhevsky2012ImageNetCW}
\bibinfo{author}{Krizhevsky, A.}, \bibinfo{author}{Sutskever, I.},
  \bibinfo{author}{Hinton, G.E.}, \bibinfo{year}{2012}.
\newblock \bibinfo{title}{Imagenet classification with deep convolutional
  neural networks}.
\newblock \bibinfo{journal}{Communications of the ACM} \bibinfo{volume}{60},
  \bibinfo{pages}{84 -- 90}.
\bibitem[{Lee et~al.(2022)Lee, Gomes and Zhang}]{Lee2022BalancingTS}
\bibinfo{author}{Lee, A.}, \bibinfo{author}{Gomes, H.M.},
  \bibinfo{author}{Zhang, Y.}, \bibinfo{year}{2022}.
\newblock \bibinfo{title}{Balancing the stability-plasticity dilemma with
  online stability tuning for continual learning}.
\newblock \bibinfo{journal}{2022 International Joint Conference on Neural
  Networks (IJCNN)} , \bibinfo{pages}{1--8}.
\bibitem[{Li and Hoiem(2016)}]{Li2016LearningWF}
\bibinfo{author}{Li, Z.}, \bibinfo{author}{Hoiem, D.}, \bibinfo{year}{2016}.
\newblock \bibinfo{title}{Learning without forgetting}.
\newblock \bibinfo{journal}{IEEE Transactions on Pattern Analysis and Machine
  Intelligence} \bibinfo{volume}{40}, \bibinfo{pages}{2935--2947}.
\bibitem[{Lomonaco et~al.(2020)Lomonaco, Pellegrini, L{\'o}pez, Caccia, She,
  Chen, Jodelet, Wang, Mai, V{\'a}zquez, Parisi, Churamani, Pickett, Laradji
  and Maltoni}]{Lomonaco2020CVPR2C}
\bibinfo{author}{Lomonaco, V.}, \bibinfo{author}{Pellegrini, L.},
  \bibinfo{author}{L{\'o}pez, P.R.}, \bibinfo{author}{Caccia, M.},
  \bibinfo{author}{She, Q.}, \bibinfo{author}{Chen, Y.},
  \bibinfo{author}{Jodelet, Q.}, \bibinfo{author}{Wang, R.},
  \bibinfo{author}{Mai, Z.}, \bibinfo{author}{V{\'a}zquez, D.},
  \bibinfo{author}{Parisi, G.I.}, \bibinfo{author}{Churamani, N.},
  \bibinfo{author}{Pickett, M.}, \bibinfo{author}{Laradji, I.H.},
  \bibinfo{author}{Maltoni, D.}, \bibinfo{year}{2020}.
\newblock \bibinfo{title}{Cvpr 2020 continual learning in computer vision
  competition: Approaches, results, current challenges and future directions}.
\newblock \bibinfo{journal}{Artif. Intell.} \bibinfo{volume}{303},
  \bibinfo{pages}{103635}.
\bibitem[{Lopez-Paz and Ranzato(2017)}]{lopez2017}
\bibinfo{author}{Lopez-Paz, D.}, \bibinfo{author}{Ranzato, M.},
  \bibinfo{year}{2017}.
\newblock \bibinfo{title}{Gradient episodic memory for continual learning}, in:
  \bibinfo{booktitle}{Proceedings of the 31st International Conference on
  Neural Information Processing Systems}, p. \bibinfo{pages}{6470–6479}.
\bibitem[{Loshchilov and Hutter(2017)}]{Loshchilov2016SGDRSG}
\bibinfo{author}{Loshchilov, I.}, \bibinfo{author}{Hutter, F.},
  \bibinfo{year}{2017}.
\newblock \bibinfo{title}{{SGDR}: Stochastic gradient descent with warm
  restarts}, in: \bibinfo{booktitle}{ICLR}.
\bibitem[{McCloskey and Cohen(1989)}]{McCloskey1989CatastrophicII}
\bibinfo{author}{McCloskey, M.}, \bibinfo{author}{Cohen, N.J.},
  \bibinfo{year}{1989}.
\newblock \bibinfo{title}{Catastrophic interference in connectionist networks:
  The sequential learning problem}.
\newblock \bibinfo{journal}{Psychology of Learning and Motivation}
  \bibinfo{volume}{24}, \bibinfo{pages}{109--165}.
\bibitem[{Mermillod et~al.(2013)Mermillod, Bugaiska and
  Bonin}]{Mermillod2013TheSD}
\bibinfo{author}{Mermillod, M.}, \bibinfo{author}{Bugaiska, A.},
  \bibinfo{author}{Bonin, P.}, \bibinfo{year}{2013}.
\newblock \bibinfo{title}{The stability-plasticity dilemma: investigating the
  continuum from catastrophic forgetting to age-limited learning effects}.
\newblock \bibinfo{journal}{Frontiers in Psychology} \bibinfo{volume}{4}.
\bibitem[{Radford et~al.(2021)Radford, Kim, Hallacy, Ramesh, Goh, Agarwal,
  Sastry, Askell, Mishkin, Clark, Krueger and
  Sutskever}]{Radford2021LearningTV}
\bibinfo{author}{Radford, A.}, \bibinfo{author}{Kim, J.W.},
  \bibinfo{author}{Hallacy, C.}, \bibinfo{author}{Ramesh, A.},
  \bibinfo{author}{Goh, G.}, \bibinfo{author}{Agarwal, S.},
  \bibinfo{author}{Sastry, G.}, \bibinfo{author}{Askell, A.},
  \bibinfo{author}{Mishkin, P.}, \bibinfo{author}{Clark, J.},
  \bibinfo{author}{Krueger, G.}, \bibinfo{author}{Sutskever, I.},
  \bibinfo{year}{2021}.
\newblock \bibinfo{title}{Learning transferable visual models from natural
  language supervision}, in: \bibinfo{booktitle}{International Conference on
  Machine Learning}.
\bibitem[{Rebuffi et~al.(2017)Rebuffi, Kolesnikov, Sperl and
  Lampert}]{Rebuffi2017}
\bibinfo{author}{Rebuffi, S.A.}, \bibinfo{author}{Kolesnikov, A.},
  \bibinfo{author}{Sperl, G.}, \bibinfo{author}{Lampert, C.H.},
  \bibinfo{year}{2017}.
\newblock \bibinfo{title}{icarl: Incremental classifier and representation
  learning}, in: \bibinfo{booktitle}{2017 IEEE Conference on Computer Vision
  and Pattern Recognition (CVPR)}, pp. \bibinfo{pages}{5533--5542}.
\bibitem[{Shi et~al.(2023)Shi, Zhao and Fu}]{Shi2023}
\bibinfo{author}{Shi, L.}, \bibinfo{author}{Zhao, K.}, \bibinfo{author}{Fu,
  Z.}, \bibinfo{year}{2023}.
\newblock \bibinfo{title}{Boosting separated softmax with discrimination for
  class incremental learning}.
\newblock \bibinfo{journal}{Journal of Visual Communication and Image
  Representation} \bibinfo{volume}{95}, \bibinfo{pages}{103899}.
\bibitem[{Shim et~al.(2020)Shim, Mai, Jeong, Sanner, Kim and
  Jang}]{Shim2020OnlineCC}
\bibinfo{author}{Shim, D.}, \bibinfo{author}{Mai, Z.}, \bibinfo{author}{Jeong,
  J.}, \bibinfo{author}{Sanner, S.}, \bibinfo{author}{Kim, H.J.},
  \bibinfo{author}{Jang, J.}, \bibinfo{year}{2020}.
\newblock \bibinfo{title}{Online class-incremental continual learning with
  adversarial shapley value}, in: \bibinfo{booktitle}{AAAI Conference on
  Artificial Intelligence}.
\bibitem[{Shmelkov et~al.(2017)Shmelkov, Schmid and Alahari}]{Shmelkov2017}
\bibinfo{author}{Shmelkov, K.}, \bibinfo{author}{Schmid, C.},
  \bibinfo{author}{Alahari, K.}, \bibinfo{year}{2017}.
\newblock \bibinfo{title}{Incremental learning of object detectors without
  catastrophic forgetting}, in: \bibinfo{booktitle}{2017 IEEE International
  Conference on Computer Vision (ICCV)}, pp. \bibinfo{pages}{3420--3429}.
\bibitem[{Smith et~al.(2022)Smith, Karlinsky, Gutta, Cascante-Bonilla, Kim,
  Arbelle, Panda, Feris and Kira}]{Smith2022CODAPromptCD}
\bibinfo{author}{Smith, J.}, \bibinfo{author}{Karlinsky, L.},
  \bibinfo{author}{Gutta, V.}, \bibinfo{author}{Cascante-Bonilla, P.},
  \bibinfo{author}{Kim, D.}, \bibinfo{author}{Arbelle, A.},
  \bibinfo{author}{Panda, R.}, \bibinfo{author}{Feris, R.S.},
  \bibinfo{author}{Kira, Z.}, \bibinfo{year}{2022}.
\newblock \bibinfo{title}{Coda-prompt: Continual decomposed attention-based
  prompting for rehearsal-free continual learning}.
\newblock \bibinfo{journal}{2023 IEEE/CVF Conference on Computer Vision and
  Pattern Recognition (CVPR)} , \bibinfo{pages}{11909--11919}.
\bibitem[{Sun et~al.(2023)Sun, Zhou, Ye and Zhan}]{sun2023pilot}
\bibinfo{author}{Sun, H.L.}, \bibinfo{author}{Zhou, D.W.}, \bibinfo{author}{Ye,
  H.J.}, \bibinfo{author}{Zhan, D.C.}, \bibinfo{year}{2023}.
\newblock \bibinfo{title}{Pilot: A pre-trained model-based continual learning
  toolbox}.
\newblock \bibinfo{journal}{arXiv preprint arXiv:2309.07117} .
\bibitem[{Vaswani et~al.(2017)Vaswani, Shazeer, Parmar, Uszkoreit, Jones,
  Gomez, Kaiser and Polosukhin}]{Vaswani2017AttentionIA}
\bibinfo{author}{Vaswani, A.}, \bibinfo{author}{Shazeer, N.M.},
  \bibinfo{author}{Parmar, N.}, \bibinfo{author}{Uszkoreit, J.},
  \bibinfo{author}{Jones, L.}, \bibinfo{author}{Gomez, A.N.},
  \bibinfo{author}{Kaiser, L.}, \bibinfo{author}{Polosukhin, I.},
  \bibinfo{year}{2017}.
\newblock \bibinfo{title}{Attention is all you need}, in:
  \bibinfo{booktitle}{Neural Information Processing Systems}.
\bibitem[{van~de Ven and Tolias(2019)}]{vandeVen2019ThreeSF}
\bibinfo{author}{van~de Ven, G.M.}, \bibinfo{author}{Tolias, A.S.},
  \bibinfo{year}{2019}.
\newblock \bibinfo{title}{Three scenarios for continual learning}.
\newblock \bibinfo{journal}{ArXiv} \bibinfo{volume}{abs/1904.07734}.
\bibitem[{van~de Ven et~al.(2022)van~de Ven, Tuytelaars and
  Tolias}]{vandeVen2022ThreeTO}
\bibinfo{author}{van~de Ven, G.M.}, \bibinfo{author}{Tuytelaars, T.},
  \bibinfo{author}{Tolias, A.S.}, \bibinfo{year}{2022}.
\newblock \bibinfo{title}{Three types of incremental learning}.
\newblock \bibinfo{journal}{Nature Machine Intelligence} \bibinfo{volume}{4},
  \bibinfo{pages}{1185 -- 1197}.
\bibitem[{Wah et~al.(2011)Wah, Branson, Welinder, Perona and Belongie}]{cub200}
\bibinfo{author}{Wah, C.}, \bibinfo{author}{Branson, S.},
  \bibinfo{author}{Welinder, P.}, \bibinfo{author}{Perona, P.},
  \bibinfo{author}{Belongie, S.}, \bibinfo{year}{2011}.
\newblock \bibinfo{title}{The {C}altech-{UCSD} birds-200-2011 dataset} .
\bibitem[{Wang et~al.(2023)Wang, Zhou, Liu, Ye, Bian, chuan Zhan and
  Zhao}]{Wang2023BEEFBC}
\bibinfo{author}{Wang, F.L.}, \bibinfo{author}{Zhou, D.W.},
  \bibinfo{author}{Liu, L.}, \bibinfo{author}{Ye, H.J.}, \bibinfo{author}{Bian,
  Y.}, \bibinfo{author}{chuan Zhan, D.}, \bibinfo{author}{Zhao, P.},
  \bibinfo{year}{2023}.
\newblock \bibinfo{title}{Beef: Bi-compatible class-incremental learning via
  energy-based expansion and fusion}, in: \bibinfo{booktitle}{International
  Conference on Learning Representations}.
\bibitem[{Wang et~al.(2022a)Wang, Huang and Hong}]{Wang2022SPromptsLW}
\bibinfo{author}{Wang, Y.}, \bibinfo{author}{Huang, Z.}, \bibinfo{author}{Hong,
  X.}, \bibinfo{year}{2022}a.
\newblock \bibinfo{title}{S-prompts learning with pre-trained transformers: An
  occam's razor for domain incremental learning}, in:
  \bibinfo{booktitle}{NeurIPS}.
\bibitem[{Wang et~al.(2022b)Wang, Ma, Huang, Wang, Su and
  Hong}]{Wang2022IsolationAI}
\bibinfo{author}{Wang, Y.}, \bibinfo{author}{Ma, Z.}, \bibinfo{author}{Huang,
  Z.}, \bibinfo{author}{Wang, Y.}, \bibinfo{author}{Su, Z.},
  \bibinfo{author}{Hong, X.}, \bibinfo{year}{2022}b.
\newblock \bibinfo{title}{Isolation and impartial aggregation: A paradigm of
  incremental learning without interference}, in: \bibinfo{booktitle}{AAAI
  Conference on Artificial Intelligence}.
\bibitem[{Wang et~al.(2022c)Wang, Zhang, Ebrahimi, Sun, Zhang, Lee, Ren, Su,
  Perot, Dy and Pfister}]{Wang2022DualPromptCP}
\bibinfo{author}{Wang, Z.}, \bibinfo{author}{Zhang, Z.},
  \bibinfo{author}{Ebrahimi, S.}, \bibinfo{author}{Sun, R.},
  \bibinfo{author}{Zhang, H.}, \bibinfo{author}{Lee, C.Y.},
  \bibinfo{author}{Ren, X.}, \bibinfo{author}{Su, G.}, \bibinfo{author}{Perot,
  V.}, \bibinfo{author}{Dy, J.G.}, \bibinfo{author}{Pfister, T.},
  \bibinfo{year}{2022}c.
\newblock \bibinfo{title}{{D}ual{P}rompt: Complementary prompting for
  rehearsal-free continual learning}, in: \bibinfo{booktitle}{ECCV}.
\bibitem[{Wang et~al.(2021)Wang, Zhang, Lee, Zhang, Sun, Ren, Su, Perot, Dy and
  Pfister}]{Wang2021LearningTP}
\bibinfo{author}{Wang, Z.}, \bibinfo{author}{Zhang, Z.}, \bibinfo{author}{Lee,
  C.Y.}, \bibinfo{author}{Zhang, H.}, \bibinfo{author}{Sun, R.},
  \bibinfo{author}{Ren, X.}, \bibinfo{author}{Su, G.}, \bibinfo{author}{Perot,
  V.}, \bibinfo{author}{Dy, J.G.}, \bibinfo{author}{Pfister, T.},
  \bibinfo{year}{2021}.
\newblock \bibinfo{title}{Learning to prompt for continual learning}.
\newblock \bibinfo{journal}{2022 IEEE/CVF Conference on Computer Vision and
  Pattern Recognition (CVPR)} , \bibinfo{pages}{139--149}.
\bibitem[{Welling(2009)}]{welling2009}
\bibinfo{author}{Welling, M.}, \bibinfo{year}{2009}.
\newblock \bibinfo{title}{Herding dynamical weights to learn}, in:
  \bibinfo{booktitle}{Proceedings of the 26th Annual International Conference
  on Machine Learning}, \bibinfo{publisher}{Association for Computing
  Machinery}, \bibinfo{address}{New York, NY, USA}. p.
  \bibinfo{pages}{1121–1128}.
\bibitem[{Yun et~al.(2019)Yun, Han, Oh, Chun, Choe and Yoo}]{Yun2019CutMixRS}
\bibinfo{author}{Yun, S.}, \bibinfo{author}{Han, D.}, \bibinfo{author}{Oh,
  S.J.}, \bibinfo{author}{Chun, S.}, \bibinfo{author}{Choe, J.},
  \bibinfo{author}{Yoo, Y.J.}, \bibinfo{year}{2019}.
\newblock \bibinfo{title}{Cutmix: Regularization strategy to train strong
  classifiers with localizable features}.
\newblock \bibinfo{journal}{2019 IEEE/CVF International Conference on Computer
  Vision (ICCV)} , \bibinfo{pages}{6022--6031}.
\bibitem[{Zhao et~al.(2019)Zhao, Xiao, Gan, Zhang and
  Xia}]{Zhao2019MaintainingDA}
\bibinfo{author}{Zhao, B.}, \bibinfo{author}{Xiao, X.}, \bibinfo{author}{Gan,
  G.}, \bibinfo{author}{Zhang, B.}, \bibinfo{author}{Xia, S.},
  \bibinfo{year}{2019}.
\newblock \bibinfo{title}{Maintaining discrimination and fairness in class
  incremental learning}.
\newblock \bibinfo{journal}{2020 IEEE/CVF Conference on Computer Vision and
  Pattern Recognition (CVPR)} , \bibinfo{pages}{13205--13214}.
\bibitem[{Zhao et~al.(2020)Zhao, Wang, Fu, Wu and
  Li}]{Zhao2020MemoryEfficientCL}
\bibinfo{author}{Zhao, H.}, \bibinfo{author}{Wang, H.}, \bibinfo{author}{Fu,
  Y.}, \bibinfo{author}{Wu, F.}, \bibinfo{author}{Li, X.},
  \bibinfo{year}{2020}.
\newblock \bibinfo{title}{Memory-efficient class-incremental learning for image
  classification}.
\newblock \bibinfo{journal}{IEEE Transactions on Neural Networks and Learning
  Systems} \bibinfo{volume}{33}, \bibinfo{pages}{5966--5977}.
\bibitem[{Zhou et~al.(2024)Zhou, Cai, Ye, Zhan and Liu}]{aper}
\bibinfo{author}{Zhou, D.W.}, \bibinfo{author}{Cai, Z.W.}, \bibinfo{author}{Ye,
  H.J.}, \bibinfo{author}{Zhan, D.C.}, \bibinfo{author}{Liu, Z.},
  \bibinfo{year}{2024}.
\newblock \bibinfo{title}{Revisiting class-incremental learning with
  pre-trained models: Generalizability and adaptivity are all you need}.
\newblock \bibinfo{journal}{International Journal of Computer Vision}
  \DOIprefix\doi{10.1007/s11263-024-02218-0}.
\bibitem[{Zhou et~al.(2023)Zhou, Wang, Ye and Zhan}]{Zhou2022AMO}
\bibinfo{author}{Zhou, D.W.}, \bibinfo{author}{Wang, Q.}, \bibinfo{author}{Ye,
  H.J.}, \bibinfo{author}{Zhan, D.C.}, \bibinfo{year}{2023}.
\newblock \bibinfo{title}{A model or 603 exemplars: Towards memory-efficient
  class-incremental learning}, in: \bibinfo{booktitle}{ICLR}.
\bibitem[{Zhou et~al.(2021)Zhou, Yang, Loy and Liu}]{Zhou2021LearningTP}
\bibinfo{author}{Zhou, K.}, \bibinfo{author}{Yang, J.}, \bibinfo{author}{Loy,
  C.C.}, \bibinfo{author}{Liu, Z.}, \bibinfo{year}{2021}.
\newblock \bibinfo{title}{Learning to prompt for vision-language models}.
\newblock \bibinfo{journal}{International Journal of Computer Vision}
  \bibinfo{volume}{130}, \bibinfo{pages}{2337 -- 2348}.
\bibitem[{Zhou et~al.(2019)Zhou, Mai, Zhang, Xu, Wu and Davis}]{Zhou2019M2KDMA}
\bibinfo{author}{Zhou, P.}, \bibinfo{author}{Mai, L.}, \bibinfo{author}{Zhang,
  J.}, \bibinfo{author}{Xu, N.}, \bibinfo{author}{Wu, Z.},
  \bibinfo{author}{Davis, L.S.}, \bibinfo{year}{2019}.
\newblock \bibinfo{title}{M2kd: Multi-model and multi-level knowledge
  distillation for incremental learning}.
\newblock \bibinfo{journal}{ArXiv} \bibinfo{volume}{abs/1904.01769}.

\end{thebibliography}

\end{document}